\documentclass[twoside,11pt]{article}

\usepackage{blindtext}

\usepackage{jmlr2e}
\usepackage{nicefrac}              
\usepackage{microtype}             
\usepackage{xcolor}                
\usepackage{xspace}
\usepackage{algorithm, algorithmicx}
\usepackage{mathtools}
\usepackage{enumitem}
\usepackage{subfigure}
\usepackage{threeparttable}
\usepackage{multirow}
\usepackage{multicol}
\usepackage{graphicx}
\usepackage{caption}
\usepackage{MnSymbol}
\usepackage{marvosym}

\newcommand{\model}{\textsc{DIFFormer}\xspace}

\newcommand{\revise}[1]{{\color{black} #1}}

\usepackage{lastpage}
\jmlrheading{26}{2025}{1-\pageref{LastPage}}{12/23; Revised
3/25}{6/25}{23-1672}{Qitian Wu, David Wipf and Junchi Yan}
\ShortHeadings{Transformers from Diffusion: A Unified Framework for Neural Message Passing}{Wu, Wipf and Yan}

\firstpageno{1}

\begin{document}

\title{Transformers from Diffusion: A Unified Framework for Neural Message Passing}

\author{\name Qitian Wu$^*$ \email wuqitian@mit.edu \\
       \addr Eric and Wendy Schmidt Center, Broad Institute of MIT and Harvard
       \AND
       \name David Wipf \email davidwipf@gmail.com \\
       \addr Amazon Web Services AI Lab
       \AND
       \name Junchi Yan\thanks{Correspondence authors.} \email yanjunchi@sjtu.edu.cn \\
       \addr School of Artificial Intelligence, Shanghai Jiao Tong University}

\editor{Tong Zhang}

\maketitle

\begin{abstract}
    Learning representations for structured data with certain geometries (e.g., observed or unobserved) is a fundamental challenge, wherein message passing neural networks (MPNNs) have become a de facto class of model solutions. In this paper, inspired by physical systems, we propose an energy-constrained diffusion model, which integrates the inductive bias of diffusion on manifolds with layer-wise constraints of energy minimization. We identify that the diffusion operators have a one-to-one correspondence with the energy functions implicitly descended by the diffusion process, and the finite-difference iteration for solving the energy-constrained diffusion system induces the propagation layers of various types of MPNNs operating on observed or latent structures. This leads to a unified mathematical framework for common neural architectures whose computational flows can be cast as message passing (or its special case), including MLPs, GNNs, and Transformers. Building on these insights, we devise a new class of neural message passing models, dubbed diffusion-inspired Transformers (DIFFormer), whose global attention layers are derived from the principled energy-constrained diffusion framework. Across diverse datasets ranging from real-world networks to images, texts, and physical particles, we demonstrate that the new model achieves promising performance in scenarios where the data structures are observed (as a graph), partially observed, or entirely unobserved.\footnote{Code available at \url{https://github.com/qitianwu/DIFFormer}}
\end{abstract}


\section{Introduction}\label{sec-intro}

Real-world data are generated from a convoluted interactive process whose underlying physical principles often involve inter-connections of certain forms. Such a nature violates the common hypothesis of standard representation learning paradigms assuming that observed data are independently sampled. The challenge, however, is that due to the absence of prior knowledge about ground-truth data generation, it can be practically prohibitive to build feasible methodology for uncovering the latent structures that embody the inter-connecting patterns. To address this issue, prior works, e.g., \cite{pointcloud-19,LDS-icml19,jiang2019glcn,Bayesstruct-aaai19}, consider encoding the potential interactions as estimated structures in latent space, but this requires sufficient degrees of freedom that significantly increases learning difficulty from limited labels~\citep{fatemi2021slaps} and hinders the scalability to large systems~\citep{wunodeformer}. 



Turning to a simpler problem setting where putative inter-connections are instantiated as an observed graph, remarkable progress has been made in designing expressive architectures such as message passing neural networks (MPNNs), a dominant class of graph neural networks (GNNs)~\citep{scarselli2008gnnearly,GCN-vallina,GAT,SGC-icml19,gcnii,RWLS-icml21}, for harnessing observed structures as a geometric prior~\citep{geometriclearning-2017}. 
However, the observed graphs can be incomplete or noisy, due to error-prone data collection, or generated by an artificial construction independent from downstream targets. The potential inconsistency between observation and the underlying data geometry would presumably elicit systematic bias between structured representation of graph-based learning and the true inter-dependency. While a plausible remedy is to learn more useful latent structures from the data, this unfortunately brings the previously-mentioned obstacles to the fore.

\begin{figure}[tb!]
			\centering
			\hspace{5pt}
			\includegraphics[width=0.95\textwidth]{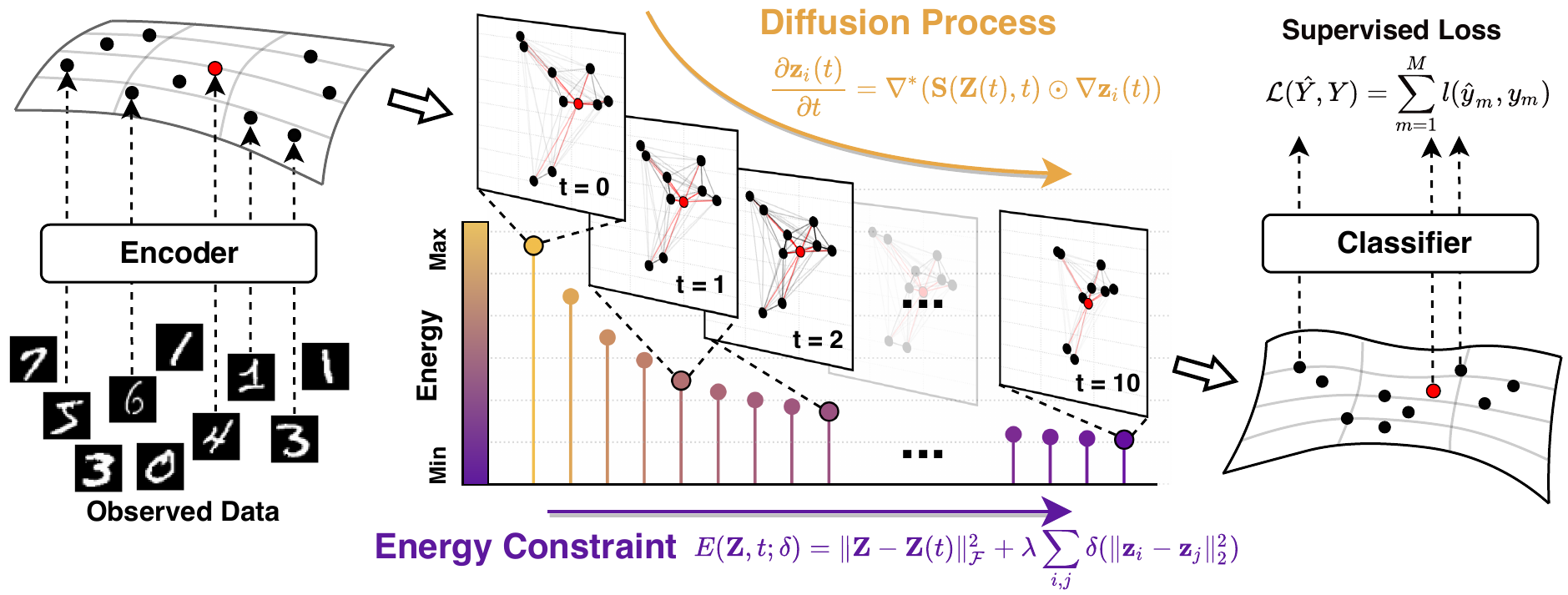}
		\caption{\looseness=-1 An illustration of the general idea behind DIFFormer induced by the energy-constrained diffusion model. It treats observed data as nodes on the manifold and encodes them into hidden states through a diffusion process aimed at minimizing a regularized energy. This design allows feature propagation among arbitrary node pairs at each layer with optimal inter-connecting structures for informed prediction in downstream tasks.}
		\label{fig:model}
\end{figure}

To resolve the dilemma, we propose a principled theoretical framework stemming from a two-fold physics-based inspiration as illustrated in Figure~\ref{fig:model}. The model is defined through feed-forward continuous dynamics (i.e., a diffusion PDE equation) involving observed data as locations (a.k.a. nodes) on Riemannian manifolds with \emph{latent} structures, upon which the features of each node act as heat flowing over the underlying geometry~\citep{hamzi2021learning}. Such a diffusion model serves an important \emph{inductive bias} for leveraging global information from other data points to obtain more informative representations of each individual. Particularly, in the general case (i.e., the non-local, non-homogeneous diffusion), the model allows for feature propagation between arbitrary node pairs at each layer, and adaptively navigates this process by layer-dependent pairwise connectivity weights. Moreover, for guiding the representations towards some ideal constraints of internal consistency, we introduce a principled energy function that enforces layer-wise \emph{regularization} on the evolutionary directions. The energy function provides another view (from a macroscopic standpoint) into the desired representations with low global energy that are produced, i.e., soliciting a steady state that gives rise to informed predictions in downstream tasks. 


As a justification for the tractability of the above general methodology, our analysis reveals the underlying equivalence between the finite-difference iterations of the diffusion process and the unfolded minimization dynamics for an associated regularized energy. This result suggests a closed-form optimal solution for the diffusion dynamics trajectory that updates node representations by the ones of all the others towards giving a rigorous decrease of the global energy. Based on this, we also show that the energy-constrained diffusion model has essential connections with various types of existing MPNNs (like GCN, GIN, APPNP, GAT, etc.) as well as Transformers that can be considered as an extension of MPNN on latent complete graphs. Furthermore, as by-product results, we derive the convergence speed of the energy minimization by the diffusion dynamics and discuss how to alleviate the potential risk of over-smoothing that can manifest as a degenerate global optimum.

On top of the theory, we propose a new class of neural encoders, Diffusion-inspired Transformers (DIFFormer), along with two practical instantiations. The first model version is equipped with a simple diffusion-inspired attention function that only requires $\mathcal O(N)$ complexity ($N$ for node number) for computing all-pair interactions in each layer. The second model version is endowed with a more expressive non-linear attention function that can learn complex latent structures. We apply these models to an extensive range of experimental datasets, showcasing wide applicability and practical efficacy for learning effective representations of structured data. In particular, we consider experiments on both graph-based predictive tasks, where the input graphs have disparate properties (such as homophilous graphs, heterophilic graphs, large-sized graphs and incompletely observed graphs), and standard predictive tasks, where the structures are unobserved (such as images, texts and physical particles). The results consistently show the superiority of our models.

\subsection{Related Works}\label{sec-intro-related}

To provide more background information and properly position our contributions within the community, we review salient related work and discuss connections with ours. 

\subsubsection{Neural Diffusion on Graphs}\label{sec-intro-related-diff}

The diffusion-based learning has gained increasing research interests, as the continuous dynamics can serve as an inductive bias incorporated with prior knowledge of the tasks at hand~\citep{lagaris1998artificial,chen2018neuralode}. One category directly solves a continuous process of differential equations, e.g., ~\cite{grand} 
revealing the association between the discretization of graph diffusion equations and the feed-forward updating rules of GNNs. Along this direction, recent works~\citep{beltrami,GRAND++,bodnar2022neural,choi2023gread} leverage diffusion equations as a mathematical framework for analyzing GNN behavior and devising continuous models that utilize differentiable PDE-solving tools for training. 

Another line of research investigates PDE-inspired learning using the diffusion perspective as a principled guideline on top of which (discrete) neural network-based approaches are designed for node classification~\citep{atwood2016diffusion,klicpera2019diffusion,xu2020heat}, addressing over-smoothing~\citep{rusch2022gradient}, knowledge distillation~\citep{yang2022geometric} and topological generalization~\citep{wu2025advective}. Our work leans on PDE-inspired learning and introduces a new diffusion model that is implicitly defined as minimizing a regularized energy. Our analysis also reveals the underlying equivalence between the numerical iterations of diffusion equations and unfolding the minimization dynamics of a corresponding energy. The results illuminate the fundamental connection between graph diffusion equations and energy optimization systems. More importantly, as we will show in the following sections, such a principled perspective (from the energy-constrained diffusion) brings up an interpretable framework for existing message passing neural networks and can be used for navigating new architecture designs.

\subsubsection{Message Passing Neural Networks}\label{sec-intro-related-gnn}

Graph neural networks~\citep{scarselli2008gnnearly} 
have become the mainstream class of neural encoders for representation learning on structured data with observed geometries. Most existing GNNs adopt message-passing-based architectures, and are to a large extent interchangeably called message passing neural networks (MPNNs) in the literature.
With the pioneering work of graph convolution networks~\citep{GCN-vallina} that show promising performance on semi-supervised (node) classification tasks, there is a surge of recent work exploring various expressive MPNN architectures equipped with advanced message passing designs e.g., ~\cite{gin,graphsage,jknet-icml18,mixhop-icml19,appnp,gcnii,h2gcn-neurips20,gprgnn-iclr21}. The challenge, however, is that due to the diversity of graph-structured data that can have disparate scales, sizes, topological properties, etc., current models designed for particular tasks are often hard to transfer to others outside its experimental settings. 

Furthermore, from an architectural view, the majority of existing MPNNs operates the message passing per layer within observed edges of input graphs, which could limit efficacy in scenarios where the graphs are noisy or incomplete. To resolve this, several recent works propose to learn latent graph structures from data that can boost MPNNs towards better representations~\citep{LDS-icml19,IDGL-neurips20,jiang2019glcn,fatemi2021slaps,wunodeformer,wu2023difformer,wu2023sgformer,deng2024polynormer}. These approaches generalize message-passing-based schemes to broader regimes where interactions are modeled by latent structures. Our work aims to provide a theoretical framework that can interpret the message passing rules of GNNs as numerical iterations of a diffusion process that descends a regularized energy in an interpretable form. As we will show in later sections, this principled perspective can be utilized to help understand the behavior of various MPNNs and the mechanism of message passing over observed or latent graph structures.

\subsection{Contributions and Organization}\label{sec-intro-contribution}

Before we delve into the proposed model, we summarize the main contributions of this paper along with pointers to the relevant sections.

\begin{itemize}
    \item We propose a principled theoretical framework for representation learning on structured data. The framework is built upon an energy-constrained diffusion model that integrates the continuous dynamics of diffusion equations with minimization constraints of a global energy. The model offers a new aspect for learning effective representations with either observed structures or unobserved latent structures. (See Section~\ref{sec-model}).

    \item Our analysis shows that when the diffusion equations are linear with constant diffusivity, the energy functions minimized by the diffusion dynamics take a linear, quadratic form. Furthermore, the finite-difference iterations of the energy-constrained diffusion dynamics induce the propagation layers of common \emph{convolutional} MPNNs, such as GCN, GIN, APPNP, etc. We also illuminate the convergence speed of energy minimization by the diffusion process and theoretically discuss how to avoid the potential risk of over-smoothing caused by the degenerate solutions. (See Section~\ref{sec-gnn}).

    \item Pushing further, we consider the more general case where the diffusivity can change with time that gives rise to non-linear diffusion equations. We show that in such cases, the diffusion process descends a non-convex energy function that assigns certain tolerance on node pairs with large distances in latent space. Correspondingly, the finite-difference iterations of the energy-constrained diffusion dynamics induce the propagation layers of \emph{attentional} MPNNs (e.g., GAT) as well as Transformers. On top of these results, we propose a new class of neural encoders, inspired by the energy-constrained diffusion, that resort to message passing between arbitrary node pairs with diffusion-based attention functions. The model implementation possesses expressivity for learning all-pair interactions and scalability with linear complexity w.r.t.~node numbers. (See Section~\ref{sec-trans}).
    
    \item To validate the effectiveness of our model and demonstrate its applicability, we apply the model to a wide spectrum of predictive tasks on experimental datasets ranging from real-world networks (including homophilous graphs, heterophilic graphs, large graphs and incompletely observed graphs) to images and physical particles. The results show that our model can significantly outperform strong MPNNs competitors in the scenarios where the graph structures are observed or unobserved. 
    (See Section~\ref{sec-exp}).
\end{itemize}

\textbf{Comparison with the Conference Paper on ICLR 2023~\citep{wu2023difformer}.} On the basis of our conference paper, we have made substantial extensions that entail new theoretical analysis and empirical results enriching and deepening the technical contents. In Section~\ref{sec-preliminary}, we add more technical background about the manifold diffusion as foundations of our proposed model. In Section~\ref{sec-gnn-connection}, we analyze linear diffusion equations with constant diffusivity, and show its connection with the minimization of the convex energy function of quadratic forms. We also illustrate how to derive various types of GNNs, such as GCN, GIN and APPNP, starting from the energy-constrained diffusion with constant diffusivity. In Section~\ref{sec-gnn-dis}, we derive the upper and lower bound of the energy at each layer and discuss how to avoid the over-smoothing issue with a source term incorporated into the diffusion equation. In Section~\ref{sec-trans-attn-existing}, we present discussions on how to derive dot-then-exponential Softmax attention via the energy-constrained diffusion with time-dependent diffusivity. And in Section~\ref{sec-trans-dis}, we supplement detailed discussions along with analysis on how to extend our model with feature transformations, non-linear activations, and graph inductive biases. For the experiments, we add empirical comparisons on five additional datasets including heterophilic graphs (in Section~\ref{sec-exp-observed}) and physical particles (in Section~\ref{sec-exp-unobserved}) and discussions on addressing the potential over-smoothing (in Section~\ref{sec-exp-dis}). 

\section{Preliminary and Background}\label{sec-preliminary}

In this section, we introduce some technical background about diffusion on manifolds as preliminary to our model. 
We consider an abstract domain denoted by $\Omega$, which, for the purposes of our study, is assumed to be a Riemannian manifold~\citep{eells1964harmonic}. 
A fundamental distinction between an $n$-dimensional Riemannian manifold and a Euclidean space lies in its unique property of being locally Euclidean. This suggests that for each point $u\in\Omega$, there exists a $n$-dimensional Euclidean tangent space $T_u\Omega \cong \mathbb{R}^n$ that locally represents the structure of $\Omega$. We denote by $T\Omega$ the collection of these tangent spaces, which has a smoothly varying inner product (often known as the \emph{Riemannian metric}). 

For some physical quantity (e.g., temperature), it can be described by a function of the form $z:\Omega \rightarrow \mathbb{R}$ that is a \emph{scalar field} on $\Omega$. This also associates to every point $u\in \Omega$ a tangent vector $\mathtt z(u)\in T_u\Omega$ that can be considered as a local infinitesimal displacement of a {\em (tangent) vector field} $\mathtt z:\Omega \rightarrow T\Omega$. Let $\mathcal Z(\Omega)$ and $\mathcal Z(T\Omega)$ denote the functional spaces of scalar and (tangent) vector fields on $\Omega$, respectively. Then the inner products on $\mathcal Z(\Omega)$ and $\mathcal Z(T\Omega)$ can be denoted by $\langle z, z'\rangle$ and $\llangle \mathtt z, \mathtt z'\rrangle$, respectively. The {\em gradient} operator $\nabla: \mathcal Z(\Omega) \rightarrow \mathcal Z(T\Omega)$ transforms scalar fields into vector fields that represent the local direction of the steepest change of $\mathcal Z(\Omega)$.
The {\em divergence} operator
$\nabla^*: \mathcal Z(T\Omega) \rightarrow \mathcal Z(\Omega)$ transforms the vector fields into scalar fields that quantify the flow of $\mathtt z$ through an infinitesimal volume. These two operators are adjoint w.r.t.~the inner products: $\langle \nabla z, \mathtt z \rangle = \llangle 
z, \nabla^* \mathtt z \rrangle $.

\begin{figure}
    \centering
    \includegraphics[width=0.98\linewidth]{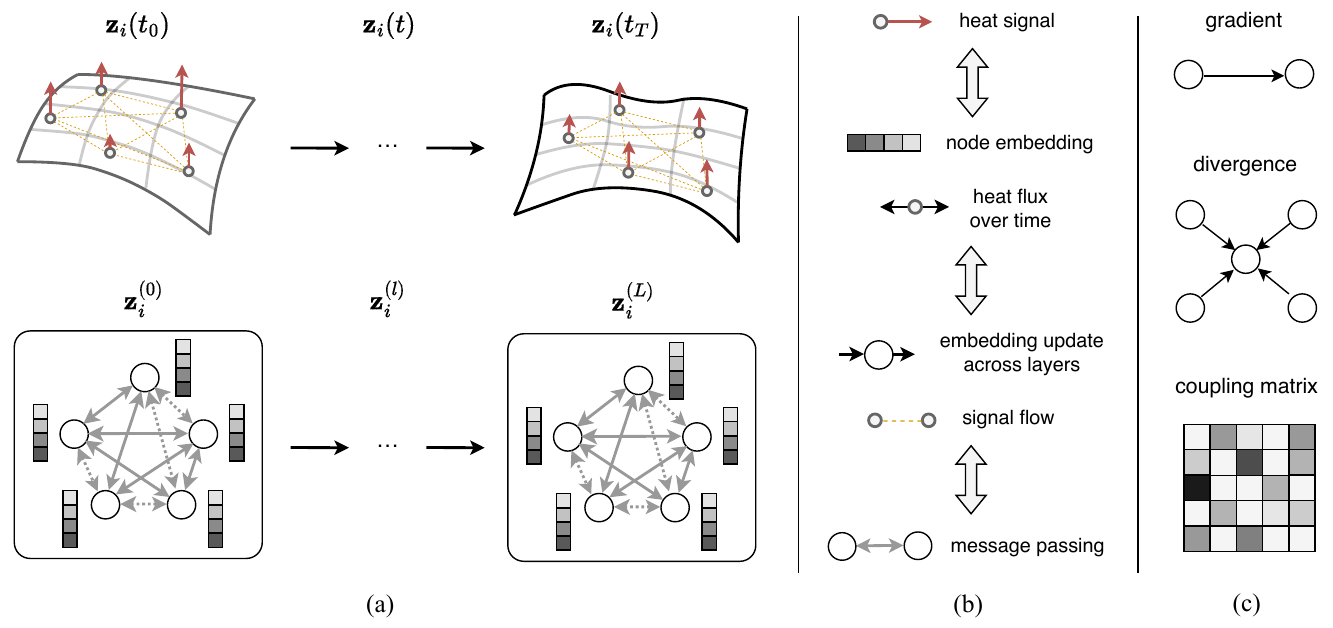}
    \caption{\revise{Illustration of (a) the connection between diffusion process on manifolds and message passing neural networks (MPNNs), (b) the analogy bridging the concepts of two models, and (c) definitions of key components in diffusion equation models on graphs.}}
    \label{fig:concept-illustration}
\end{figure}

The concept of diffusion is widely used in a variety of fields, including physics, chemistry, sociology, economics, etc. In general sense, the diffusion process describes the transfer of a certain quantity (e.g., heat, density, ideas, price values, etc.) inside a physical system due to concentration differences: the quantity spreads out from the points (or locations) with high concentrations of the quantity to others. For the quantity over time $z(u, t): \Omega \times [0, \infty) \rightarrow \mathbb R$, the diffusion process is described via a \emph{diffusion equation}, a PDE with initial conditions~\citep{freidlin1993diffusion,medvedev2014nonlinear,romeny2013geometry}:
\begin{equation}\label{eqn-def-diff}
    \frac{\partial z(u, t)}{\partial t} = \nabla^*\left( F(u, t) \odot \nabla  z(u, t)\right), ~~~ \mbox{s. t.} ~~ z(u, 0) = z_0(u), t\geq 0, u\in \Omega,
\end{equation}
where $F(u, t)$ denotes the diffusivity, $\odot$ denotes the Hadamard product, and Eqn.~\ref{eqn-def-diff} can be incorporated with additional boundary conditions if $\Omega$ has a boundary.
The physical implication of Eqn.~\ref{eqn-def-diff} is that the temporal change of $z(u, t)$ at location $u$ equals to the flow that spatially enters into $u$ within infinitesimal time. \revise{This resembles the main design of message passing neural networks (MPNNs), where as layers increase (time goes by), the embeddings (physical quantity) of connected nodes (adjacent locations) are propagated to update that of each other. More illustration on how these two perspectives are bridged through analogy is outlined in Fig.~\ref{fig:concept-illustration} and will be elaborated in Sec.~\ref{sec-model}.}

The diffusivity in diffusion equations determines the evolutionary direction of the diffusion system. If the diffusion is homogeneous, $F(u, t)$ remains a constant scalar for arbitrary $u$ and $t$. For non-homogeneous systems, the diffusion can be isotropic ($F(u, t)$ becomes a scalar-valued function and is location-dependent) and anisotropic ($F(u, t)$ becomes a $n\times n$ matrix-valued function and is location- and direction-dependent)~\citep{weickert1998anisotropic}. \revise{As will be introduced in later sections, different instantiations of diffusivity will give rise to specific forms of MPNNs such as MLP, GCN, GAT and Transformers.}

Another concept intimately related to diffusion is the energy, a measure of how variable the physical quantity is in the system~\citep{evans2022partial}. For the quantity $z(u)$ on $\Omega$, the Dirichlet energy is a quadratic functional on the Sobolev space and returns a real number
\begin{equation}\label{eqn-dirichlet-energy}
    E(z) = \int_{u\in \Omega} \|\nabla z(u)\|^2 du.
\end{equation}
The Dirichlet energy is non-negative, due to that it is the integral of a non-negative quantity.

\section{Model Formulation: Energy-Constrained Diffusion}\label{sec-model}

In this section, we introduce the general formulation of our energy-constrained diffusion model and its inherent connection with neural message passing. We will begin with a geometric diffusion model which is characterized by a diffusion PDE equation with flexible instantiations of the diffusivity. The latter enables us to bridge the numerical iterations of the PDE with different types of MPNNs. Built upon this, we will probe how the energy minimization perspective can be organically incorporated as a physics-inspired prior of the diffusion system in the form of constraints, which gives rise to energy-constrained diffusion. 

\textbf{Graph Notations.} We assume a graph to be $\mathcal G = (\mathcal V, \mathcal E)$ where $\mathcal V=\{i\}$ denotes the node set and $\mathcal E=\{(i,j)\}$ denotes the edge set. For node $i\in \mathcal V$, it has an input feature vector $\mathbf x_i \in \mathbb R^D$. The edge set is associated with an adjacency matrix $\mathbf A = [a_{ij}]_{i, j \in \mathcal V}$ where $a_{ij} = 1$ if $(i, j)\in \mathcal E$ and 0 otherwise. We use $\mathbf D = \mbox{diag}(d_i)_{i\in \mathcal V}$ to denote the diagonal degree matrix of $\mathbf A$ where $d_i$ denotes the degree of node $i$. The problem of our interest is how to obtain effective node-level representations (a.k.a. embeddings) $\mathbf z_i \in \mathbb R^d$ based on their initial features and graph structures. Beyond the observed edges $\mathcal E$, the non-trivial challenge stems from the latent structures that are not observed as input yet inter-connect the nodes in data generation. Without loss of generality, there also exist cases where no graph structure is observed, i.e., $\mathcal E = \varnothing$, though the inter-dependence among nodes cannot be ignored. 


\subsection{Geometric Diffusion with Observed/Latent Structures}\label{sec-model-diff}

The starting point of our model is rooted on an analogy that treats nodes in the graph (i.e., $i\in \mathcal V$) as locations on a Riemannian manifold (i.e., $u \in \Omega$)~\citep{rosenberg1997laplacian}, node embeddings as the physical quantity of interest
and the update of node embeddings per layer as heat flux through time~\citep{grand}. \revise{A high-level illustration is presented in Fig.~\ref{fig:concept-illustration}.}

To be specific, each node $i\in \mathcal V$ has a $d$-dimensional node embedding $\mathbf z_i$ (where $d$ is the hidden size) that is updated layer by layer, and we model the node embedding as a vector-valued function $\mathbf z_i(t): [0, \infty) \rightarrow \mathbb R^d$ that evolves with time\footnote{Since node embeddings are often $d$-dimensional vectors, we extend the scalar-valued quantity $z(u, t)$ commonly studied in physical systems to a vector-valued function $\mathbf z_i(t)$ in the analogy.}. We denote by $\mathbf Z(t) = [\mathbf z_i(t)]_{i\in \mathcal V}$ the stack of node embeddings, and the (heat) diffusion process that describes the evolution of $\mathbf Z(t)$ can be written as a partial differential equation (PDE):
\begin{equation}\label{eqn-diffuse}
    \frac{\partial \mathbf Z(t)}{\partial t} = \nabla^*\left(\mathbf F(t) \odot \nabla \mathbf Z(t)\right), ~~~ \mbox{s. t.} ~~ \mathbf Z(0) = [\mathbf x_i]_{i=1}^N, ~~ t\geq 0, 
\end{equation}
where the function $\mathbf F(t): [0, \infty) \rightarrow \mathbb R_+^{|\mathcal V| \times |\mathcal V|}$ defines the \emph{diffusivity} between any pair at time $t$. The gradient operator $\nabla$ converts node features (analogous
to scalar fields on manifolds) into edge features (analogous
to vector fields on manifolds) that measure the difference between source and target nodes, i.e., $(\nabla \mathbf Z(t))_{ij} = \mathbf z_j(t) - \mathbf z_i(t)$. The divergence operator $\nabla^*$ takes edge features into node features, by summing up information flows through a point:
\begin{equation}
    \nabla^*\left( \mathbf F(t) \odot \nabla \mathbf Z(t)\right)_i = \left (\mathbf S(t) \cdot \nabla \mathbf Z(t) \right )_i = \sum_{j\in \mathcal V} s_{ij}(t) \left(\nabla \mathbf Z(t)\right)_{ij},
\end{equation}
where $\mathbf S(t) = [s_{ij}(t)]_{i, j\in \mathcal V}$ is a coupling matrix associated with the diffusivity $\mathbf F(t)$.

Then with the gradient and divergence operators incorporated, Eqn.~\ref{eqn-diffuse} can be explicitly written as
\begin{equation}\label{eqn-diffuse2}
    \frac{\partial \mathbf z_i(t)}{\partial t} = \sum_{j\in \mathcal V} s_{ij}(t) (\mathbf z_j(t) - \mathbf z_i(t)).
\end{equation} 
Such a diffusion process can serve as an inductive bias that guides the model to use other nodes' information at every layer (which can be seen as the discretization of time) for learning informative node representations. 

We can adopt numerical methods to solve the continuous dynamics in Eqn.~\ref{eqn-diffuse2}. For instance, using the explicit Euler scheme involving finite differences with step size $\tau$, i.e., $\frac{\partial \mathbf z_i(t)}{\partial t} \approx \frac{\mathbf z_i^{(k+1)} - \mathbf z_i^{(k)}}{\tau}$, after some re-arranging we have
\begin{equation}\label{eqn-diffuse-iter}
    \mathbf z_i^{(k+1)} = \left (1 - \tau \sum_{j\in \mathcal V} s_{ij}^{(k)} \right ) \mathbf z_i^{(k)} + \tau \sum_{j\in \mathcal V} s_{ij}^{(k)} \mathbf z_j^{(k)},
\end{equation}
where $s_{ij}^{(k)}$ is given by the trajectory $\mathbf S^{(k)}$ of $\mathbf S(t)$ at the discrete step $k$.
The above numerical iteration coincides with the updating rule of (graph) neural networks from layer $k$ to $k+1$, where the first term in Eqn.~\ref{eqn-diffuse-iter} acts as the residual connection and the second term accommodates the global information from other nodes. 

\noindent\textbf{\emph{Remark.}} The coupling matrix $\mathbf S(t)$ quantifies the pairwise influence at each layer. Since we do not enforce any spatial constraint (from input graphs) on the gradient and divergence operators, the interactions among nodes are fully determined by $\mathbf S(t)$ and there remains much flexibility for its specification.
\begin{itemize}
    \item A basic choice is to fix $\mathbf S(t)$ as an identity matrix which constrains the propagation in Eqn.~\ref{eqn-diffuse-iter} to self-loops and the model degrades to a multi-layer perceptron (MLP) that treats all the nodes independently. With all the nodes isolated from each other, there is no interaction among different locations throughout the diffusion process.

    \item One could also specify $\mathbf S(t)$ as some propagation matrix induced by observed graph structures. In such a case, information flows at each layer are restricted within neighboring nodes in the graph, as is done by common GNNs. This corresponds to a \emph{local diffusion} system where the spatial constraints are determined by the graph.

    \item An ideal case could be to allow $\mathbf S(t)$ to have non-zero values for arbitrary $(i, j)$ and evolve with time, i.e., the node embeddings at each layer can efficiently and adaptively propagate to all the others. In such a case, the information flows involve the interactions among arbitrary location pairs, giving rise to a \emph{non-local diffusion} system~\citep{chasseigne2006asymptotic}.
\end{itemize}  

\subsection{Diffusion with Layer-wise Energy Constraints}\label{sec-model-energy}

As mentioned previously, the crux is how to define a proper coupling function to induce a desired diffusion process that can maximize the information utility and accord with the geometry behind observed data. 
Since we have no prior knowledge for the explicit form of $\mathbf S(t)$ (that can depend on the underlying data geometry), without loss of generality, we consider the diffusivity (and more specifically, the induced coupling matrix $\mathbf S(t)$) as a latent variable for modeling. Furthermore, to enforce a constraint w.r.t.~the presumed quality of node embeddings at an arbitrarily given layer $k$, we resort to an \emph{energy function} $E(\mathbf Z, k)$ that measures the global smoothness of node embeddings, i.e., how variable the quantity is in the diffusive system. In common physical systems, the evolution pursues steady states that minimize some global energy and achieve some equilibrium~\citep{kimmel1997high,bertozzi2012diffuse,luo2017convergence}. Inspired by this phenomenon, we incorporate layer-wise constraints of energy minimization into the diffusion model:
\begin{equation}\label{eqn-diffuse-appx}
    \begin{split}
    &\mathbf z_i^{(k+1)} = \left (1 - \tau \sum_{j\in \mathcal V} s_{ij}^{(k)} \right ) \mathbf z_i^{(k)} + \tau \sum_{j\in \mathcal V} s_{ij}^{(k)} \mathbf z_j^{(k)}, \\
    &\mbox{s. t.}~~\mathbf z_i^{(0)} = \mathbf x_i, \quad E(\mathbf Z^{(k+1)}, k) \leq E(\mathbf Z^{(k)}, k-1), \quad k\geq 1.
    \end{split}
\end{equation}
The above formulation defines a new class of diffusion process on latent manifolds whose dynamics are \emph{implicitly} defined by optimizing an energy function (see Figure~\ref{fig:model} for an illustration). Eqn.~\ref{eqn-diffuse-appx} unifies two schools of thought into a new diffusive system where the updates of node embeddings are driven by both the diffusion dynamics (as an inductive bias) and the energy constraints (as a regularization). The diffusion process describes the \emph{microscopic} behavior of each node's embedding updates through feed-forward evolution, while the energy function provides a \emph{macroscopic} view for quantifying the consistency of the global system. In general, we expect that the final states could yield a low energy, which suggests that the physical system arrives at a steady point wherein the yielded node representations have absorbed enough global information under a certain guiding principle. 
As we will show in the following sections, the updating rules of common GNNs can be cast into the general formulation of Eqn.~\ref{eqn-diffuse-appx} when specifying different forms of the coupling matrix and the energy function (Section~\ref{sec-gnn}), and furthermore, this unified framework can be utilized as a principled guidance for motivating new architecture designs (Section~\ref{sec-trans}).

\section{Graph Neural Networks as Energy-Constrained Diffusion}\label{sec-gnn}

The diffusion system of Eqn.~\ref{eqn-diffuse-appx} is hard to solve since we need to infer $\mathbf S^{(k)}$ at arbitrary layers that are coupled by the energy minimization constraints of $K$ inequalities (where $K$ denotes the number of iterations). Instead of directly resolving this difficult case, in this section, we first consider a simple case where the diffusivity $\mathbf F(t)$ in Eqn.~\ref{eqn-diffuse} is assumed to be fixed w.r.t. time $t$, in which situation Eqn.~\ref{eqn-diffuse} becomes a linear diffusion equation and the induced coupling matrix $\mathbf S(t)$ (resp. $\mathbf S^{(k)}$) remains a constant matrix over time $t$ (resp. layer $k$). Within this setting, we can show that the corresponding diffusion dynamics with energy constraints would yield the updating rules of common GNNs. The proofs for all theoretical results are deferred to Appendix~\ref{appx-proof}.

\subsection{Connection between Static Diffusivity and Energy}\label{sec-gnn-connection}

In the case of static diffusivity, the problem boils down to finding a constant coupling matrix $\mathbf S$ that gives rise to the diffusion dynamics satisfying the energy constraint at each step. We define the Laplacian of $\mathbf S$ as $\mathbf \Delta = \tilde{\mathbf D} - \mathbf S$, where $\tilde{\mathbf D}$ is the diagonal degree matrix of $\mathbf S$, and we next show that for a typical quadratic energy form, there exists $\mathbf S$ whose yielded diffusion process is the solution for Eqn.~\ref{eqn-diffuse-appx} under certain mild conditions.

\begin{theorem}\label{thm-gcn}
    Assume that the diffusivity is fixed w.r.t. time (a.k.a. layers), i.e., $\mathbf S^{(k)} = \mathbf S = [s_{ij}]_{i, j\in \mathcal V}$. Then for any step size $0<\tau\leq \frac{1}{\lambda_1}$, where $\lambda_1$ is the largest singular value of $\mathbf \Delta$, the feed-forward iteration of Eqn.~\ref{eqn-diffuse-iter} globally descends the energy of the quadratic form
    \begin{equation}\label{eqn-energy-gcn}
        E(\mathbf Z, k) = \|\mathbf Z - \mathbf Z^{(k)}\|^2_{\mathcal F} + \lambda(\tau) \sum_{i, j} s_{ij}\cdot \|\mathbf z_i - \mathbf z_j\|_2^2,
    \end{equation}
    where $\lambda$ is a weight dependent on the step size $\tau$, formally $E(\mathbf Z^{(k+1)}, k) \leq E(\mathbf Z^{(k)}, k-1)$, with equality iff~ $\mathbf Z^{(k)}$ is a stationary point of $E(\mathbf Z, k)$.
\end{theorem}

The energy function Eqn.~\ref{eqn-energy-gcn} integrates two-fold effects. The first term enforces the \emph{local} smoothness that penalizes the large gap between the next-layer embedding and the one of the current layer. The second term, which can be essentially seen as the spatially discretized counterpart of Eqn.~\ref{eqn-dirichlet-energy} with the instantiation of $\Omega$ as a graph~\citep{zhou2005regularization}, enforces the \emph{global} smoothness that penalizes the difference between the embeddings of different node pairs at the next layer. Thereby, Theorem~\ref{thm-gcn} suggests that the diffusion process with static diffusivity inherently minimizes a convex energy that facilitates the consistency of node embeddings throughout the feed-forward updating. Moreover, when instantiating the coupling matrix $\mathbf S$ as certain particular choices, we can connect the energy-constrained diffusion dynamics and the message passing rules of common GNNs, as illustrated by the following examples.

\begin{example}\label{exa-gcn}
    If we assume $\mathbf S = \mathbf D^{-\frac{1}{2}}\mathbf A \mathbf D^{-\frac{1}{2}}$, Eqn.~\ref{eqn-diffuse-iter} would become $\mathbf z_i^{(k+1)} = (1 - \tau) \mathbf z_i^{(k)} + \tau \sum_{j\in \mathcal N(i)} \frac{1}{\sqrt{d_id_j}} \mathbf z_j^{(k)} $, i.e., one-layer updating of graph convolution networks~\citep{GCN-vallina} with residual connection.
\end{example}

\begin{example}
    If we assume $\mathbf S = \mathbf A + \mathbf I$, Eqn.~\ref{eqn-diffuse-iter} can be equivalently written as $\mathbf z_i^{(k+1)} = (1 + \tau) \mathbf z_i^{(k)} + \tau \sum_{j\in \mathcal N(i)} \mathbf z_j^{(k)}$, i.e., one-layer updating of graph isomorphism network~\citep{gin} up to a re-scaling factor.
\end{example}

Apart from these two instantiations, there also exist many other choices for $\mathbf S$ whose corresponding diffusion iterations coincide with existing GNNs' message passing. The above results indicate that the message passing layers can be seen as trajectories of the diffusion dynamics minimizing the associated energy. While the above analysis focuses on the discrete iterations induced by the diffusion, we can further extend the results to the continuous PDE dynamics and show that the updates of node embeddings within infinitesimal time equals to the negative gradient of the energy.

\begin{corollary}\label{corollary-gcn-cont}
    The diffusion dynamics of Eqn.~\ref{eqn-diffuse2} with static diffusivity that induces a constant coupling matrix $\mathbf S(t) = \mathbf S$ is a gradient flow $\frac{\partial \mathbf z_i(t)}{\partial t} = - \frac{1}{2}\nabla_{\mathbf z_i} E(\mathbf Z, t)$, where $E(\mathbf Z, t) = \|\mathbf Z - \mathbf Z(t)\|^2_{\mathcal F} + \lambda \sum_{i, j} s_{ij} \|\mathbf z_i - \mathbf z_j\|_2^2$.
\end{corollary}

Another property we can show based on Theorem~\ref{thm-gcn} is the global convergence of the diffusion-induced iterations w.r.t. energy minimization, which reveals the final state of the iterations analogous to certain equilibrium of the system.

\begin{corollary}\label{cora-conv}
Under the same conditions as Theorem~\ref{thm-gcn}, the numerical iteration of Eqn.~\ref{eqn-diffuse-iter} converges to the global optimum of $E(\mathbf Z, k)$.
\end{corollary}

In spite of the convergence property, the diffusion iterations will eventually arrive at the global optimum, where the energy is minimized to zero, with infinite steps of iterations. In such a case, all the node embeddings converge to a single point in the latent space, corresponding to the well-known over-smoothing phenomenon when stacking deep GNN layers. However, in practice, we do not require deep propagation layers for desired performance or the need to minimize the energy to the global minimum, the over-smoothing issue can be alleviated in this regard. Yet, as we will discuss in the next subsection, the risk of over-smoothing can be avoided with slight modification of the diffusion equation. 

\subsection{Further Discussions and Extensions}\label{sec-gnn-dis}

In the previous subsection, we pinpoint the underlying energy descended by the feed-forward diffusion dynamics. Some follow-up questions still remain. First, it is unclear how much quantity each iteration step contributes to the energy descent, which is linked with the convergence rate of the iterations based on the result of Corollary~\ref{cora-conv}. Second, the over-smoothing issue caused by the global convergence adds to the lingering concern on the robustness of the diffusion system.

To answer the above questions and supplement further discussions based on our theory, we next derive upper and lower bounds of the energy $E(\mathbf Z^{(k)}, k-1)$ at each step, shedding light on the convergence speed of the iterations, and furthermore, discuss how to amend the diffusion equation with a source term to resolve the over-smoothing issue.

\subsubsection{Upper and Lower Bounds for the Layer-wise Energy}\label{sec-gnn-dis-bound}

The diffusion iterations in Eqn.~\ref{eqn-diffuse-iter} are determined by two factors: the coupling matrix $\mathbf S^{(k)}$ (which is assumed to be a constant matrix $\mathbf S$ in our case) and the step size $\tau$. The former determines the rate of information flows across different nodes, while the latter controls the forward speed of one-step iteration. We can show that the minimization of the global energy can be further characterized by the upper and lower bounds at each step, which reveals the descending speed by each diffusion iteration, and the bounds are associated with the coupling matrix $\mathbf S$ and the step size $\tau$.

\begin{proposition}\label{prop-bound}
    On the same conditions of Theorem~\ref{thm-gcn}, for arbitrarily given $k$, the energy yielded by the diffusion iteration Eqn.~\ref{eqn-diffuse-iter} with $\mathbf S^{(k)} = \mathbf S$ is bounded by: 
    \begin{equation}
        (1 - \tau\lambda_1)^2E(\mathbf Z^{(k)}, k-1) \leq E(\mathbf Z^{(k+1)}, k) \leq (1 - \tau\lambda_2)^2 E(\mathbf Z^{(k)}, k-1),
    \end{equation}
    where $\lambda_2$ is the smallest singular value of $\mathbf \Delta$.
\end{proposition}

This proposition indicates that the energy yielded by the next layer lies in a certain interval dependent on the energy of the previous layer. It further suggests that the convergence rate of energy minimization is $(1 - \tau \lambda_2)^{2}$ depending on the smallest eigenvalue of the Laplacian of $\mathbf S$ and the step size $\tau$. 

\subsubsection{Global Convergence with A Source Term}\label{sec-gnn-dis-source}

Another concerning issue of the diffusion iteration is its convergence to the global optimum of the energy Eqn.~\ref{eqn-energy-gcn} that corresponds with a degenerate solution where all the node embeddings degrade to a single point in the latent space. Critically though, such a potential risk can be overcome by augmenting the diffusion equation Eqn.~\ref{eqn-diffuse2} with a source term
\begin{equation}\label{eqn-diffuse-source}
    \frac{\partial \mathbf z_i(t)}{\partial t} = \sum_{j\in \mathcal V} s_{ij}(t) (\mathbf z_j(t) - \mathbf z_i(t)) + \beta \mathbf h_i,
\end{equation} 
where $\mathbf h_i$ with the weight $\beta$ can be considered as some extra input signals from external sources to each point within the system. Correspondingly, the induced numerical iteration would become the counterpart of Eqn.~\ref{eqn-diffuse-iter} augmented with an additional term $\tau\beta \mathbf h_i$ for node $i$ at each step. By extending the analysis of Theorem~\ref{thm-gcn}, we can obtain the energy function descended by the new diffusion system (where we assume $\mathbf H = [\mathbf h_i]_{i\in \mathcal V}$).

\begin{proposition}\label{prop-convergence}
    For the diffusion dynamics Eqn.~\ref{eqn-diffuse-source} with a constant coupling matrix $\mathbf S^{(k)} = \mathbf S = [s_{ij}]_{i, j\in \mathcal V}$ and step size $0 < \tau \leq \frac{1}{\lambda_1}$, the induced numerical iteration from $\mathbf z_i^{(k)}$ to $\mathbf z_i^{(k+1)}$ satisfies the energy constraint $E(\mathbf Z^{(k+1)}, k) \leq E(\mathbf Z^{(k)}, k-1)$ with the energy of the form
    \begin{equation}\label{eqn-energy-appnp}
        E(\mathbf Z, k) = \|\mathbf Z - (\mathbf Z^{(k)} + \eta(\beta, \tau)\mathbf H)\|^2_{\mathcal F} + \lambda(\tau) \sum_{i, j} s_{ij}\cdot \|\mathbf z_i - \mathbf z_j\|_2^2,
    \end{equation}
    where $\eta$ is a coefficient dependent on the step size $\tau$ and the weight for source term $\beta$.
\end{proposition}

One can derive the global optimum of Eqn.~\ref{eqn-energy-appnp}, i.e., $\mathbf Z^* = \frac{\eta}{\lambda}(\tilde{\mathbf D} - \mathbf S)^{-1} \mathbf H$, by letting $\frac{\partial E(\mathbf Z, k)}{\partial \mathbf Z} = 0$. This suggests that as time goes to the infinity, the diffusion process would globally converge to, critically, a non-degenerate fixed state where the final representations of different nodes preserve enough diversity given proper settings of $\{\mathbf h_i\}_{i\in \mathcal V}$. For example, one can simply set $\mathbf h_i = \mathbf z^{(0)}$ with the initial embedding of each node to reinforce the information of the centered node, in which case the diffusion iteration intersects with the message passing design of some GNN architectures that are invulnerable to over-smoothing as the layers go deep, as illustrated by the example below.

\begin{example}
    For $\mathbf S^{(k)} = \mathbf D^{-\frac{1}{2}}\mathbf A \mathbf D^{-\frac{1}{2}}$ and $\mathbf H = \mathbf Z^{(0)}$, the feed-forward iteration induced by Eqn.~\ref{eqn-diffuse-source} yields the updating rule $\mathbf z_i^{(k+1)} = (1 - \tau) \mathbf z_i^{(k)} + \tau \sum_{j\in \mathcal N(i)} \frac{1}{\sqrt{d_id_j}} \mathbf z_j^{(k)} + \tau\beta \mathbf z^{(0)}$, the form of which is adopted by APPNP~\citep{appnp} and loosely adopted by GCNII~\citep{gcnii}.
\end{example}

\section{Transformer Backbones Induced by Energy-Constrained Diffusion}\label{sec-trans}

In the previous section, we assume the diffusivity to be dependent on specific locations yet stay unchanged over time (a.k.a. layers). While we have shown that the diffusion process in such a case implicitly minimizes a principled energy which regularizes the internal consistency of the produced embeddings at each step, the static diffusivity may limit the flexibility of the diffusion system, in particular for accommodating the adaptive pairwise influence among data points. In real-world complex physical systems, the diffusivity often goes through both spatial and temporal variations. For example, in cells, the diffusivity of ions and molecules can change due to fluctuations in temperature, local concentration gradients, and cellular activity~\citep{heitjans2006diffusion}; besides, in fluid dynamics, turbulent flows can exhibit varying diffusivity due to the chaotic nature of the flow~\citep{pope2000turbulent,csanady1973turbulent}.

We next investigate a more expressive model that is comprised of time-dependent diffusivity, in which case Eqn.~\ref{eqn-diffuse} becomes a non-linear diffusion equation and the induced coupling matrix $\mathbf S^{(k)}$ in Eqn.~\ref{eqn-diffuse2} can flexibly change at different layers. In such a case, the information among arbitrary node pairs can flow at an adaptive rate dependent on specific locations and time. We will show that in such a situation, there also exists an associated global energy function that are implicitly descended by the diffusion dynamics. Furthermore, we will link the time-dependent coupling matrix with the attention mechanism that is often inserted in-between two neural layers to model the pairwise influence based on the embeddings computed at the current layer.

\subsection{Connection between Time-Dependent Diffusivity and Energy}\label{sec-trans-connection}

When the diffusivity can vary over time, Eqn.~\ref{eqn-diffuse-appx} becomes hard to solve since we need to infer the value for a series of coupled $\mathbf S^{(k)}$'s that need to satisfy $K$ inequalities by the energy minimization constraints. Instead of solving Eqn.~\ref{eqn-diffuse-appx} directly, we can notice a natural corollary based on Theorem~\ref{thm-gcn} that the $k$-th step iteration of Eqn.~\ref{eqn-diffuse-iter} contributes to a descent step on the local energy at the current layer: 
\begin{equation}\label{eqn-energy-layer}
    E(\mathbf Z, k; \mathbf S^{(k)}) = \|\mathbf Z - \mathbf Z^{(k)}\|^2_{\mathcal F} + \lambda(\tau) \sum_{i, j} s_{ij}^{(k)}\cdot \|\mathbf z_i - \mathbf z_j\|_2^2,
\end{equation}
where $s_{ij}^{(k)}$ is the $(i, j)$-th entry of the coupling matrix $\mathbf S^{(k)}$ at the $k$-th step. We can thereby extend the analysis and results in Section~\ref{sec-gnn} for interpreting the behavior of one-step diffusion iteration from the $k$-th layer to the $(k+1)$-th. However, since the energy function Eqn.~\ref{eqn-energy-layer} depends on the coefficient $s_{ij}^{(k)}$ at the $k$-th layer that varies throughout the diffusion process, it is still unclear the global behavior of diffusion dynamics, particularly if there is a global energy (shared across all layers) that is minimized by the whole trajectory.

In the following, we aim to unlock the black box of the diffusion system with time-dependent diffusivity and reveal a global energy associated with the diffusion, which boils down to finding closed-form solutions for $\mathbf S^{(k)}$ that give rise to a diffusion process satisfying Eqn.~\ref{eqn-diffuse-appx}. To achieve this goal, we first prove a preliminary result that suggests a surrogate energy that serves as a strict upper bound of a non-convex regularized energy.

\begin{proposition}\label{prop-energy}
    For the regularized energy function of the form
    \begin{equation}\label{eqn-energy}
        E(\mathbf Z, k; \delta) = \|\mathbf Z - \mathbf Z^{(k)}\|_{\mathcal F}^2 + \lambda\sum_{i,j} \delta(\|\mathbf z_i - \mathbf z_j\|_2^2),
    \end{equation}
    where $\delta:\mathbb R^+ \rightarrow \mathbb R$ is defined as a function that is \emph{non-decreasing} and \emph{concave} on a particular interval of our interest, we have its upper bound $E(\mathbf Z, k; \delta) \leq \tilde{E}(\mathbf Z, k; \mathbf \Omega^{(k)}, \tilde \delta):$
    \begin{equation}\label{eqn-prop1-bound}
        \tilde E(\mathbf Z, k; \mathbf\Omega^{(k)}, \tilde \delta) = \|\mathbf Z - \mathbf Z^{(k)}\|_{\mathcal F}^2 + \lambda \left[\sum_{i,j} \omega_{ij}^{(k)} \|\mathbf z_i - \mathbf z_j\|_2^2 - \tilde \delta(\omega_{ij}^{(k)})\right ],
    \end{equation}
    where $\mathbf \Omega^{(k)} = [\omega_{ij}^{(k)}]_{i, j\in \mathcal V}$ and $\tilde \delta$ denotes the concave conjugate of $\delta$,
    and the equality holds if and only if the variaitonal parameter $\omega_{ij}^{(k)}$ satisfies $\omega_{ij}^{(k)} = \left. \frac{\partial \delta(z^2)}{\partial z^2} \right |_{z = \|\mathbf z_i - \mathbf z_j\|_2}$.
\end{proposition}

In light of the proposition, we notice that if treating the coupling matrix $\mathbf S^{(k)}$ as the variaitonal parameters $\mathbf\Omega^{(k)}$, then one-step iteration of Eqn.~\ref{eqn-diffuse-iter} contributes to minimizing the upper bound of the non-convex energy Eqn.~\ref{eqn-energy}. Pushing further, if the coupling matrix at the $k$-th layer is given by $\mathbf S^{(k)}=\mathbf \Omega^{(k)} =[\omega_{ij}^{(k)}]_{i, j\in \mathcal V}$ where $\omega_{ij}^{(k)}=\left. \frac{\partial \delta(z^2)}{\partial z^2} \right |_{z = \|\mathbf z_i^{(k)} - \mathbf z_j^{(k)}\|_2}$, then Eqn.~\ref{eqn-diffuse-iter} serves to minimize the global energy Eqn.~\ref{eqn-energy} that equals to the quantity of the surrogate Eqn.~\ref{eqn-prop1-bound} with the variational parameters $\omega_{ij}^{(k)}$ fixed as $s^{(k)}_{ij}$. One concern, however, is the convergence of the gradient descent on Eqn.~\ref{eqn-prop1-bound} with the iteration of Eqn.~\ref{eqn-diffuse-iter}, which, as shown by previous analysis, depends on $\mathbf S^{(k)}$ and the step size $\tau$. Since $\mathbf S^{(k)}$ can be different at each layer, we need the step size smaller than the inverse of the largest eigenvalue among all the $\mathbf S^{(k)}$'s to guarantee the convergence. To make the result more concise, we consider $\mathbf S^{(k)}$ to be row-normalized, which further enables us to link the layer-dependent coupling matrix with the attention weight (more discussions are in Section~\ref{sec-trans-attn}). We formulate the main result as the following theorem that implies a closed-form solution for $\mathbf S^{(k)}$ for the diffusion system Eqn.~\ref{eqn-diffuse-appx} and further reveals the underlying energy-descending property of the diffusion process with time-dependent diffusivity.

\begin{theorem}\label{thm-main}
    For any regularized energy function $E(\mathbf Z, k; \delta)$ of the generic form Eqn.~\ref{eqn-energy}, there exists step size $0<\tau\leq 1$ such that the diffusion process of Eqn.~\ref{eqn-diffuse-iter} with the coupling matrix $\mathbf S^{(k)} = [s_{ij}^{(k)}]_{i, j\in \mathcal V}$ at the $k$-th step given by
    \begin{equation}\label{eqn-optimal-diffuse}
        \hat s_{ij}^{(k)} = \frac{\omega_{ij}^{(k)}}{\sum_{l=1}^N \omega_{il}^{(k)}}, \quad \omega_{ij}^{(k)} = \left. \frac{\partial \delta(z^2)}{\partial z^2} \right |_{z^2 = \|\mathbf z_i^{(k)} - \mathbf z_j^{(k)}\|_2^2},
    \end{equation}
    yields a descent step on the energy, i.e., $E(\mathbf Z^{(k+1)}, k; \delta) \leq E(\mathbf Z^{(k)}, k-1; \delta)$ for any $k\geq 1$. 
\end{theorem}

Now we obtain the global energy function $E(\mathbf Z, k; \delta)$ minimized by the diffusion dynamics with the layer-dependent coupling matrix $\mathbf S^{(k)}$. In the definition of $E(\mathbf Z, k; \delta)$, i.e., Eqn.~\ref{eqn-energy}, the concave and non-decreasing function $\delta$ can be seen as a penalty function designed to promote robustness against node embedding differences across spurious pairs. Or stated differently, since $\delta$ is concave, large errors across spurious node pairs will not accrue and dominate the objective \citep{RWLS-icml21}. In this way, $\delta$ can be loosely viewed as introducing an implicit form of latent structure inference, with Eqn.~\ref{eqn-energy} serving as a robust non-convex energy for learning local and global consistency in the spirit of \cite{globallocal-2003}.

Pushing further, similar to the case studied in Section~\ref{sec-gnn}, we can extend the result to a continuous PDE diffusion system where the evolution of node embeddings described by the diffusion equation is implicitly given by the negative gradient direction of the regularized energy, as formulated by the following corollary.

\begin{corollary}\label{coro-main}
    The non-linear diffusion equation with a time-dependent coupling matrix $\frac{\partial \mathbf z_i(t)}{\partial t} = \sum_{j\in \mathcal V} s_{ij}(t) (\mathbf z_j(t) - \mathbf z_i(t))$ is a gradient flows $\frac{\partial \mathbf z_i(t)}{\partial t} = - \frac{1}{2}\nabla_{\mathbf z_i} E(\mathbf Z, t)$, where 
    \begin{equation}\label{eqn-energy-con}
        E(\mathbf Z, t; \delta) = \|\mathbf Z - \mathbf Z(t)\|^2_{\mathcal F} + \lambda(\tau) \sum_{i, j} \delta(\|\mathbf z_i - \mathbf z_j\|_2^2).
    \end{equation}
\end{corollary}


We next shed some insights on the implications of Theorem~\ref{thm-main} (as well as Corollary~\ref{coro-main}). According to Theorem~\ref{thm-main}, the inherent connection between the diffusion process and the associated energy lies in the correspondence between the penalty function $\delta$ and the coupling matrix $\mathbf S^{(k)}$. The latter bridges the two perspectives into a unified framework. Specifically, we can re-state the conclusion of Theorem~\ref{thm-main} via two statements below.

\noindent \textbf{Statement 1.} For a regularized energy $E(\mathbf Z, k; \delta)$ with a given penalty function $\delta$, there exists a diffusion process with the coupling matrix $\mathbf S^{(k)} = [s_{ij}^{(k)}]_{i, j\in \mathcal V}$ given by Eqn.~\ref{eqn-optimal-diffuse} that satisfies the energy constraints, i.e., the closed-form solution for Eqn.~\ref{eqn-diffuse-appx}.

\noindent \textbf{Statement 2.} For a given diffusion process with the coupling matrix $\mathbf S^{(k)}$ instantiated as Eqn.~\ref{eqn-optimal-diffuse}, there exists an underlying global energy $E(\mathbf Z, k; \delta)$ descended by the whole dynamics, where $\delta$ satisfies the condition of Eqn.~\ref{eqn-optimal-diffuse}. 

We next leverage these two statements as principled guidance for motivating new message-passing-based model architectures and interpreting existing attention networks, via aligning $\mathbf S^{(k)}$ with the layer-dependent attention weights inferred by modern Transformer-like models.

\subsection{Principled Attention Layers Derived From Diffusion Process}\label{sec-trans-attn}

Theorem~\ref{thm-main} suggests the existence for the optimal $\mathbf S^{(k)}$ at each time step for the diffusion process satisfying the energy minimization constraint (i.e., Eqn.~\ref{eqn-diffuse-appx}). The result enables us to unfold the implicit process of Eqn.~\ref{eqn-diffuse-appx} and compute $\mathbf S^{(k)}$ in a feed-forward way from the initial embeddings $\mathbf Z^{(0)}$. Specifically, the condition of Eqn.~\ref{eqn-optimal-diffuse} implies that the optimal $\mathbf S^{(k)}$ in the form of a function over the $l_2$ distance between node embeddings, i.e., $z = \|\mathbf z_i^{(k)} - \mathbf z_j^{(k)}\|_2$. We thereby introduce a pairwise distance function $f(z^2)$ and define a new family of neural model architectures with layer-wise updating rules specified by:
{\small
\begin{center}
\fcolorbox{black}{gray!10}{\parbox{0.97\linewidth}{
\begin{equation}\label{eqn-model}
    \begin{split}
        & \mbox{Diffusivity Inference:} \quad \hat s_{ij}^{(k)} = \frac{f(\|\mathbf z_i^{(k)} - \mathbf z_j^{(k)}\|_2^2)}{\sum_{l\in \mathcal V} f(\|\mathbf z_i^{(k)} - \mathbf z_l^{(k)}\|_2^2)}, \leq i, j \in \mathcal V,\\
        & \mbox{State Updating:} \quad \mathbf z_i^{(k+1)} = \underbrace{\left (1 - \tau \sum_{j\in \mathcal V} \hat s_{ij}^{(k)} \right ) \mathbf z_i^{(k)} }_{\mbox{state conservation}} + \underbrace{\tau \sum_{j\in \mathcal V} \hat s_{ij}^{(k)} \mathbf z_j^{(k)} }_{\mbox{state propagation}}, \quad i \in \mathcal V.
    \end{split}
\end{equation}
}
}
\end{center} }
The model layer defined above consists of two consecutive operations where the \emph{diffusivity inference} estimates the pairwise attention using the current node embeddings and the \emph{state updating} computes the next-layer embeddings with attention-based propagation. According to the results of Proposition~\ref{prop-energy} and Theorem~\ref{thm-main}, Eqn.~\ref{eqn-model} can be seen as an execution of a minimization-minimization algorithm towards optimizing the energy target Eqn.~\ref{eqn-energy}: 1) with fixed $\mathbf Z^{(k)}$, the \emph{diffusivity inference} returns the optimal variational parameters $\hat{\mathbf S}^{(k)}=[\hat s_{ij}^{(k)}]_{i, j\in \mathcal V}$ that decrease the upper bound, i.e., surrogate energy Eqn.~\ref{eqn-prop1-bound}, to approximate the target Eqn.~\ref{eqn-energy}; 2) the \emph{state updating} proceeds to descend the surrogate energy with the variational parameters $\mathbf \Omega^{(k)}=\hat{\mathbf S}^{(k)}$ fixed, which equivalently minimizes the energy target Eqn.~\ref{eqn-energy}. 

\textbf{\emph{Remark.}} Since $f$ is the first-order derivative of $\delta$, the choice of function $f$ in above formulation is not arbitrary, but needs to be a non-negative and decreasing function of $z^2$, so that the associated $\delta$ in Eqn.~\ref{eqn-energy} is guaranteed to be non-decreasing and concave w.r.t. $z^2$ (i.e., the condition of Proposition~\ref{prop-energy}). Critically though, there remains much room for us to properly design the specific $f$, 
so as to provide adequate capacity and scalability. Also, in our model presented by Eqn.~\ref{eqn-model} we only have one hyper-parameter $\tau$ in practice, noting that the weight $\lambda$ in the regularized energy is determined through $\tau$ by Theorem \ref{thm-main}, which reduces the cost of hyper-parameter searching.

\subsubsection{Attention Designs Inspired by Time-Dependent Diffusivity}\label{sec-trans-attn-difformer}

We next go into model instantiations based on the above theory, and introduce two specified $f$'s as practical versions of our model. To begin with, because $\|\mathbf z_i - \mathbf z_j\|_2^2 = \|\mathbf z_i\|_2^2 + \|\mathbf z_j\|_2^2 - 2\mathbf z_i^\top \mathbf z_j$, we can convert $f(\|\mathbf z_i - \mathbf z_j\|_2^2)$ into the form $g(\mathbf z_i^\top \mathbf z_j)$ on the condition that $\|\mathbf z_i\|_2$ remains constant. Notice that this condition is relatively mild since the normalization is often used to rescale the node embeddings before attention in practice. In particular, we use the L2 normalization to constrain $\mathbf z_i$ to have unit norm.

\textbf{Simple Diffusivity Attention.} A straightforward design is to adopt the linear function $g(x) = 1+x$ that gives rise to the dot-product attention:
\begin{equation}\label{eqn-S1}
    f(\|\mathbf z_i^{(k)} - \mathbf z_j^{(k)}\|_2^2) = 1 + (\mathbf z_i^{(k)})^\top 
    \mathbf z_j^{(k)}.
\end{equation}
By treating $z = \|\mathbf z_i^{(k)} - \mathbf z_j^{(k)}\|_2$, Eqn.~\ref{eqn-S1} can be written as $f(z^2) = 2 - \frac{1}{2}z^2$, which yields a non-negative result and is decreasing on the interval $[0, 2]$ in which $z^2$ lies. By simple calculation, we can obtain the corresponding penalty function $\delta(z^2) = 2z^2 - \frac{1}{4}z^4$ which is non-decreasing and concave w.r.t. $z^2$ within $[0, 2]$. One scalability concern for the model Eqn.~\ref{eqn-model} arises because of the need to compute all-pair attention scores and propagation for each individual node, inducing at least $\mathcal O(|\mathcal V|^2)$ complexity. In such a case, the model can be difficult in scaling to large-scale systems where $|\mathcal V|$ is prohibitively large. Remarkably, the simple diffusivity model allows a significant acceleration by noting that the state propagation term of Eqn.~\ref{eqn-model} can be re-arranged via
\begin{equation}
     \sum_{j\in \mathcal V} s_{ij}^{(k)} \mathbf z_j^{(k)} = \sum_{j\in \mathcal V} \frac{1 + (\mathbf z_i^{(k)})^\top \mathbf z_j^{(k)}}{\sum_{l\in \mathcal V}\left (1 + (\mathbf z_i^{(k)})^\top \mathbf z_l^{(k)} \right ) } \mathbf z_j^{(k)} = \frac{\sum_{j\in \mathcal V}\mathbf z_j^{(k)} + \left (\sum_{j\in \mathcal V} \mathbf z_j^{(k)}\cdot  (\mathbf z_j^{(k)})^\top \right ) \cdot \mathbf z_i^{(k)} }{|\mathcal V| + (\mathbf z_i^{(k)})^\top \sum_{l\in \mathcal V} \mathbf z_l^{(k)}}.
\end{equation}
The two summation terms above can be computed once and shared to every node $i$, reducing the complexity in each iteration to $\mathcal O(|\mathcal V|)$ (see more details in Appendix~\ref{appx-alg} for how we achieve linear complexity w.r.t. $|\mathcal V|$ in the matrix form for model implementation). We refer to this version of our model implementation as DIFFormer-s. 

\textbf{Advanced Diffusivity Attention.} The simple model facilitates efficiency and scalability, yet may sacrifice the capacity for learning complex latent geometry. We thus propose an advanced version with non-linearity incorporated $g(x) = \frac{1}{1+\exp(-x)}$: 
\begin{equation}\label{eqn-S2}
    f(\|\mathbf z_i^{(k)} - \mathbf z_j^{(k)}\|_2^2) = \frac{1}{1+\exp{\left(- (\mathbf z_i^{(k)} )^\top 
    (\mathbf z_j^{(k)} ) \right ) }}.
\end{equation}
In such a case, the corresponding $f$ and $\delta$ can be written as $f(z^2) = \frac{1}{1+e^{z^2/2-1}}$ and $\delta(z^2) = z^2 - 2 \log (e^{z^2/2-1} + 1)$, respectively, where the latter satisfies the non-decreasing and concavity properties w.r.t. $z^2$. We dub this model version as DIFFormer-a. Appendix~\ref{appx-inst} further compares the two models (i.e., different $f$'s and $\delta$'s) through synthetic results. Real-world empirical comparisons are in Section~\ref{sec-exp}.



\subsubsection{Connection with Existing Attention Mechanisms}\label{sec-trans-attn-existing}

Another interesting perspective is to leverage the energy-constrained diffusion framework to interpret the existing attention networks. From Eqn.~\ref{eqn-model} one can naturally connect $\hat s_{ij}^{(k)}$ with the attention score and consider $f$ as a similarity measure. For example, in the original Transformers~\citep{transformer}, $f$ is instantiated as a dot-then-exponential operator:
\begin{equation}\label{eqn-trans-attn}
    f(\|\mathbf z_i^{(k)} - \mathbf z_j^{(k)}\|_2^2) = \exp\left (\frac{(\mathbf z_i^{(k)})^\top \mathbf z_j^{(k)}}{\sqrt{d}} \right ), \quad \hat{s}_{ij}^{(k)} = \frac{\exp((\mathbf z_i^{(k)})^\top \mathbf z_j^{(k)}) }{\sum_{l\in \mathcal V } \exp((\mathbf z_i^{(k)})^\top \mathbf z_l^{(k)})}.
\end{equation}
In such a case, $f(z^2) = e^{1 - \frac{1}{2}z^2} e^{\frac{1}{\sqrt{d}}}$, which is non-negative and decreasing w.r.t. $z^2$. Therefore, there exists a corresponding $\delta$ that satisfies the condition of Theorem~\ref{thm-main} and gives rise to an associated regularized energy globally minimized by a sequence of Softmax attention layers.

Apart from the dot-then-exponential $f$ used by \cite{transformer}, there also exist a series of other attention networks explored by its follow-ups enriching the family of Transformers. Among these different models, we can consider the un-normalized attention as the output of a general pairwise similarity function $c(\mathbf z_i^{(k)}, \mathbf z_j^{(k)})$. Based on our analysis, once $c$ is non-negative and decreasing w.r.t. $\|\mathbf z_i^{(k)} - \mathbf z_j^{(k)}\|_2^2$ (where the latter condition is equivalent to increasing w.r.t. $(\mathbf z_i^{(k)})^\top \mathbf z_j^{(k)}$ on condition that $\mathbf z_i^{(k)}$ has unit norm), then the model stacking multiple attention layers can be cast into the forward iterations of the energy-constrained diffusion.

\begin{example}
    For $s^{(k)}_{ij} = \frac{c(\mathbf z_i^{(k)}, \mathbf z_j^{(k)})}{\sum_{l\in \mathcal V} c(\mathbf z_i^{(k)}, \mathbf z_l^{(k)})}$, where $c$ is non-negative and decreasing w.r.t. $\|\mathbf z_i^{(k)} - \mathbf z_j^{(k)}\|_2^2$, the feed-forward diffusion iteration $\mathbf z_i^{(k+1)} = (1 - \tau) \mathbf z_i^{(k)} + \tau \sum_{j\in \mathcal V} s^{(k)}_{ij} \mathbf z_j^{(k)}$ yields the updating rule widely adopted by the family of attention-based models.
\end{example}

Another model class that combines the spirits of GNNs and attention mechanism resorts to constraining the attention computation within the neighboring nodes. The typical example is the Graph Attention Networks (GAT)~\citep{GAT}. While the original GAT instantiates the similarity function as $c(\mathbf z_i^{(k)}, \mathbf z_j^{(k)}) = \mbox{LeakyReLU}(\mathbf a^\top [\mathbf W^{(k)}\mathbf z_i^{(k)} \| \mathbf W^{(k)}\mathbf z_j^{(k)}])$, which does not guarantee larger scores for inputs that are closer in the latent space, we can still consider an extended version of GAT via generalizing the similarity function.
\begin{example}
    For $s^{(k)}_{ij} = \frac{c(\mathbf z_i^{(k)}, \mathbf z_j^{(k)})}{\sum_{l\in \mathcal N(i)} c(\mathbf z_i^{(k)}, \mathbf z_l^{(k)})}$, where $c$ is non-negative and decreasing w.r.t. $\|\mathbf z_i^{(k)} - \mathbf z_j^{(k)}\|_2^2$, the feed-forward diffusion iteration $\mathbf z_i^{(k+1)} = (1 - \tau) \mathbf z_i^{(k)} + \tau \sum_{j\in \mathcal N(i)} s^{(k)}_{ij} \mathbf z_j^{(k)}$ serves as an extension of the updating rule of GAT~\citep{GAT} with residual connection.
\end{example}

\subsection{Model Extensions and Further Discussions}\label{sec-trans-dis}

The analysis so far focuses on the message passing rules of different models, particularly the propagation of node embeddings in-between layers. In consideration of practical implementation, apart from the feature propagation, common neural networks involve feature transformation (e.g., the trainable weight matrices involved in the feed-forward layer) and non-linear activation to endow the model with capacity for expressing complex functions. Moreover, in the context of learning on graphs, there often exist observed graphs that can be informative for improving the quality of node representations. We next probe into how our theory can be generalized to practical neural networks and how to incorporate the graph inductive bias into the global attentions. Besides, similar to the diffusion with static diffusivity discussed in Section~\ref{sec-gnn}, the diffusion model presented in this section is susceptible to the over-smoothing problem as well, and we will discuss how to resolve this issue in the case of time-dependent diffusivity.

\subsubsection{Incorporating Feature Transformations in-between Layers}\label{sec-trans-dis-w}

Common neural networks utilize feature transformation and non-linear activation in-between two (propagation) layers. More specifically, after inserting the layer-wise transformation, the updating rule of the $k$-th step can be written as:
\begin{equation}\label{eqn-diff-iter-w}
    \mathbf z_i^{(k+1)} = \sigma \left [ \left (1 - \tau \sum_{j\in \mathcal V} s_{ij}^{(k)} \right ) h^{(k)}(\mathbf z_i^{(k)}) + \tau \sum_{j\in \mathcal V} s_{ij}^{(k)} h^{(k)}(\mathbf z_j^{(k)}) \right ],
\end{equation}
where $h^{(k)}$ is the feature transformation of the $k$-th layer (e.g., a fully-connected layer) and $\sigma$ denotes the non-linear activation (e.g., ReLU). We can generalize our previous result to show that the above diffusion iteration decreases a layer-specific energy.

\begin{proposition}\label{prop-feattrans}
    For any non-decreasing element-wise activation function $\sigma$ and step size $0<\tau\leq 1$, there exists a penalty function $\psi_{\sigma}$ such that the iteration of Eqn.~\ref{eqn-diff-iter-w} with $\mathbf S^{(k)}$ given by Eqn.~\ref{eqn-optimal-diffuse} contributes to a descent step from $\mathbf Z^{(k)}$ to $\mathbf Z^{(k+1)}$ on the energy
    \begin{equation}\label{eqn-energy-w}
        E(\mathbf Z, k;\delta, h^{(k)}) = \|\mathbf Z - h^{(k)}(\mathbf Z^{(k)})\|_2^2 + \sum_{i,j} \delta(\|\mathbf z_i - \mathbf z_j\|_2^2) + \sum_{i} \psi_{\sigma}(\mathbf z_i).
    \end{equation}
\end{proposition}

\emph{\textbf{Remark.}} The trainable transformation $h^{(k)}$ increases the model capacity and can be optimized w.r.t. the learning objective to map the node embeddings into a proper latent space. For large datasets with complex inter-connecting patterns, the feature transformation $h^{(k)}$ allows the diffusion model to propagate embeddings over layer-specific latent manifolds. Our experiments found that the layer-wise transformation is not necessary for small datasets, but contributes to some performance gains for datasets with a large number of instances. Besides, while as shown by the analysis our theory can be extended to incorporate the non-linear activation $\sigma$, we empirically found that omitting the non-linearity can still perform well in quite a few cases.

\subsubsection{Incorporating Graph Inductive Bias}\label{sec-trans-dis-graph}

The model presented in Section~\ref{sec-trans-attn-difformer} does \emph{not} assume any observed graph as input. For situations with observed structures (e.g., in the form of graphs), we can leverage the structural information as a geometric prior. \revise{There potentially exist different ways to utilize the graph information. For instance, \cite{graphtransformer-2020} uses the features of Laplacian decomposition on the input graphs as absolute positional embeddings, and incorporates the positional embeddings with node features as the input for Transformers. This scheme, however, is not scalable for large graphs due to the $\mathcal O(|\mathcal V|^3)$ complexity of Laplacian decomposition. Furthermore, \cite{graphformer-neurips21} considers the encodings of node-pair-wise distances as relative positional embeddings that are used to reinforce the attention weight for any node pair. Since the distance encodings introduce extra trainable parameters and involve arbitrary node pairs in the graph (whose numbers scale quadratically w.r.t. the graph size), this approach significantly increases the training cost on large graphs.} In consideration of both effectiveness and efficiency criteria, we turn to a simple-yet-effective scheme for incorporating the graph structural information.
Denote by $\mathcal G = (\mathcal V, \mathcal E)$ the input graph, and we modify the layer-wise updating rule as:
\begin{equation}\label{eqn-diffuse-graph}
    \mathbf z_i^{(k+1)} = \left (1 - \frac{\tau}{2} \sum_{j\in \mathcal V} \left(\hat s_{ij}^{(k)} + \tilde a_{ij} \right ) \right ) \mathbf z_i^{(k)} + \frac{\tau}{2} \sum_{j\in \mathcal V} \left (\hat s_{ij}^{(k)} + \tilde a_{ij}  \right )\mathbf z_j^{(k)},
\end{equation}
where $\tilde a_{ij}$ is the connectivity weight of the edge $(i,j)\in \mathcal E$. The above model can be seen as an integration of the attention-based propagation and the graph-based propagation. In particular, one can consider the normalized adjacency for the graph-based propagation, in which case $\tilde a_{ij} = \frac{1}{\sqrt{d_id_j}}$ (or $\tilde a_{ij} = \frac{1}{d_i}$) if $(i,j)\in \mathcal E$ and 0 otherwise. By extending the proof of Theorem~\ref{thm-main}, we can show that the diffusion iteration of Eqn.~\ref{eqn-diffuse-graph} is equivalent (up to a re-scaling factor on the adjacency matrix) to a sequence of descending steps on the following regularized energy: 
\begin{equation}\label{eqn-energy-graph}
    E(\mathbf Z, k; \delta) = \|\mathbf Z - \mathbf Z^{(k)}\|_{\mathcal F}^2 + \frac{\lambda}{2}\sum_{i,j} \delta(\|\mathbf z_i - \mathbf z_j\|_2^2) + \frac{\lambda}{2}\sum_{(i,j)\in \mathcal E} \tilde a_{ij} \|\mathbf z_i - \mathbf z_j\|_2^2,
\end{equation}
where the last term contributes to a penalty for observed edges in the input graph~\citep{graphkernel}. The energy function Eqn.~\ref{eqn-energy-graph} combines the regularization effects of the non-local diffusion model with time-dependent diffusivity (i.e., enforced by the second term) and the local diffusion model with constant diffusivity (i.e., enforced by the third term).

\subsubsection{Diffusion with A Source Term: Resolving Over-Smoothing}\label{sec-trans-dis-source}

While the energy defined by Eqn.~\ref{eqn-energy} is non-convex, the minimization of the energy can still lead to the degenerate solution where all embeddings are equal to one another, in which case the first term of Eqn.~\ref{eqn-energy} is minimized to zero, and the second term is minimized because $\delta$ is non-decreasing. In this situation, there exists the potential risk for the over-smoothing problem. To fundamentally avoid such an issue, we can leverage the remedy in Section~\ref{sec-gnn-dis-source} and consider the diffusion equation with a source term. As a natural extension of our analysis in this section, we can show that Eqn.~\ref{eqn-diffuse-source} with time-dependent diffusivity descends a regularized energy whose global optimum does not cause over-smoothing.

\begin{proposition}\label{prop-trans-oversmooth}
    For step size $0 < \tau \leq 1$, the diffusion dynamics Eqn.~\ref{eqn-diffuse-source} with $\mathbf S^{(k)} = [s_{ij}^{(k)}]_{i, j\in \mathcal V}$ given by Eqn.~\ref{eqn-optimal-diffuse} and step size $0 < \tau \leq \frac{1}{\lambda_1}$, the induced numerical iteration from $\mathbf Z^{(k)}$ to $\mathbf Z^{(k+1)}$ satisfies the energy constraint $E(\mathbf Z^{(k+1)}, k; \delta) \leq E(\mathbf Z^{(k)}, k-1; \delta)$ with the global energy of the form
    \begin{equation}\label{eqn-energy2-appnp}
        E(\mathbf Z, k; \delta) = \|\mathbf Z - (\mathbf Z^{(k)} + \eta(\beta, \tau)\mathbf H)\|^2_{\mathcal F} + \lambda(\tau) \sum_{i, j} \delta( \|\mathbf z_i - \mathbf z_j\|_2^2).
    \end{equation}
\end{proposition}

\subsubsection{Scaling Up DIFFormer to Large-Scale Systems}\label{sec-trans-dis-scale}

One remaining issue for our model is how to scale the global attention to large-scale systems that involve structures (observed or unobserved) among massive numbers of nodes, e.g., up to millions. The large number of inter-connected points makes it hard for full-batch training on a single GPU. 
However, thanks to the reduced reliance on input graphs, we can harness a simple strategy for improving the space efficiency for training DIFFormer. To be specific, in each epoch, we partition the dataset into random mini-batches and feed one mini-batch (including the input features of nodes within the current mini-batch and if any, the graph adjacency composed by the observed structures among these nodes) for one feed-forward and backward computation.
In particular, for DIFFormer-s that only requires linear complexity w.r.t. node numbers, we can set a large batch size in practice which gives rise to enough global information of the interactions among nodes in one mini-batch. \revise{Also, for common large graphs with millions of nodes, the mini-batch partition is only required for training, so the global interactions can be fully accommodated at test time.}
The flexibility of mini-batch training also accommodates parallel acceleration if needed in practice. 

\begin{table}[t!]
\centering
\caption{A head-to-head comparison of various models, including multi-layer perceptrons (MLP), graph neural networks (GNN) and DIFFormer, from the perspective of our proposed energy-constrained geometric diffusion framework. The table compares these models in terms of the corresponding energy function forms, coupling matrices (induced by the diffusivity) and algorithmic complexity of one-layer propagation. \label{tbl-existing}}
\small
    \resizebox{0.99\textwidth}{!}{
\begin{tabular}{@{}c|c|c|c@{}}
\hline
\textbf{Models} & \textbf{Energy Function} $E(\mathbf Z, k; \delta)$ & \textbf{Coupling Matrix} $\mathbf S^{(k)}$ & \textbf{Complexity} \\
\hline
MLP & $\|\mathbf Z - \mathbf Z^{(k)}\|_2^2$ & $s_{ij}^{(k)} = \left\{ 
    \begin{aligned}
         &1, \quad \mbox{if} \; i = j  \\
         &0, \quad otherwise
    \end{aligned}
    \right. $ & $\mathcal O(|\mathcal V|d^2)$ \\
\hline
GCN & $\|\mathbf Z - \mathbf Z^{(k)}\|^2_{\mathcal F} + \lambda\sum_{(i,j)\in \mathcal E} s_{ij}^{(k)}\|\mathbf z_i - \mathbf z_j\|_2^2$ & $s_{ij}^{(k)} = \left\{ 
    \begin{aligned}
         &\frac{1}{\sqrt{d_id_j}}, \quad \mbox{if} \; (i,j) \in \mathcal E  \\
         &0, \quad otherwise
    \end{aligned}
    \right. $ & $\mathcal O(|\mathcal E|d^2)$  \\
\hline
GIN & $\|\mathbf Z - \mathbf Z^{(k)}\|^2_{\mathcal F} + \lambda\sum_{(i,j)\in \mathcal E} s_{ij}^{(k)}\|\mathbf z_i - \mathbf z_j\|_2^2$ & $s_{ij}^{(k)} = \left\{ 
    \begin{aligned}
         &1, \quad \mbox{if} \; (i,j) \in \mathcal E  \\
         &2, \quad \mbox{if} \; i=j~\mbox{and}~(i,j) \in \mathcal E  \\
         &0, \quad otherwise
    \end{aligned}
    \right. $ & $\mathcal O(|\mathcal E|d^2)$  \\
\hline
APPNP & $\|\mathbf Z - \mathbf Z^{(k)} - \eta\mathbf Z^{(0)}\|^2_{\mathcal F} + \lambda\sum_{(i,j)\in \mathcal E} s_{ij}^{(k)}\|\mathbf z_i - \mathbf z_j\|_2^2$ & $s_{ij}^{(k)} = \left\{ 
    \begin{aligned}
         &\frac{1}{\sqrt{d_id_j}}, \quad \mbox{if} \; (i,j) \in \mathcal E  \\
         &0, \quad otherwise
    \end{aligned}
    \right. $ & $\mathcal O(|\mathcal E|d^2)$  \\
\hline
GCNII & $\|\mathbf Z - \mathbf Z^{(k)} - \eta\mathbf Z^{(0)}\|^2_{\mathcal F} + \lambda\sum_{(i,j)\in \mathcal E} s_{ij}^{(k)}\|\mathbf z_i - \mathbf z_j\|_2^2$ & $s_{ij}^{(k)} = \left\{ 
    \begin{aligned}
         &\frac{1}{\sqrt{d_id_j}}, \quad \mbox{if} \; (i,j) \in \mathcal E  \\
         &0, \quad otherwise
    \end{aligned}
    \right. $ & $\mathcal O(|\mathcal E|d^2)$  \\
\hline
GAT & $\|\mathbf Z - \mathbf Z^{(k)}\|^2_{\mathcal F} + \lambda\sum_{(i,j)\in \mathcal E} \delta(\|\mathbf z_i - \mathbf z_j\|_2^2)$ & $s_{ij}^{(k)} = \left\{ 
    \begin{aligned}
         &\frac{c(\mathbf z_i^{(k)},\mathbf z_j^{(k)})}{\sum_{l: (i,l)\in \mathcal E} c(\mathbf z_i^{(k)},\mathbf z_l^{(k)})}, \quad \mbox{if} \; (i,j) \in \mathcal E  \\
         &0, \quad otherwise
    \end{aligned}
    \right.  $& $\mathcal O(|\mathcal E|d^2)$  \\
\hline
DIFFormer-s & $\|\mathbf Z - \mathbf Z^{(k)}\|_{\mathcal F}^2 + \lambda \sum_{i,j} \delta(\|\mathbf z_i - \mathbf z_j\|_2^2)$ & $s_{ij}^{(k)} = \frac{f(\|\mathbf z_i^{(k)} - \mathbf z_j^{(k)}\|_2^2)}{\sum_{l\in \mathcal V} f(\|\mathbf z_i^{(k)} - \mathbf z_l^{(k)}\|_2^2)}$ & $\mathcal O(|\mathcal V|d^2) $ \\
\hline
DIFFormer-a & $\|\mathbf Z - \mathbf Z^{(k)}\|_{\mathcal F}^2 + \lambda \sum_{i,j} \delta(\|\mathbf z_i - \mathbf z_j\|_2^2)$ & $s_{ij}^{(k)} = \frac{f(\|\mathbf z_i^{(k)} - \mathbf z_j^{(k)}\|_2^2)}{\sum_{l\in \mathcal V} f(\|\mathbf z_i^{(k)} - \mathbf z_l^{(k)}\|_2^2)}$ & $\mathcal O(|\mathcal V|^2d^2) $ \\
\hline
\end{tabular}}
\end{table}

\subsection{Systematic Comparisons of Different Model Classes}\label{sec-trans-comp}

To clearly compare different models within our proposed energy-constrained diffusion framework, we summarize their corresponding energy function forms and coupling matrix instantiations in Table~\ref{tbl-existing}. Specifically, we discuss their connections and differences in details. 
\begin{itemize}
    \item Multi-layer perceptrons (MLP) only consider the local consistency regularization in the energy function and only allow information flows for self-loops at each layer, i.e., the coupling matrix $\mathbf S$ only has non-zero entries $s_{ii}$'s on the diagonal.

    \item For common GNNs such as GCN~\citep{GCN-vallina}, GIN~\citep{gin} and APPNP~\citep{appnp}, they inherently minimize the energy of the quadratic form that regularizes the global consistency over the neighboring nodes of the observed graphs. The propagation rule of these GNN models induces message passing through observed edges, which can be seen as a local diffusion system with static location-dependent diffusivity and constant coupling matrix $\mathbf S$.

    \item For graph attention networks~\citep{GAT}, the energy function is similar to GCN's except that the non-linearity $\delta$ remains (as a certain specific form depending on the attention function). The concave and non-decreasing function $\delta$ introduces tolerance for some node pairs with large differences of node embeddings, particularly the node pairs that are connected but should be separated in the latent space (e.g., from different classes). In terms of the diffusivity, the GAT model only assumes non-zero $s_{ij}^{(k)}$ for the connected node pairs, yet the difference is that in such a case, $\mathbf S^{(k)}$ can flexibly change at different layers.

    \item In contrast, DIFFormer possesses the flexibility for learning adaptive pairwise influence among arbitrary node pairs, and its corresponding energy regularizes the global consistency over all node pairs in the system. The penalty function $\delta$ serves to automatically down-weigh the spurious node pairs that should be separated in the latent space. Different from MLP and GNNs, the message passing rules of DIFFormer correspond to a non-local, non-homogeneous diffusion system where the diffusivity has non-zero values for all location pairs and can temporally change as time goes by (i.e., the coupling matrix $\mathbf S^{(k)}$ is a layer-dependent dense matrix). With different attention networks, the corresponding instantiations of $\mathbf S^{(k)}$ would be different. For example, our proposed DIFFormer-s with the simple attention function can achieve significant acceleration and only requires $\mathcal O(|\mathcal V|d^2)$ linearly scaling w.r.t. the number of nodes. 
\end{itemize}

\revise{Apart from these, the proposed model framework is also related to Label Propagation~\citep{zhou2003learning} which is a classic semi-supervised learning algorithm that propagates the labels of training samples to predict those of testing samples. One straightforward way to bridge both of the worlds is to replace the node embeddings $\mathbf Z$ with labels $\mathbf Y$, on top of which one can derive the diffusion process and energy functions induced by different label propagation algorithms. Pushing further, a recent work~\citep{yang2024graph} identifies the inherent connection between MPNNs and label propagation, where the training dynamics of MPNNs can be written as a form of label propagation and the propagation matrix evolves during training. This work sheds lights on another perspective that dissects the optimization process of MPNNs. Though in general the results of \cite{yang2024graph} are orthogonal to our work, our analysis can be extended to investigate into the optimization dynamics of different MPNN models during training and reveal their implicit bias by linking the label propagation view with energy-constrained diffusion.}

\section{Experiments}\label{sec-exp}

The goal of our experiments is to validate the effectiveness of DIFFormer in diverse situations and datasets, where representation learning on complex structured data is a fundamental problem. In this regard, given the diversity of experimental tasks, the SOTA models can differ case by case, so our main target is to show the wide applicability and desired competitiveness of our models against commonly used MPNNs as well as some powerful bespoke methods tailored for different specific tasks. In the following, we first describe the overview of our experimental setup and delve into the results and discussions in each case.

The datasets in our experiments encompass geometric structures that are observed, partially observed or completely unobserved. The first scenario we study is graph-based node-level prediction tasks where input graphs are given as observation (Section~\ref{sec-exp-observed}), and we will consider graph datasets with different properties including homophilous graphs, heterophilic graphs and large-sized graphs. The second scenario we consider involves relational structures that are partially observed (Section~\ref{sec-exp-partial}), wherein we consider predictive tasks over dynamically evolving graphs as the evaluation task. The third experimental scenario is the generic predictive tasks without observed structures (Section~\ref{sec-exp-unobserved}), where one needs to infer the unobserved geometry behind data. In the last situation, we will experiment on diverse data formats that entail images, texts and physical particles. 

In each dataset, we compare with a different set of competing models closely associated and specifically designed for the particular task at hand. Also, unless otherwise stated, for datasets where input graphs are available, we incorporate them for feature propagation as is defined by Eqn.~\ref{eqn-diffuse-graph}. Detailed information about datasets and pre-processing is presented in Appendix~\ref{appx-dataset}. More implementation details including hyper-parameter searching space are deferred to Appendix~\ref{appx-implementation}.

\begin{table}[t]
\centering
\caption{Mean and standard deviation of testing Accuracy (\%) on node classification (with five different random initializations). All the models are split into groups with a comparison of non-linearity (whether the model requires activation for layer-wise transformations), PDE-solver (whether the model requires PDE-solver) and Input-G (whether the propagation purely relies on input graphs).\label{tbl-bench}}
\small
    \resizebox{\textwidth}{!}{
\begin{tabular}{@{}c|c|ccc|ccc@{}}
\hline
\textbf{Type} & \textbf{Model} & \textbf{Non-linearity} & \textbf{PDE-solver} & \textbf{Input-G}  & \textbf{Cora} & \textbf{Citeseer} & \textbf{Pubmed}   \\ 
\hline
\multirow{3}{*}{Basic models} & MLP & R   & -  & -   & 56.1 $\pm$ 1.6 & 56.7 $\pm$ 1.7 & 69.8 $\pm$ 1.5  \\
& LP & -  & -  & R   & 68.2 & 42.8 & 65.8 \\
& ManiReg & R    & -  & R  & 60.4 $\pm$ 0.8 & 67.2 $\pm$ 1.6 & 71.3 $\pm$ 1.4 \\
\hline
\multirow{8}{*}{MPNNs on Observed Graphs} 
& GCN & R   & -  & R   & 81.5 $\pm$ 1.3  & 71.9 $\pm$ 1.9 & 77.8 $\pm$ 2.9 \\
& GAT & R   & -  & R   & 83.0 $\pm$ 0.7 & 72.5 $\pm$ 0.7 & 79.0 $\pm$ 0.3 \\
& SGC & -  & -  & R   & 81.0 $\pm$ 0.0 & 71.9 $\pm$ 0.1 & 78.9 $\pm$ 0.0 \\
& GRAND-l & -  & R   & R   & 83.6 $\pm$ 1.0 & 73.4 $\pm$ 0.5 & 78.8 $\pm$ 1.7 \\
& GRAND & R   & R   & R    & 83.3 $\pm$ 1.3 & 74.1 $\pm$ 1.7 & 78.1 $\pm$ 2.1 \\
& GRAND++ & R   & R   & R    & 82.2 $\pm$ 1.1 & 73.3 $\pm$ 0.9 & 78.1 $\pm$ 0.9 \\
& GDC & R   & -  & R   & 83.6 $\pm$ 0.2 & 73.4 $\pm$ 0.3 & 78.7 $\pm$ 0.4 \\
& GraphHeat & R   & -  & R  & 83.7 & 72.5 & 80.5 \\
\hline
\multirow{8}{*}{MPNNs on Latent Graphs} 
& GCN-$k$NN & R   & -  & -    & 72.2 $\pm$ 1.8 & 56.8 $\pm$ 3.2 & 74.5 $\pm$ 3.2 \\
& GAT-$k$NN & R   & -  & -    & 73.8 $\pm$ 1.7 & 56.4 $\pm$ 3.8 & 75.4 $\pm$ 1.3 \\
& Dense GAT & R   & -  & -    & 78.5 $\pm$ 2.5 & 66.4 $\pm$ 1.5 & 66.4 $\pm$ 1.5 \\
& LDS & R   & -  & -    & 83.9 $\pm$ 0.6 & \color{brown}\textbf{74.8 $\pm$ 0.3} & out-of-memory \\
& GLCN & R   & -  & -    & 83.1 $\pm$ 0.5 & 72.5 $\pm$ 0.9 & 78.4 $\pm$ 1.5 \\
& Graphormer & R & - & R & 74.2 $\pm$ 0.9 & 63.6 $\pm$ 1.0 & out-of-memory \\
& GraphGPS & R & - & R & 80.9 $\pm$ 1.1 & 68.6 $\pm$ 1.5 & 78.5 $\pm$ 0.7 \\
& NodeFormer & -  & -  & -  & 83.4 $\pm$ 0.2 & 73.0 $\pm$ 0.3 & \color{brown}\textbf{81.5 $\pm$ 0.4} \\
\hline
\multirow{2}{*}{Ours} 
& {DIFFormer}-s & -  & -  & -    & \color{purple}\textbf{85.9 $\pm$ 0.4} & 73.5 $\pm$ 0.3 & \color{purple}\textbf{81.8 $\pm$ 0.3} \\
& {DIFFormer}-a & -   & -  & -    & \color{brown}\textbf{84.1 $\pm$ 0.6} & \color{purple}\textbf{75.7 $\pm$ 0.3} & 80.5 $\pm$ 1.2 \\
\hline
\end{tabular}}
\end{table}

\subsection{Learning with Observed Structures}\label{sec-exp-observed}

We first consider predictive tasks with observed structures and evaluate the model on diverse datasets involving homophilous graphs, heterophilic graphs and large graphs.

\textbf{Results on Homophilous Graphs.} We test DIFFormer on three citation networks \texttt{Cora}, \texttt{Citeseer} and \texttt{Pubmed} which are commonly used as benchmarks for evaluating graph representation learning approaches. Table~\ref{tbl-bench} reports the testing accuracy. We compare with several sets of baselines linked with our model from different aspects. The first category of competitors includes basic models: \emph{MLP}~\citep{mlp} and two classical graph-based semi-supervised learning approaches Label Propagation (\emph{LP})~\citep{lp-icml2003} and \emph{ManiReg}~\citep{belkin2006manireg}. The second category of competitors belongs to MPNNs on observed graphs, including commonly used GNNs (\emph{SGC}~\citep{SGC-icml19}, \emph{GCN}~\citep{GCN-vallina} and \emph{GAT}~\citep{GAT}), diffusion-based models (\emph{GRAND}~\citep{grand}, its linear variant GRAND-l, and \emph{GRAND++}~\citep{GRAND++}), and diffusion-inspired models (\emph{GDC}~\citep{klicpera2019diffusion} and \emph{GraphHeat}~\citep{xu2020heat}. The third model category extends MPNNs to latent graphs. In particular, we consider MPNNs operated on $k$-nearest-neighbor graphs (\emph{GCN-$k$NN} and \emph{GAT-$k$NN}) and latent complete graphs that connect all node pairs (\emph{Dense GAT}). Furthermore, we compare with several strong structure learning models that learn to optimize the latent structures on which MPNNs run, including LDS~\citep{LDS-icml19}, GLCN~\citep{jiang2019glcn}, and three recently proposed graph Transformer models including NodeFormer~\citep{wunodeformer}, \revise{Graphormer~\citep{graphformer-neurips21} and GraphGPS~\citep{rampavsek2022recipe}}. Table~\ref{tbl-bench} shows that DIFFormer achieves the best results on three datasets with significant improvements. Also, we notice that the simple diffusivity model DIFFormer-s significantly exceeds the counterparts without non-linearity (SGC, GRAND-l and DGC-Euler) and even comes to the first on \texttt{Cora} and \texttt{Pubmed}. These results suggest that DIFFormer can serve as a very competitive encoder backbone for node-level prediction that learns inter-instance interactions for generating informative representations and boosting downstream performance.


\begin{table}[t]
\centering
\caption{Testing ROC-AUC (\%) for \texttt{Proteins} and Accuracy (\%) for \texttt{Pokec} on large-scale graph datasets. $*$ denotes using mini-batch training. \label{tbl-largebench}}
\small
    \resizebox{\textwidth}{!}{
\begin{tabular}{@{}c|cccccc|c@{}}
\hline
\textbf{Models} & MLP & LP & SGC & GCN & GAT & NodeFormer & DIFFormer-s   \\ 
\hline
\textbf{Proteins} & 72.41 $\pm$ 0.10 & 74.73 & 49.03 $\pm$ 0.93 & 74.22 $\pm$ 0.49$^*$ & 75.11 $\pm$ 1.45$^*$ & \color{brown}\textbf{77.45 $\pm$ 1.15}$^*$ & \color{purple}\textbf{79.49 $\pm$ 0.44}$^*$ \\
\hline
\textbf{Pokec} & 60.15 $\pm$ 0.03 & 52.73 &  52.03 $\pm$ 0.84 & 62.31 $\pm$ 1.13$^*$ &  65.57 ± 0.34$^*$ & \color{brown}\textbf{68.32 $\pm$ 0.45}$^*$ & \color{purple}\textbf{69.24 $\pm$ 0.76}$^*$\\
\hline
\end{tabular}}
\end{table}



\textbf{Results on Heterophily Graphs.} We next study heterophily graphs where the connected nodes tend to have different classes. We consider three widely used heterophily graph datasets \texttt{Chameleon}, \texttt{Squirrel} and \texttt{Actor}.\footnote{For the first two datasets, a recent work~\citep{heter-iclr23} identifies that their original public splits are problematic (with overlapped nodes in the training and testing sets), so we adopt the new splits introduced by \cite{heter-iclr23}.} We basically consider the competitors MLP and three common GNNs, i.e., SGC, GCN and GAT. Moreover, to demonstrate the effectiveness of our model, we also compare with several strong models that are tailored for handling heterophily graphs, including H2GCN~\citep{h2gcn-neurips20}, FSGNN~\citep{fsgnn}, CPGNN~\citep{cpgnn} and GloGNN~\citep{glognn}. As shown by the results in Table~\ref{tbl-heter}, DIFFormer achieves the highest scores among three cases and even outperforms the bespoke GNNs designed for heterophily graphs, which verifies the efficacy of our model in non-homophilous datasets. In such a situation, the input graphs may contain some irrelevant information and are not reliable for propagating beneficial information. However, the global attention of DIFFormer can help to learn adaptive structures and flexibly propagate useful information across dis-connected nodes in the graph.

\textbf{Results on Large-Sized Graphs.} To demonstrate the scalability of DIFFormer, we conduct experiments on two large-scale graph datasets \texttt{ogbn-Proteins}, a multi-task protein-protein interaction network, and \texttt{Pokec}, a social network. 
Table~\ref{tbl-largebench} presents the results. Due to the dataset size (0.13M/1.63M nodes for two graphs) and scalability issues that many of the competitors in Table~\ref{tbl-bench} as well as DIFFormer-a would potentially experience, we only compare DIFFormer-s with the scalable competitors. In particular, we found GCN/GAT/NodeFormer/DIFFormer-s are still hard for full-graph training on a single V100 GPU with 16GM memory. We thus consider mini-batch training with batch size 10K/100K for \texttt{Proteins}/\texttt{Pokec}. We found that DIFFormer outperforms common GNNs by a large margin, which suggests its desired efficacy on large datasets. As mentioned previously, we prioritize the efficacy of DIFFormer as a general encoder backbone for solving node-level prediction tasks on large graphs. While there are quite a few practical tricks shown to be effective for training GNNs for this purpose, e.g., hop-wise attention~\citep{adadiffusion-2022} or various label re-use strategies, these efforts are largely orthogonal to our contribution here and can be applied to most of models to further boost performance. For further investigation, we supplement more results using different mini-batch sizes for training and study its impact on testing performance in Appendix~\ref{appx-res-batch}. Furthermore, we compare the training time and memory costs in Appendix~\ref{appx-res-time}.

\begin{figure}[t]
\begin{minipage}{0.42\linewidth}
    \centering
	\captionof{table}{Testing Accuracy (\%) on heterophily graphs.\label{tbl-heter}}
    \resizebox{0.98\textwidth}{!}{
\begin{tabular}{@{}l|ccc@{}}
\hline
\textbf{Models} & \textbf{Chameleon} & \textbf{Squirrel} & \textbf{Actor}   \\ 
\hline
MLP & 36.7 ± 4.7 & 36.5 ± 1.8 & 28.9 ± 0.8 \\
SGC & 36.0 ± 3.2 & 38.1 ± 1.8 & 29.2 ± 1.2 \\
GCN & 41.3 ± 3.0 & 38.6 ± 1.8 & 30.1 ± 0.2 \\
GAT & 39.2 ± 3.0 & 35.6 ± 2.0 & 29.6 ± 0.6 \\
H2GCN & 26.7 ± 3.6 & 35.1 ± 1.1 & 34.5 ± 1.6 \\
FSGNN & 40.6 ± 2.9 & 35.9 ± 1.3 & 35.7 ± 0.9 \\
CPGNN &  33.0 ± 3.1 & 30.0 ± 2.0 & 34.7 ± 0.7 \\
GloGNN & 25.9 ± 3.5  & 35.1 ± 1.2 & 36.0 ± 1.6 \\
\hline
DIFFormer-s & \color{brown}\textbf{42.5 ± 2.5} & \color{brown}\textbf{38.8 ± 0.8} & \color{purple}\textbf{36.5 ± 0.7} \\
DIFFormer-a & \color{purple}\textbf{42.8 ± 4.4} & \color{purple}\textbf{40.5 ± 1.8} & \color{brown}\textbf{36.4 ± 0.8} \\
\hline
\end{tabular}}
    \end{minipage}
    \hspace{5pt}
    \begin{minipage}{0.55\linewidth}
    \centering
	\captionof{table}{Mean and standard deviation of MSE on spatial-temporal prediction datasets.\label{tbl-temporal}}
    \resizebox{0.98\textwidth}{!}{
\begin{tabular}{@{}l|ccc@{}}
\hline
\textbf{Models} & \textbf{Chickenpox} & \textbf{Covid} & \textbf{WikiMath}   \\ 
\hline
MLP & 0.924 ± 0.001 & 0.956 ± 0.198 & 1.073 ± 0.042 \\
GCN & 0.923 ± 0.001 & 1.080 ± 0.162 & 1.292 ± 0.125 \\
GAT & 0.924 ± 0.002 & 1.052  ± 0.336 & 1.339 ± 0.073 \\
Dense GAT & 0.935 ± 0.002 & 1.052 ± 0.336 & 1.339 ± 0.073 \\
GAT-kNN & 0.926 ± 0.004 & 0.861 ± 0.123 & 0.882 ± 0.015 \\
GCN-kNN &  0.936 ± 0.004 & 1.475 ± 0.560 & 1.023 ± 0.058 \\
\hline
DIFFormer-s & \color{purple}\textbf{0.914 ± 0.006}  & 0.779 ± 0.037 & 0.731 ± 0.007 \\
DIFFormer-a & \color{brown}\textbf{0.915 ± 0.008} & \color{brown}\textbf{0.757 ± 0.048} & 0.763 ± 0.020 \\
DIFFormer-s w/o g & 0.916 ± 0.006 & 0.779 ± 0.028 & \color{brown}\textbf{0.727 ± 0.025} \\
DIFFormer-a w/o g & 0.916 ± 0.006 & \color{purple}\textbf{0.741 ± 0.052} & \color{purple}\textbf{0.716 ± 0.030} \\
\hline
\end{tabular}}
    \end{minipage}
\end{figure}




\subsection{Learning with Partially Observed Structures}\label{sec-exp-partial}

There also exist practical scenarios where the observed graphs are incomplete and dynamically evolved. We consider three spatial-temporal datasets with details described in Appendix~\ref{appx-dataset}. Each dataset consists of a series of graph snapshots where nodes are treated as instances and each of them has a integer label (e.g., reported cases for \texttt{Chickenpox} or \texttt{Covid}). The task is to predict the labels of one snapshot based on the previous ones. 
Table~\ref{tbl-temporal} compares testing MSE of four DIFFormer variants (here DIFFormer-s w/o g denotes the model DIFFormer-s without using input graphs) with the scalable competitors. We can see that two DIFFormer variants without input graphs even outperform the counterparts using input structures in four out of six cases. This implies that our attention module could learn useful structures for informed prediction, and the input structure might not always contribute to positive effect. In fact, for temporal dynamics, the underlying relations that truly influence the trajectory evolution can be much complex and the observed relations could be unreliable with missing or noisy links, in which case GNN models relying on input graphs may perform undesirably. Compared to the competitors, our models utilizing the latent interactions rank the first with significant improvements. We present more analysis based on visualization in Appendix~\ref{sec-exp-dis}.

\subsection{Learning with Unobserved Structures}\label{sec-exp-unobserved}

The last scenario we consider requires the model to handle unobserved structures, e.g., data manifold geometries or unknown interactions among physical particles.

\textbf{Results on Images and Texts.} As mentioned previously, DIFFormer can be applied to no-graph tasks where the inter-dependencies of input instances are unknown.
We next conduct experiments on \texttt{CIFAR-10}, \texttt{STL-10} and \texttt{20News-Group} datasets to test DIFFormer for standard classification tasks with limited label rates.
For \texttt{20News} provided by \cite{pedregosa2011scikit}, we take 10 topics and use words with TF-IDF more than 5 as features. 
For \texttt{CIFAR} and \texttt{STL}, two public image datasets, we first use the self-supervised approach SimCLR~\citep{chen2020simple} (that does not use labels for training) to train a ResNet-18 for extracting the feature maps as input features of instances.
These datasets contain no graph structure, so we use the $k$-nearest-neighbor to construct a graph over input features for GNN competitors and do \emph{not} use input graphs for DIFFormer. Table~\ref{tab:image classification} reports the testing accuracy of DIFFormer and competitors including the basic models (MLP and ManiReg) and MPNNs operated on latent graphs (GCN-$k$NN, GAT-$k$NN, GLCN and NodeFormer). We found that two DIFFormer models perform much better than MLP in nearly all cases, suggesting the effectiveness of learning the inter-dependencies over instances. Besides, DIFFormer yields large improvements over GCN and GAT which are in some sense limited by the handcrafted graph that leads to sub-optimal propagation. Moreover, DIFFormer significantly outperforms GLCN and NodeFormer, two strong competitors that learn new graph structures for message passing, which demonstrates the superiority of our proposed model in leveraging the global information from all-pair interactions.


\textbf{Results on Physical Particles.} We proceed to apply our model to particle property prediction which has extensive application scenarios in particle physics~\citep{guest2018deep,shlomi2020graph} and molecular analysis~\citep{mccloskey2019using}. The task is to predict the property of particles, and each particle is composed of a sets of points in 3D Euclidean space that have unobserved physical interactions. Thereby, the labels are dependent on not only the node features (i.e., attributes of points) but also the latent structures that are unavailable in data yet affect the data generation. The predictive task is binary classification, and the detailed dataset information is deferred to Appendix~\ref{appx-dataset}.
Similar to the image and text datasets, since there is no input structure, we use $k$NN to construct a graph for each sample by computing the distance between points in the observed Euclidean space. We only use the $k$NN graph for GNN competitors and do not use it for DIFFormer.
Table~\ref{tab:particle} presents the testing results with the evaluation metric ROC-AUC. 
The results show that our two models achieve significantly superior classification performance, which demonstrates its practical efficacy for learning effective representations without any observed structures. 

\begin{table}[tb!]
		\centering
		\small
		\caption{Testing Accuracy (\%) for image (\texttt{CIFAR} and \texttt{STL}) and text (\texttt{20News}) classification.}
		\label{tab:image classification}
		\resizebox{\textwidth}{!}{
		\begin{tabular}{c|c|cccccc|cc}
		\hline
		\textbf{Dataset} & \textbf{\# Labels}  & MLP  & ManiReg  & GCN-$k$NN & GAT-$k$NN  & GLCN  & NodeFormer & DIFFormer-s & DIFFormer-a  \\
		\hline
		\multirow{3}{*}{\textbf{CIFAR}} 
		& 100 &65.9 ± 1.3&67.0 ± 1.9&66.7 ± 1.5&66.0 ± 2.1  &66.6 ± 1.4&67.5 ± 1.0&\color{brown}\textbf{69.1 ± 1.1}&\color{purple}\textbf{69.3 ± 1.4} \\
		& 500 &73.2 ± 0.4&72.6 ± 1.2&72.9 ± 0.4&72.4 ± 0.5 &72.8 ± 0.5&72.6 ± 0.3&\color{purple}\textbf{74.8 ± 0.5}&\color{brown}\textbf{74.0 ± 0.6} \\
		& 1000 &75.4 ± 0.6&74.3 ± 0.4&74.7 ± 0.5&74.1 ± 0.5 &74.7 ± 0.3&75.0 ± 0.2&\color{purple}\textbf{76.6 ± 0.3}&\color{brown}\textbf{75.9 ± 0.3} \\
		\hline
		\multirow{3}{*}{\textbf{STL}} 
		& 100 &66.2 ± 1.4&66.5 ± 1.9&\color{brown}\textbf{66.9 ± 0.5}&66.5 ± 0.8 &66.4 ± 0.8&65.9 ± 1.0&\color{purple}\textbf{67.8 ± 1.1}&66.8 ± 1.1 \\
		& 500 &\color{brown}\textbf{73.0 ± 0.8}&72.5 ± 0.5&72.1 ± 0.8&72.0 ± 0.8 & 72.4 ± 1.3&72.1 ± 0.8&\color{purple}\textbf{73.7 ± 0.6}&72.9 ± 0.7 \\
		& 1000 &75.0 ± 0.8&74.2 ± 0.5&73.7 ± 0.4&73.9 ± 0.6 & 74.3 ± 0.7&74.2 ± 0.4&\color{purple}\textbf{76.4 ± 0.5}&\color{brown}\textbf{75.3 ± 0.6} \\
		\hline
		\multirow{3}{*}{\textbf{20News}} 
		& 1000  &54.1 ± 0.9&56.3 ± 1.2&56.1 ± 0.6&55.2 ± 0.8 &56.2 ± 0.8&56.4 ± 0.7&\color{brown}\textbf{57.7 ± 0.3}&\color{purple}\textbf{57.9 ± 0.7} \\
		& 2000   &57.8 ± 0.9&60.0 ± 0.8&60.6 ± 1.3&59.1 ± 2.2 &60.2 ± 0.7&59.5 ± 0.9&\color{brown}\textbf{61.2 ± 0.6}&\color{purple}\textbf{61.3 ± 1.0}  \\
		& 4000  &62.4 ± 0.6&63.6 ± 0.7&64.3 ± 1.0&62.9 ± 0.7 &64.1 ± 0.8&64.1 ± 0.7&\color{purple}\textbf{65.9 ± 0.8}&\color{brown}\textbf{64.8 ± 1.0}  \\
		\hline
	\end{tabular}				
		}	\end{table}


\begin{table}[tb!]
    \centering
    \small
    \caption{Testing ROC-AUC (\%) for particle property prediction on \texttt{ActsTrack} and \texttt{SynMol}.}
    \label{tab:particle}
    \resizebox{\textwidth}{!}{
    \begin{tabular}{c|ccccc|cc}
    \hline
    \multicolumn{1}{c|}{\textbf{Dataset}}  & MLP & GCN-$k$NN & GAT-$k$NN & GLCN & NodeFormer & DIFFormer-s & DIFFormer-a \\
    \hline
    \multirow{1}{*}{\textbf{ActsTrack}} 
    &99.94 ± 0.01 &99.89 ± 0.01&99.80 ± 0.05  & 99.10 ± 0.04 & 99.86 ± 0.04 &\color{purple}\textbf{100.00 ± 0.00}&\color{purple}\textbf{100.00 ± 0.00} \\
    \hline
    \multirow{1}{*}{\textbf{SynMol}} 
    &69.07 ± 0.42&68.79 ± 0.05&65.71 ± 0.87 & 68.68 ± 0.52& 69.92 ± 0.42 & \color{brown}\textbf{71.76 ± 0.25}& \color{purple}\textbf{73.28 ± 1.65} \\
    \hline
    \end{tabular}				
}	\end{table}

\begin{figure}[t!]
    \begin{minipage}{\linewidth}
    \begin{minipage}[t]{0.32\linewidth}
    \centering
    \includegraphics[width=\linewidth]{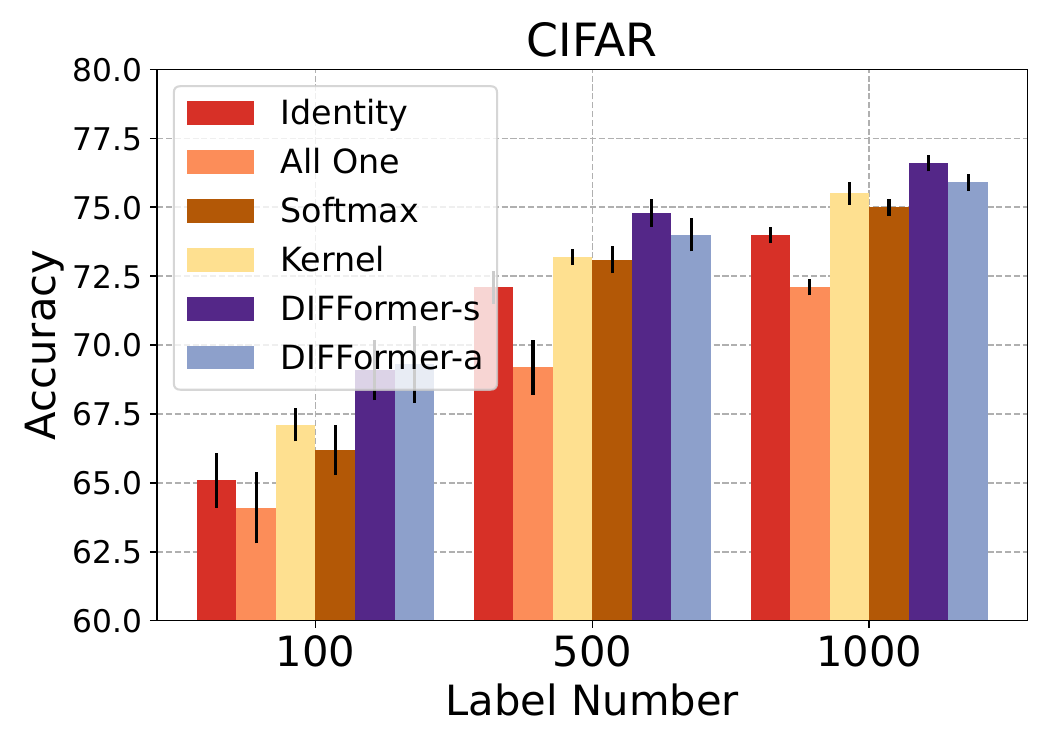}
    \end{minipage}
    \begin{minipage}[t]{0.32\linewidth}
    \centering
    \includegraphics[width=\linewidth]{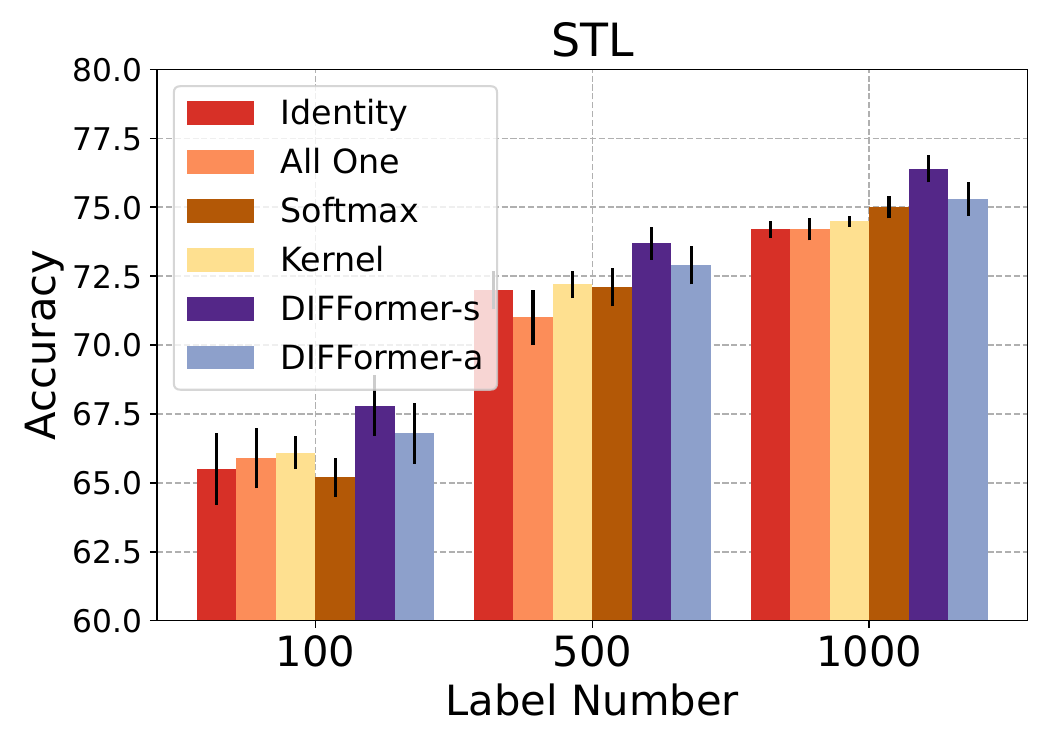}
    \end{minipage}
    \begin{minipage}[t]{0.32\linewidth}
    \centering
    \includegraphics[width=\linewidth]{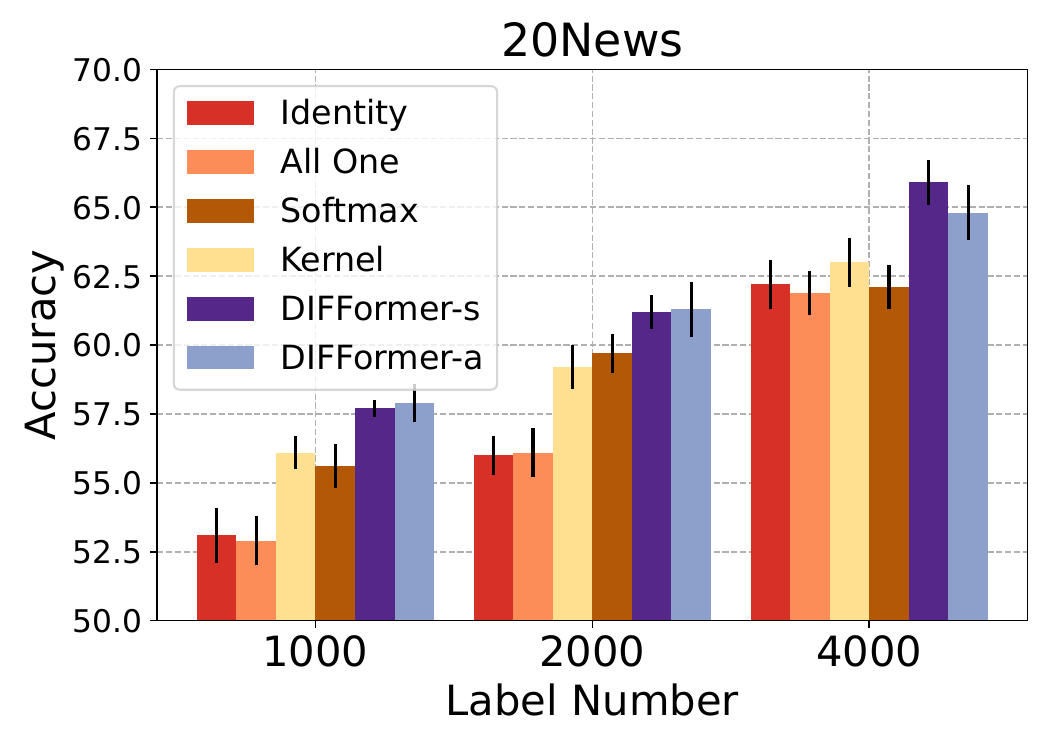}
    \end{minipage}
    \caption{Ablation studies with respect to different instantiations of attention functions.\label{fig-abl}}
    \end{minipage}
\end{figure}

\subsection{Further Results and Discussions}\label{sec-exp-dis}

We next conduct more experiments to verify the effectiveness of the model, including ablation studies on the key model component, discussions on the important hyper-parameters and visualization of the embeddings and structures.

\textbf{Ablation Studies on Attention Functions.} To verify the practical efficacy of our proposed attention functions (used by DIFFormer-s and DIFFormer-a, respectively), we compare the two instantiations proposed in Section~\ref{sec-trans-attn-difformer} with other potential choices. Figure~\ref{fig-abl} presents the comparison with four variants using different attention forms: 1) \emph{Identity} sets $\mathbf S^{(k)}$ as a fixed identity matrix; 2) \emph{All One} fixes $\mathbf S^{(k)}$ as an all-one constant matrix; 3) \emph{Softmax} parameterizes $\mathbf S^{(k)}$ with the dot-then-exponential Softmax attention networks  (i.e., Eqn.~\ref{eqn-trans-attn}) used by \cite{transformer}; 4) \emph{Kernel} adopts Gaussian kernel for computing $\mathbf S^{(k)}$. We found that across the three datasets we demonstrate, our adopted attention forms produce superior performance, which verifies the effectiveness of our attention designs derived from minimization of a principled regularized energy.

\begin{figure}[t]
    \centering
    \includegraphics[width=\linewidth]{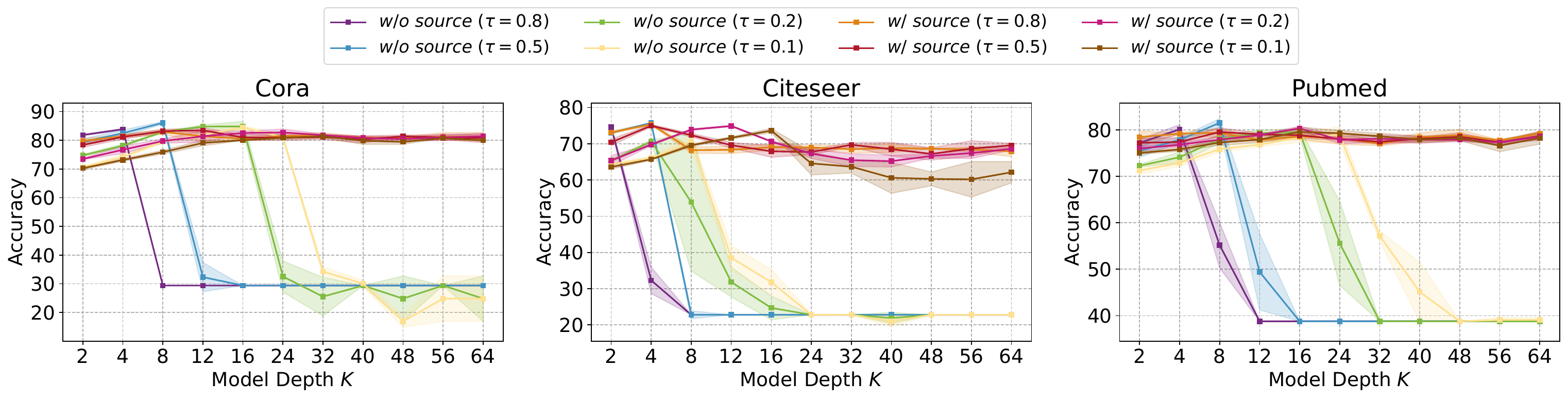}
    \caption{Hyper-parameter analysis of DIFFormer w/o and w/ the source term when using different settings of model depth $K$ and step size $\tau$.\label{fig-hyper}}
\end{figure}

\textbf{Impact of Hyperparameters $K$ and $\tau$.} We next study the influence of model depth $K$ (that controls the number of propagation layers) and step size $\tau$ (that controls the weight for residual connection and attentive propagation) on our models. Figure~\ref{fig-hyper} presents the results on three citation networks, where we compare the model implementations w/o and w/ the source term presented in Section~\ref{sec-trans-dis-source}. We found that when not using the source term, the model performance would yield the optimal performance with a moderate value of $K$, which corresponds to a proper number of propagation layers. However, when $K$ further increases, the model performance would dramatically degrade. This can be caused by the over-smoothing problem where the deep model leads to indistinguishable node embeddings that degrade to a single point in the latent space and are uninformative. We also found that for smaller $\tau$ (that assigns less importance on the attentive propagation in each layer), the safety zone for $K$ can be enlarged to a certain degree yet the over-smoothing issue cannot be avoided when $K$ is set large enough. Fortunately, the over-smoothing issue can be avoided by adding the simple source term,\footnote{We instantiate the source term as the initial embedding of each node, i.e., $\mathbf h_i = \mathbf z^{(0)}$, and set the weight $\beta=1$ throughout all the cases.} which makes the model become stable and robust for deep propagation layers.

\begin{figure}[t!]
\centering
\begin{minipage}{\linewidth}
\centering
\subfigure[The first layer]{
\begin{minipage}[t]{0.22\linewidth}
\centering
\includegraphics[width=\linewidth,trim=0cm 0cm 0cm 0cm,clip]{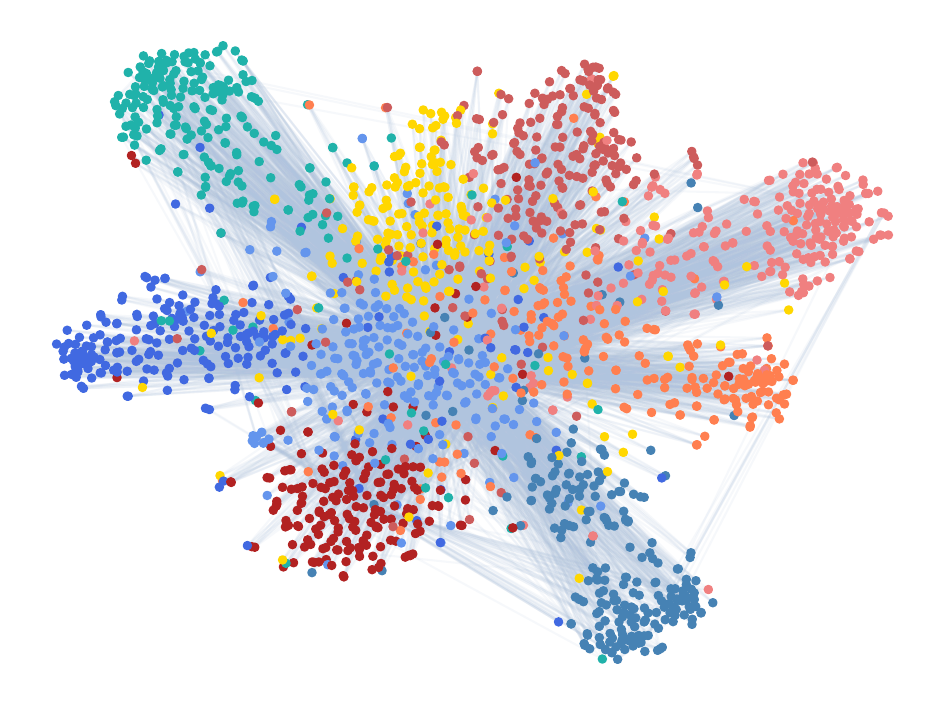}
\end{minipage}
}
\subfigure[The last layer]{
\begin{minipage}[t]{0.22\linewidth}
\centering
\includegraphics[width=\linewidth,trim=0cm 0cm 0cm 0cm,clip]{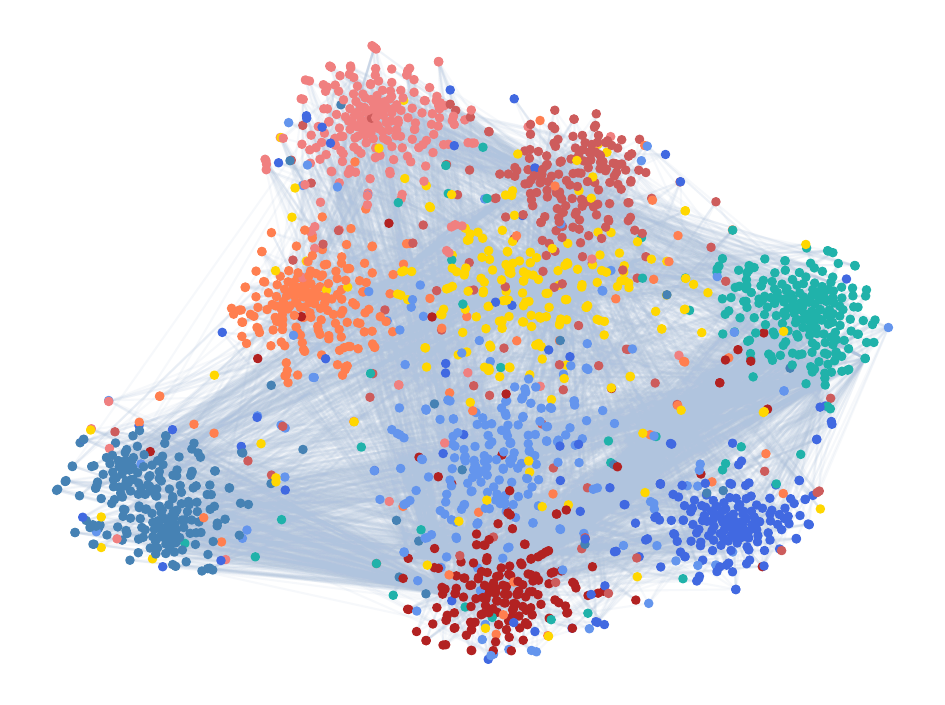}
\end{minipage}
}
\subfigure[The first layer]{
\begin{minipage}[t]{0.22\linewidth}
\centering
\includegraphics[width=\linewidth,trim=0cm 0cm 0cm 0cm,clip]{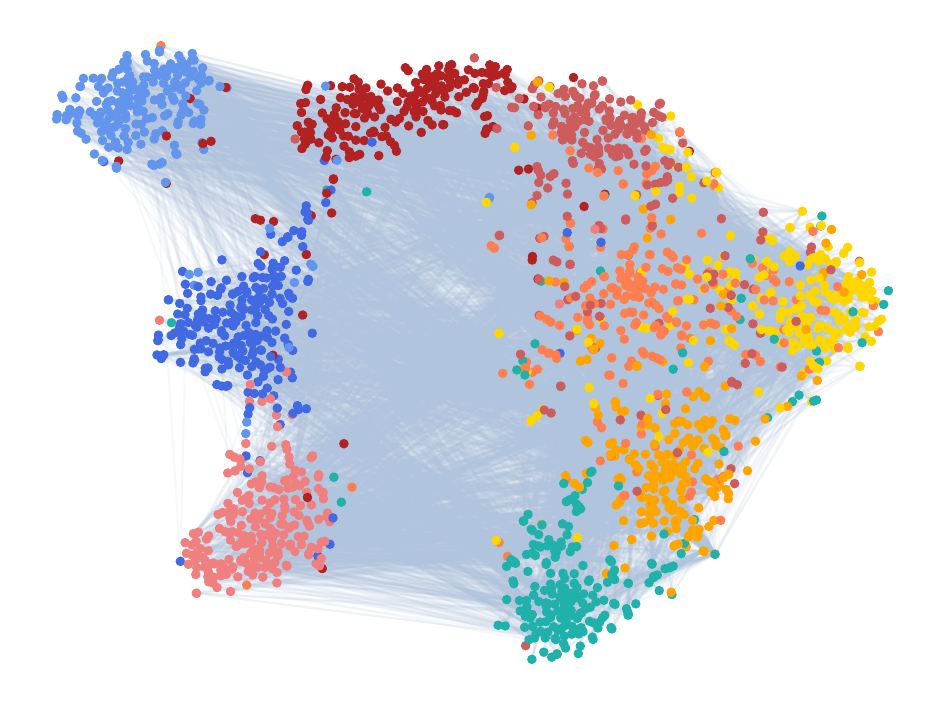}
\end{minipage}
}
\subfigure[The last layer]{
\begin{minipage}[t]{0.22\linewidth}
\centering
\includegraphics[width=\linewidth,trim=0cm 0cm 0cm 0cm,clip]{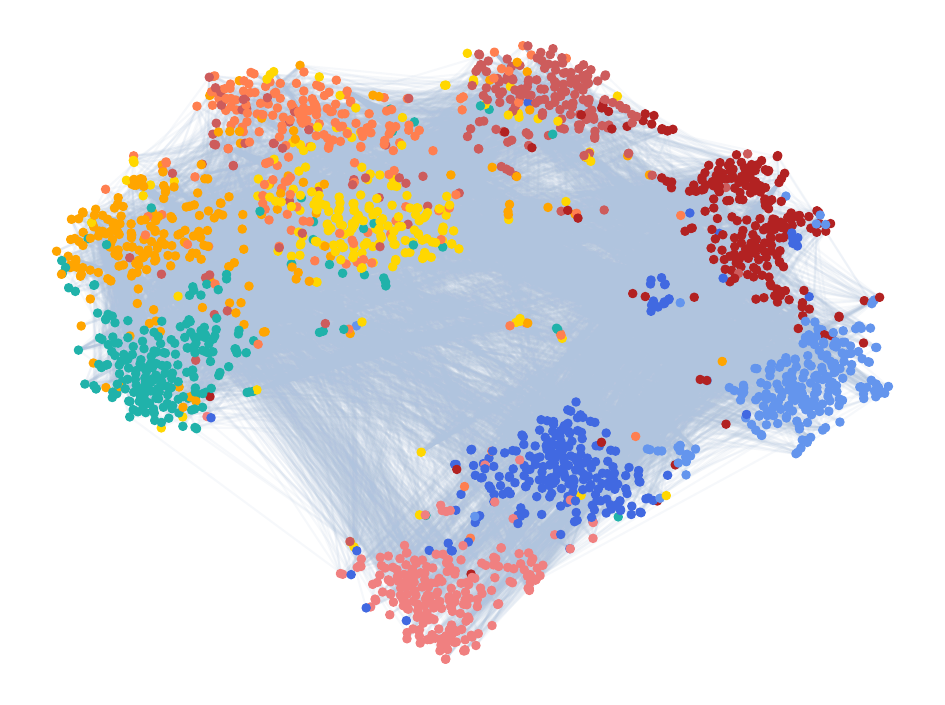}
\end{minipage}
}
\end{minipage}
\caption{Visualization of node embeddings and attention weights (we set a threshold and only plot the edges with weights more than the threshold) at different layers given by DIFFormer-s on \texttt{20News} (a)$\sim$(b) and \texttt{STL} (c)$\sim$(d).}
\label{fig:vis}
\end{figure}

\begin{figure}[t!]
\centering
\begin{minipage}{\linewidth}
\centering
\subfigure[]{
\begin{minipage}[t]{0.14\linewidth}
\centering
\includegraphics[width=\linewidth,trim=0cm 0cm 0cm 0cm,clip]{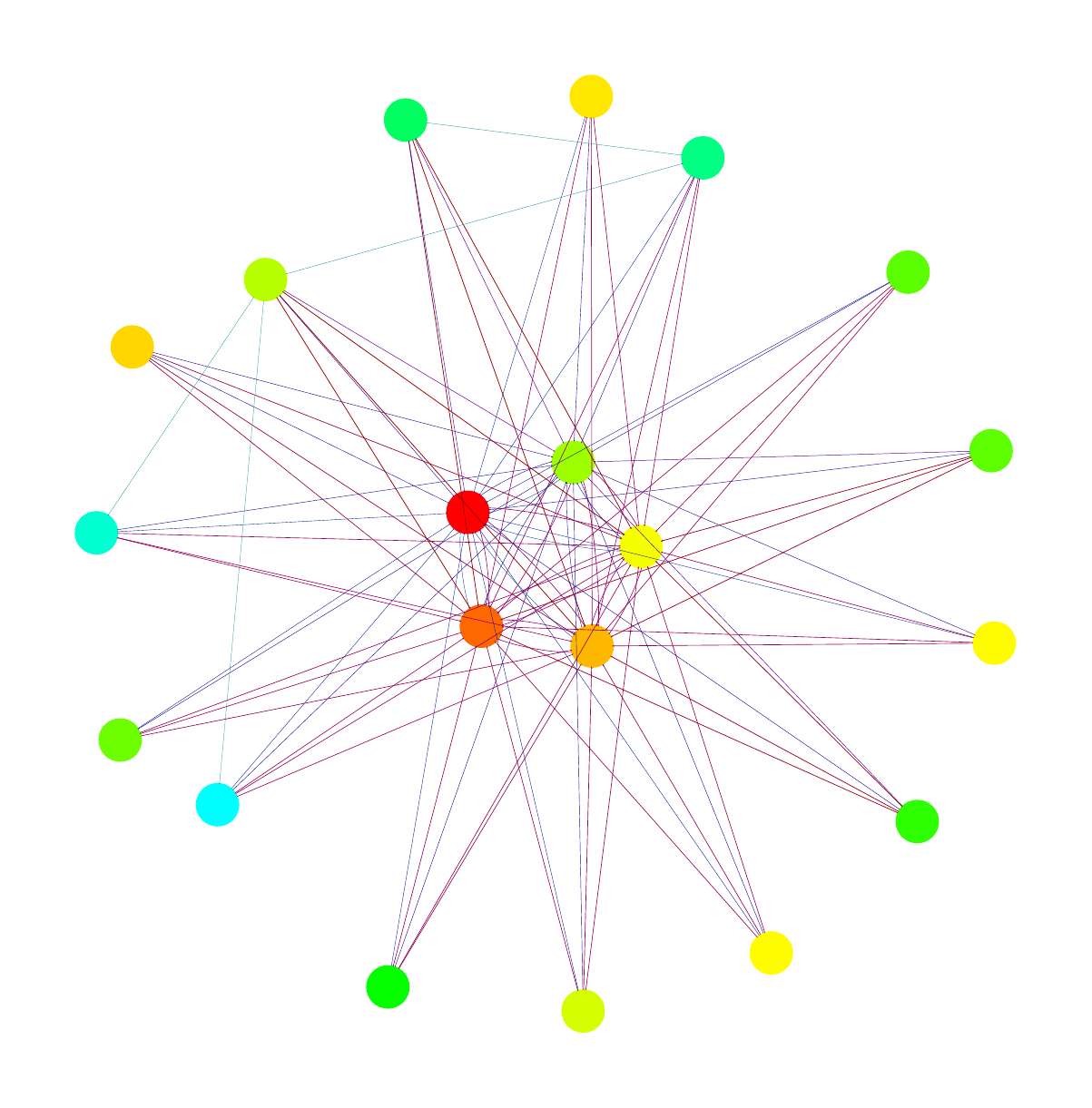}
\end{minipage}
}
\subfigure[]{
\begin{minipage}[t]{0.14\linewidth}
\centering
\includegraphics[width=\linewidth,trim=0cm 0cm 0cm 0cm,clip]{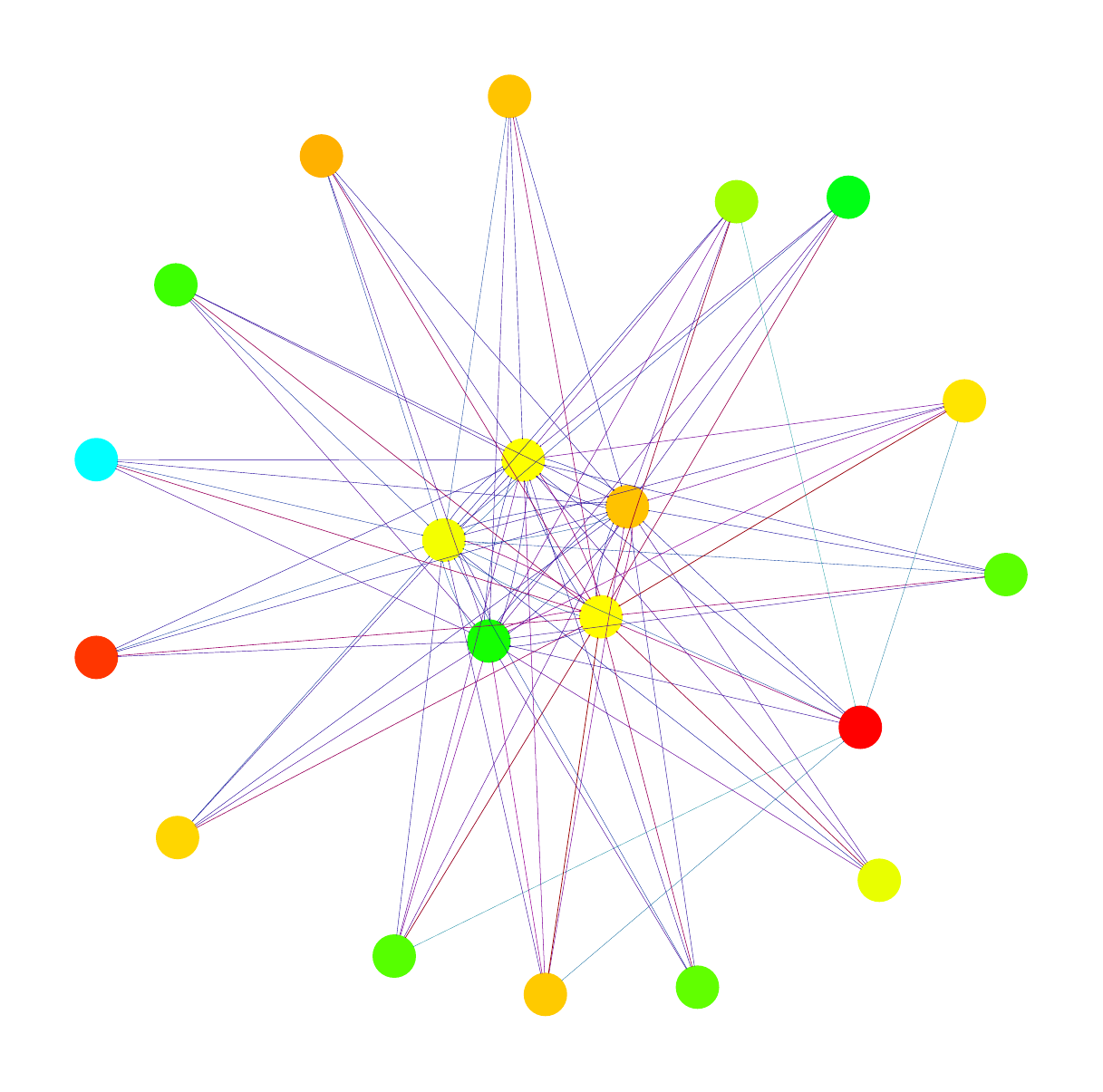}
\end{minipage}
}
\subfigure[]{
\begin{minipage}[t]{0.14\linewidth}
\centering
\includegraphics[width=\linewidth,trim=0cm 0cm 0cm 0cm,clip]{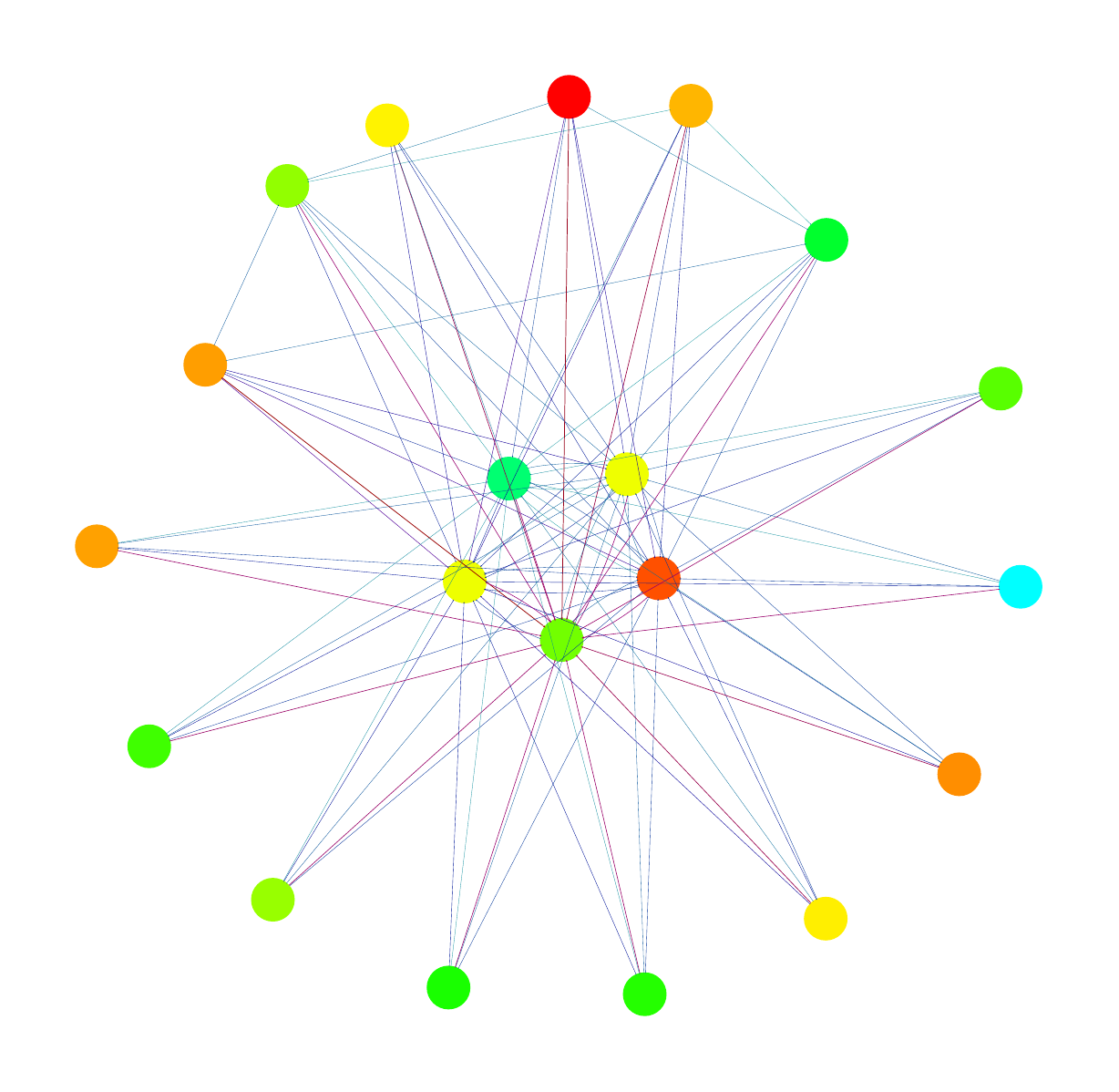}
\end{minipage}
}
\subfigure[]{
\begin{minipage}[t]{0.14\linewidth}
\centering
\includegraphics[width=\linewidth,trim=0cm 0cm 0cm 0cm,clip]{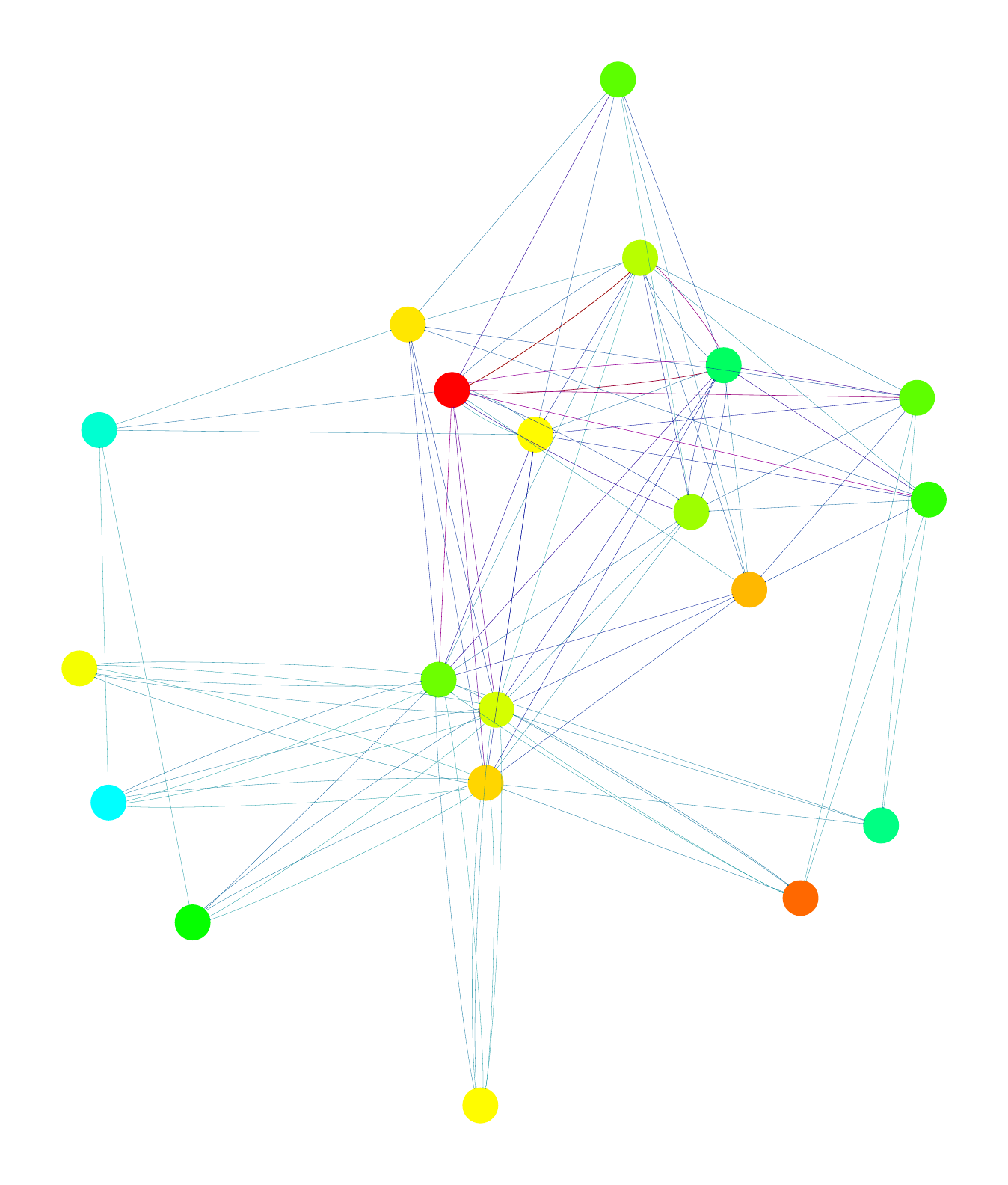}
\end{minipage}
}
\subfigure[]{
\begin{minipage}[t]{0.14\linewidth}
\centering
\includegraphics[width=\linewidth,trim=0cm 0cm 0cm 0cm,clip]{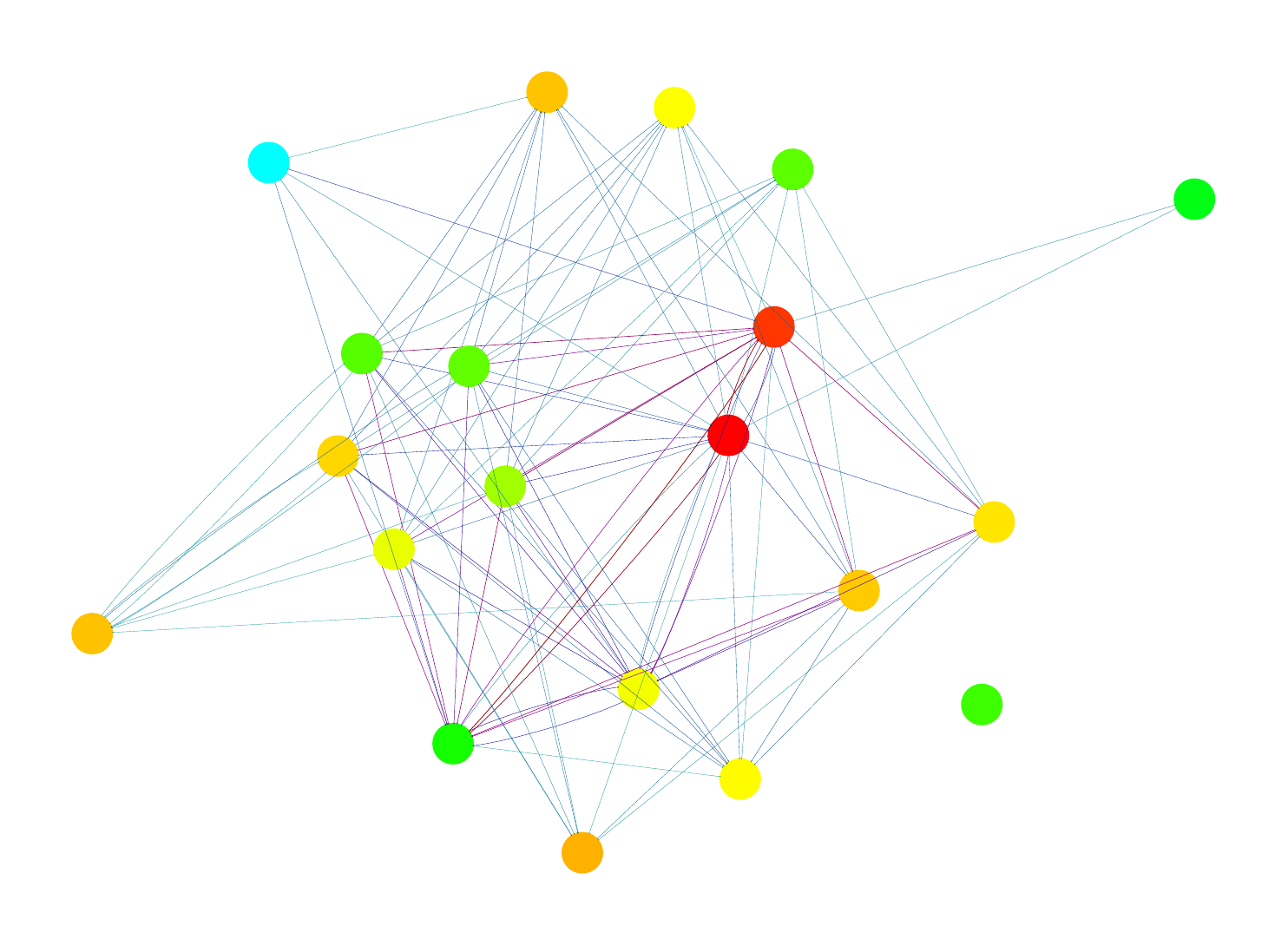}
\end{minipage}
}
\subfigure[]{
\begin{minipage}[t]{0.14\linewidth}
\centering
\includegraphics[width=\linewidth,trim=0cm 0cm 0cm 0cm,clip]{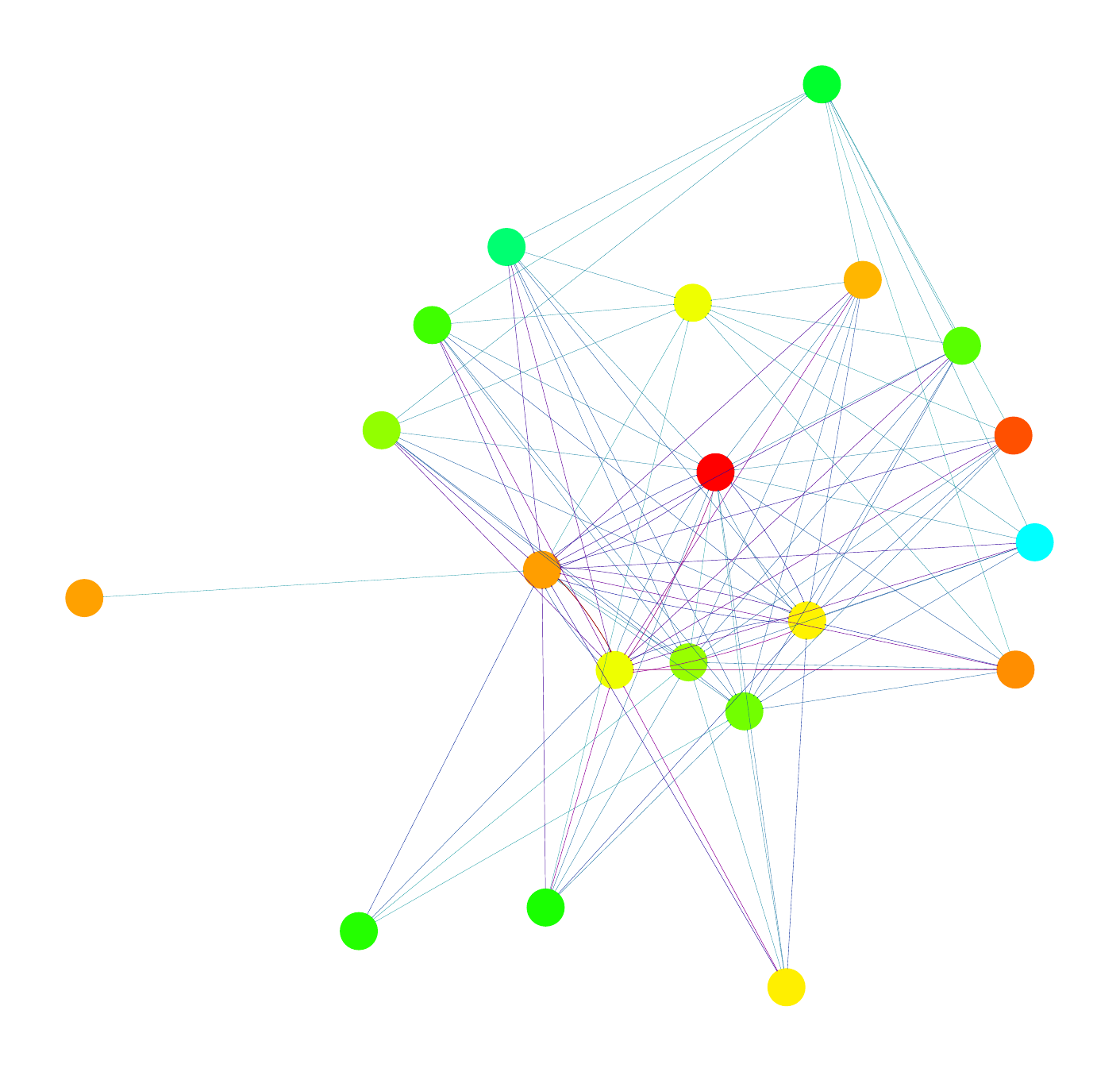}
\end{minipage}
}
\end{minipage}
\caption{The produced attentive graphs of the first layer (i.e., $\hat{\mathbf S}^{(1)}$) on \texttt{Chickenpox} across the first three snapshots, yielded by DIFFormer-s (a)$\sim$(c) and DIFFormer-a (d)$\sim$(f). Node colors correspond to ground-truth labels (i.e., reported cases), varying from red to blue as the label increases. We visualize the edges with top 100 attention weights, where edge colors change from blue to red as $\hat s^{(1)}_{ij}$ increases.}
\label{chickenpox_main_text_visualization}
\end{figure}

\textbf{Visualization.} Apart from the quantitative results, we provide some qualitative analysis on how the model behaves for learning node representations and latent structures. Figure~\ref{fig:vis}(a)-(d) plot the produced node embeddings and attention weights estimated by the model on \texttt{20News} and \texttt{STL}. We observe that the attention estimates tend to connect nodes from different clusters, which might contribute to increasing the global connectivity and facilitate absorbing other instances' information for informative representations. The node embeddings produced by our model tend to have small intra-class distance and large inter-class distance, making it easier for the classifier to distinguish instances from different classes. Figure~\ref{chickenpox_main_text_visualization} visualizes the attention estimates on \texttt{Chickenpox}. We observe that large connectivity weights usually exists between nodes with similar ground-truth labels. The produced attentive graphs on DIFFormer-s tend to have regular forms, while those on DIFFormer-a exhibit some special shapes. This suggests that DIFFormer-a indeed learns more complex underlying structures than DIFFormer-s due to its better capacity for latent structure learning.

\section{Conclusions}

This paper proposes an energy-constrained geometric diffusion model that serves as a principled theoretical framework for representation learning with complex structured data. We show that in the context of learning on graphs or latent structures, there exists a one-to-one correspondence between the diffusion operators and global energy functions implicitly minimized by the diffusion dynamics. On top of this, the finite-difference iterations of energy-constrained diffusion induce the propagational architectures of various MPNNs and Transformers. In light of these theoretical results, we propose a new class of neural encoder architectures, dubbed as DIFFormer, with two practical implementations that possess desired scalability and expressivity for learning complex all-pair interactions over the underlying data geometry. Extensive experiments demonstrate the effectiveness and superiority of the model in a wide range of tasks and datasets.

\acks{The SJTU authors were partly supported by NSFC (92370201, 62222607). The authors would like to thank the anonymous reviewers of ICLR 2023 as well as Michael Bronstein, Hongyuan Zha and Shi Jin for their insightful comments and suggestions on this work.}


\newpage

\appendix

\section{Proofs}\label{appx-proof}

\subsection{Proof for Theorem~\ref{thm-gcn}}

    We prove the theorem by construction. Define $\tilde{\mathbf D} = \mbox{diag}(\{\tilde d_i\})_{i\in \mathcal V}$ as a diagonal degree matrix associated with $\mathbf S = [s_{ij}]_{i, j \in \mathcal V}$, where $\tilde d_i = \sum_{j\in \mathcal V} s_{ij}$. Then the quadratic energy defined by Eqn.~\ref{eqn-energy-gcn} can be written as 
    \begin{equation}\label{eqn-proof-energy}
        E(\mathbf Z, k) = \|\mathbf Z - \mathbf Z^{(k)}\|^2_{\mathcal F} + \lambda \mbox{tr}(\mathbf Z^\top (\tilde{\mathbf D} - \mathbf S)\mathbf Z).
    \end{equation}
    For minimizing $E(\mathbf Z, k)$ at the $k$-th step, we can use the gradient decent updating for the proposal of next-layer node embeddings $\tilde{\mathbf Z}^{(k+1)}$ via (assuming $\frac{\alpha}{2}$ as the step size of gradient descent)
\begin{equation}\label{eqn-iter-unfolding}
\begin{split}
    \tilde{\mathbf Z}^{(k+1)} &= \mathbf Z^{(k)} - \frac{\alpha}{2} \left.\frac{\partial E(\mathbf Z, k)}{ \partial \mathbf Z}\right|_{\mathbf Z = \mathbf Z^{(k)}} \\
    & = \mathbf Z^{(k)} - \alpha\left(  \lambda (\tilde{\mathbf D} - \mathbf S) \mathbf Z^{(k)}  + \mathbf Z^{(k)} - \mathbf Z^{(k)}  \right) \\
    & = \mathbf Z^{(k)} - \alpha\lambda(\tilde{\mathbf D} - \mathbf S) \mathbf Z^{(k)}.
\end{split}
\end{equation}
Since $\lambda_1$ is the largest eigenvalue of $\mathbf \Delta = \tilde{\mathbf D} - \mathbf S$, the energy function Eqn.~\ref{eqn-proof-energy} has $\lambda_1$-Lipschitz continuous gradients 
w.r.t. $\mathbf Z$. According to the convergence theorem of the gradient descent, the iteration Eqn.~\ref{eqn-iter-unfolding} will converge on condition that $\alpha\lambda \leq \frac{1}{\lambda_1}$.

By letting $\alpha'=\alpha\lambda$ to combine two parameters as one, we have the following updating rule for next-layer node embeddings:
\begin{equation}\label{eqn-iter-unfolding2}
    \tilde{\mathbf Z}^{(k+1)} = (\mathbf I - \alpha' \tilde{\mathbf D}) \mathbf Z^{(k)} + \alpha' \mathbf S\mathbf Z^{(k)}. 
\end{equation}

One can notice that Eqn.~\ref{eqn-iter-unfolding2} shares similar forms as the numerical iteration Eqn.~\ref{eqn-diffuse-iter} for the PDE diffusion system, in particular if we write Eqn.~\ref{eqn-diffuse-iter} as a matrix form:
\begin{equation}\label{eqn-iter-pde}
    \mathbf Z^{(k+1)} = \left ( \mathbf I - \tau \tilde{\mathbf D}^{(k)} \right ) \mathbf Z^{(k)} + \tau \mathbf S^{(k)} \mathbf Z^{(k)}, 
\end{equation}
where $\tilde{\mathbf D}^{(k)}$ is the degree matrix associated with $\mathbf S^{(k)}$. On top of this,
we can see that the effect of Eqn.~\ref{eqn-iter-pde} is the same as Eqn.~\ref{eqn-iter-unfolding2}. In particular, the next-layer updated embedding $\mathbf Z^{(k+1)}$ equals to the proposal $\tilde{\mathbf Z}^{(k+1)}$ yielded by the one-step gradient descent, on condition that we let $\tau = \alpha'$ and $\mathbf S^{(k)} = \mathbf S$.
Thereby, we have proven by construction that a one-layer numerical iteration by the explicit Euler scheme, specifically shown by Eqn.~\ref{eqn-iter-unfolding2} is equivalent to a one-step gradient descent on the energy Eqn.~\ref{eqn-energy-gcn}. We thus have the result $E(\mathbf Z^{(k+1)}, k) \leq E(\mathbf Z^{(k)}, k)$, with equality if and only if $\mathbf Z^{(k)}$ is a stationary point of $E(\mathbf Z, k)$. 

Pushing further, we notice that for any fixed $\mathbf Z$, $E(\mathbf Z, k) = \|\mathbf Z - \mathbf Z^{(k)}\|^2_{\mathcal F} + \lambda \sum_{i, j} s_{ij}\|\mathbf z_i - \mathbf z_j\|^2_2$ becomes a function of $k$ and its optimum is achieved if and only if $\mathbf Z^{(k)} = \mathbf Z$. Such a fact yields that $E(\mathbf Z^{(k)}, k) \leq E(\mathbf Z^{(k)}, k-1)$. The result of the theorem follows by noting that 
\begin{equation}
    E(\mathbf Z^{(k+1)}, k) \leq E(\mathbf Z^{(k)}, k) \leq E(\mathbf Z^{(k)}, k-1).
\end{equation}

\subsection{Proof for Corollary~\ref{corollary-gcn-cont}}
    
Similar to the diffusion equation, we can use the explicit scheme involving finite differences with step size $\alpha$, i.e., $\frac{\partial \mathbf z_i(t)}{\partial t} \approx \frac{\mathbf z_i^{(k+1)} - \mathbf z_i^{(k)}}{\alpha}$, for converting the gradient flows into numerical iterations $\mathbf z_i^{(k+1)} = \mathbf z_i^{(k)} - \alpha \nabla_{\mathbf z_i} E(\mathbf Z, t)  $. The following proof can be extended from that of Theorem~\ref{thm-gcn} and setting $\tau$ and $\alpha$ to be infinitesimal.

\subsection{Proof for Corollary~\ref{cora-conv}}

According to the proof of Theorem~\ref{thm-gcn}, we know that on condition of $0<\tau< \frac{1}{\lambda_1}$, the numerical iteration of the explicit scheme (i.e., Eqn.~\ref{eqn-diffuse-iter}) contributes to a descent step on the energy $E(\mathbf Z, k)$ defined by Eqn.~\ref{eqn-energy-gcn}. Since the energy function is convex with a unique global optimum and has Lipschitz continuous gradients, the diffusion-induced iterations would decrease the energy to the converged point $\lim_{k\rightarrow \infty} E(\mathbf Z^{(k)}, k) = 0$ with a sufficient number of feed-forward steps.


\subsection{Proof for Proposition~\ref{prop-bound}}
    
We are to analyze the relationship between the energy at two consecutive layers $E(\mathbf Z^{(k+1)}, k)$ and $E(\mathbf Z^{(k)}, k-1)$, as the diffusion evolves with $\mathbf Z^{(k+1)} = (\mathbf I - \tau \tilde{\mathbf D})\mathbf Z^{(k)} + \tau \mathbf S \mathbf Z^{(k)}$. By letting $\mathbf B = \mathbf I - \tau(\tilde{\mathbf D} - \mathbf S)$, we have the following result:
\begin{equation}
    \begin{split}
        E(\mathbf Z^{(k+1)}, k) & = \|\mathbf Z^{(k+1)} - \mathbf Z^{(k)}\|_{\mathcal F}^2 + \lambda \mbox{tr}\left((\mathbf Z^{(k+1)})^\top (\tilde{\mathbf D} - \mathbf S) \mathbf Z^{(k+1)} \right ) \\
        & = \|\mathbf B (\mathbf Z^{(k)} - \mathbf Z^{(k-1)})\|_{\mathcal F}^2 + \lambda \mbox{tr}\left((\mathbf B\mathbf Z^{(k)})^\top (\tilde{\mathbf D} - \mathbf S) \mathbf B \mathbf Z^{(k)} \right ) \\
        & \leq (1 - \tau \lambda_2)^2 \|\mathbf Z^{(k)} - \mathbf Z^{(k-1)}\|_{\mathcal F}^2 + \lambda \mbox{tr}\left((\mathbf B\mathbf Z^{(k)})^\top (\tilde{\mathbf D} - \mathbf S) \mathbf B \mathbf Z^{(k)} \right ) \\
        & \leq (1 - \tau \lambda_2)^2 \|\mathbf Z^{(k)} - \mathbf Z^{(k-1)}\|_{\mathcal F}^2 + \lambda (1 - \tau \lambda_2)^2 \mbox{tr}\left((\mathbf Z^{(k)})^\top (\tilde{\mathbf D} - \mathbf S) \mathbf Z^{(k)} \right ) \\
        & = (1 - \tau \lambda_2)^2 E(\mathbf Z^{(k)}, k-1),
    \end{split}
\end{equation}
where $\lambda_2$ is the smallest eigenvalue of the positive semidefinite matrix $\mathbf \Delta = \tilde{\mathbf D} - \mathbf S$ whose eigenvalues are all non-negative. Similarly, we can derive the lower bound:
\begin{equation}
    \begin{split}
        E(\mathbf Z^{(k+1)}, k) & = \|\mathbf Z^{(k+1)} - \mathbf Z^{(k)}\|_{\mathcal F}^2 + \lambda \mbox{tr}\left((\mathbf Z^{(k+1)})^\top (\tilde{\mathbf D} - \mathbf S) \mathbf Z^{(k+1)} \right ) \\
        & \geq (1 - \tau \lambda_1)^2 \|\mathbf Z^{(k)} - \mathbf Z^{(k-1)}\|_{\mathcal F}^2 + \lambda (1 - \tau \lambda_1)^2 \mbox{tr}\left((\mathbf Z^{(k)})^\top (\tilde{\mathbf D} - \mathbf S) \mathbf Z^{(k)} \right ) \\
        & = (1 - \tau \lambda_1)^2 E(\mathbf Z^{(k)}, k-1),
    \end{split}
\end{equation}
where $\lambda_1$ is the largest eigenvalue of $\mathbf \Delta$ and the second step is based on the fact that $1 - \tau \lambda_1 \geq 1 - \frac{1}{\lambda_1}\lambda_1 = 0$.

\subsection{Proof for Proposition~\ref{prop-convergence}}
    We follow the similar reasoning line as that of Theorem~\ref{thm-gcn}. Similarly, the quadratic energy defined by Eqn.~\ref{eqn-energy-appnp} can be written as $E(\mathbf Z, k) = \|\mathbf Z - (\mathbf Z^{(k)} + \eta \mathbf H)\|^2_{\mathcal F} + \lambda \mbox{tr}(\mathbf Z^\top (\tilde{\mathbf D} - \mathbf S)\mathbf Z)$ and the gradient w.r.t. $\mathbf Z$ can be computed as
\begin{equation}\label{eqn-proof-appnp-grad}
\begin{split}
    \frac{1}{2} \left.\frac{\partial E(\mathbf Z, k)}{ \partial \mathbf Z}\right|_{\mathbf Z = \mathbf Z^{(k)}}
    & =  \lambda (\tilde{\mathbf D} - \mathbf S) \mathbf Z^{(k)}  + \mathbf Z^{(k)} - (\mathbf Z^{(k)} + \eta \mathbf H)  \\
    & = \lambda(\tilde{\mathbf D} - \mathbf S) \mathbf Z^{(k)} - \eta \mathbf H.
\end{split}
\end{equation}
Then using one-step gradient descent with step size $\frac{\alpha}{2}$ would yield the updating rule (assuming $\alpha'=\alpha\lambda$ and $\eta' = \alpha \eta$):
\begin{equation}\label{eqn-iter-unfolding-appnp2}
    \tilde{\mathbf Z}^{(k+1)} = (\mathbf I - \alpha' \tilde{\mathbf D}) \mathbf Z^{(k)} + \alpha' \mathbf S\mathbf Z^{(k)} + \eta' \mathbf H. 
\end{equation}
The sufficient condition for the convergence of the above iteration is also $\alpha \lambda \leq \frac{1}{\lambda_1}$.
On the other hand, using explicit scheme for solving Eqn.~\ref{eqn-diffuse-source} would yield the feed-forward iteration in the matrix form as
\begin{equation}\label{eqn-proof-appnp-update}
    \mathbf Z^{(k+1)} = \left ( \mathbf I - \tau \tilde{\mathbf D}^{(k)} \right ) \mathbf Z^{(k)} + \tau \mathbf S^{(k)} \mathbf Z^{(k)} + \tau \beta \mathbf H. 
\end{equation}
We can see that Eqn.~\ref{eqn-proof-appnp-update} would become the same as Eqn.~\ref{eqn-iter-unfolding-appnp2} when we let $\tau = \alpha'$ and $\tau \beta = \eta'$, thus contributing to a descent step on the energy $E(\mathbf Z, k)$. 
We therefore obtain $E(\mathbf Z^{(k+1)}, k) \leq E(\mathbf Z^{(k)}, k)$. Furthermore, we have $E(\mathbf Z^{(k)}, k) \leq E(\mathbf Z^{(k)}, k-1)$ since for any fixed $\mathbf Z$, $E(\mathbf Z, k)$ (treated as a function of $k$) achieves the optimum if and only if $\mathbf Z = \mathbf Z^{(k)}$. The proof is concluded by combining the above results, i.e., $E(\mathbf Z^{(k+1)}, k) \leq E(\mathbf Z^{(k)}, k) \leq E(\mathbf Z^{(k)}, k-1)$.

\subsection{Proof for Proposition~\ref{prop-energy}}
The proof of the proposition follows the principles of convex analysis and Fenchel duality~\citep{convex-1970}. For any concave and non-decreasing function $\rho: \mathbb R^+\rightarrow \mathbb R$, one can express it as the variational decomposition
\begin{equation}\label{eqn-proof-convex}
    \rho(z^2) = \min_{\omega \geq 0} [\omega z^2 - \tilde \rho(\omega)] \geq \omega z^2 - \tilde \rho(\omega),
\end{equation}
where $\omega$ is a variational parameter and $\tilde \rho$ is the concave conjugate of $\rho$.
Eqn.~\ref{eqn-proof-convex} essentially defines $\rho(z^2)$ as the minimal envelope of a series of quadratic bounds $\omega z^2 - \tilde \rho(\omega)$ defined by a different values of $\omega\geq 0$ and the upper bound is given for a fixed $\omega$ when removing the minimization operator. Based on this, we obtain the result of Eqn.~\ref{eqn-prop1-bound}. In terms of the sufficient and necessary condition for equality, we note that for any optimal $\omega^*$ we have
\begin{equation}
    \omega^* z^2 - \tilde \rho(\omega^*) = \rho(z^2),
\end{equation}
which is tangent to $\rho$ at $z^2$ and $\omega^* = \frac{\partial \delta(z^2)}{\partial z^2}$. We thus obtain the result of Eqn.~\ref{eqn-prop1-bound}.

\subsection{Proof for Theorem~\ref{thm-main}}
    We initiate the proof by construction. We aim to construct a descent step on the non-convex energy target Eqn.~\ref{eqn-energy} and show its equivalence to the one-step diffusion iteration Eqn.~\ref{eqn-diffuse-iter}. Due to the penalty function $\delta$ in Eqn.~\ref{eqn-energy}, it can be difficult in directly minimizing the energy by gradient descent. However, according to Proposition~\ref{prop-energy}, we can minimize the upper bound surrogate Eqn.~\ref{eqn-prop1-bound} and it becomes equivalent to a minimization of the original energy on condition that the variational parameters are given by $\omega_{ij}=\left. \frac{\partial \delta(z^2)}{\partial z^2} \right |_{z = \|\mathbf z_i^{(k)} - \mathbf z_j^{(k)}\|_2}$. Then with a one-step gradient decent of Eqn.~\ref{eqn-prop1-bound}, the proposal for the next-layer node embeddings would be (assuming the step size to be $\frac{\alpha}{2}$)
\begin{equation}\label{eqn-iter-unfolding-2}
\begin{split}
    \tilde{\mathbf Z}^{(k+1)} &= \mathbf Z^{(k)} - \frac{\alpha}{2} \left.\frac{\partial \tilde E(\mathbf Z, k;\mathbf \Omega^{(k)}, \tilde\delta)}{ \partial \mathbf Z}\right|_{\mathbf Z = \mathbf Z^{(k)}} \\
    & = \mathbf Z^{(k)} - \alpha\left(  \lambda (\overline{\mathbf D}^{(k)} - \mathbf \Omega^{(k)}) \mathbf Z^{(k)}  + \mathbf Z^{(k)} - \mathbf Z^{(k)}  \right) \\
    & = \mathbf Z^{(k)} - \alpha'(\overline{\mathbf D}^{(k)} - \mathbf \Omega^{(k)}) \mathbf Z^{(k)}
\end{split}
\end{equation}
where $\mathbf \Omega^{(k)} = \{\omega_{ij}^{(k)}\}_{N\times N}$, $\overline{\mathbf D}^{(k)}$ denotes the diagonal degree matrix associated with $\mathbf \Omega^{(k)}$ and we introduce $\alpha'=\alpha\lambda$ to combine two parameters as one. Common practice to accelerate convergence adopts a positive definite preconditioner term, e.g., $(\overline{\mathbf D}^{(k)})^{-1}$, to re-scale the updating gradient and the final updating form becomes
\begin{equation}\label{eqn-iter2-unfolding2}
    \tilde{\mathbf Z}^{(k+1)} = (1 - \alpha') \mathbf Z^{(k)} + \alpha' (\overline{\mathbf D}^{(k)})^{-1}  \mathbf \Omega^{(k)}\mathbf Z^{(k)}. 
\end{equation}

The above iteration will converge once $0<\alpha' \leq 1$. One can notice that Eqn.~\ref{eqn-iter2-unfolding2} shares the similar form as the numerical iteration Eqn.~\ref{eqn-diffuse-iter} for the PDE diffusion system, in particular if we re-write Eqn.~\ref{eqn-diffuse-iter} in a matrix form:
\begin{equation}\label{eqn-iter-pde2}
    \mathbf Z^{(k+1)} = \left ( 1 - \tau \tilde{\mathbf D}^{(k)} \right ) \mathbf Z^{(k)} + \tau \mathbf S^{(k)} \mathbf Z^{(k)}. 
\end{equation}
where $\tilde{\mathbf D}^{(k)}$ is the diagonal degree matrix associated with $\mathbf S^{(k)}$. Pushing further,
we can see that the effect of Eqn.~\ref{eqn-iter-pde2} is the same as Eqn.~\ref{eqn-iter2-unfolding2} when we let $\tau = \alpha'$ and $\mathbf S^{(k)} = (\overline{\mathbf D}^{(k)})^{-1}\mathbf \Omega^{(k)}$ (notice that since $\mathbf S^{(k)}$ is row-normalized, we have $\sum_{j\in V} s_{ij}^{(k)} = 1$ and $\tilde{\mathbf D}^{(k)} = \mathbf I$).  

Thereby, we have proven by construction that a one-step numerical iteration by the explicit Euler scheme, specifically shown by Eqn.~\ref{eqn-iter2-unfolding2} is equivalent to a one-step gradient descent on the surrogate Eqn.~\ref{eqn-proof-convex} which further equals to the original energy Eqn.~\ref{eqn-energy} with the coupling matrix given by Eqn.~\ref{eqn-optimal-diffuse}. We thus have the result 
\begin{equation}\label{eqn-proof-e1}
    E(\mathbf Z^{(k+1)}, k; \delta) \leq E(\mathbf Z^{(k)}, k; \delta).
\end{equation}
Also, we notice that for any fixed $\mathbf Z$, $E(\mathbf Z, k; \delta) = \|\mathbf Z - \mathbf Z^{(k)}\|^2_{\mathcal F} + \lambda \sum_{ij} \rho(\|\mathbf z_i - \mathbf z_j\|^2_2)$ becomes a function of $k$ and its optimum is achieved if and only if $\mathbf Z^{(k)} = \mathbf Z$. Such a fact yields that 
\begin{equation}\label{eqn-proof-e2}
E(\mathbf Z^{(k)}, k; \delta) \leq E(\mathbf Z^{(k)}, k-1; \delta).    
\end{equation}
The result of the main theorem follows by combing the results of Eqn.~\ref{eqn-proof-e1} and \ref{eqn-proof-e2}.

\subsection{Proof for Corollary~\ref{coro-main}}
    For the continuous system, we can similarly leverage the result of Proposition~\ref{prop-energy} to obtain the upper bound surrogate of Eqn.~\ref{eqn-energy-con} (assuming that $\mathbf \Omega(t) = [\omega_{ij}(t)]_{i,j\in \mathcal V}$ is a function of time)
    \begin{equation}\label{eqn-energy-su-con}
        \tilde E(\mathbf Z, t; \mathbf\Omega, \tilde \delta) = \|\mathbf Z - \mathbf Z(t)\|_{\mathcal F}^2 + \lambda \left[\sum_{i,j} \omega_{ij}(t) \|\mathbf z_i - \mathbf z_j\|_2^2 - \tilde \delta(\omega_{ij}(t))\right ],
    \end{equation}
    and the equality holds when $\omega_{ij}(t) = \left. \frac{\partial \delta(z^2)}{\partial z^2} \right |_{z = \|\mathbf z_i(t) - \mathbf z_j(t)\|_2}$. Therefore, for any given $t$, the descent step on Eqn.~\ref{eqn-energy-su-con} is equivalent to that of Eqn.~\ref{eqn-energy-con} with the $(i, j)$-th entry of the coupling matrix $s_{ij}(t)$ satisfying the condition at time $t$: 
    \begin{equation}
        \nabla_{\mathbf z_i} E(\mathbf Z, t) = \nabla_{\mathbf z_i} \tilde E(\mathbf Z, t; \mathbf\Omega, \tilde \delta), \quad \mbox{on condition that}~~~ s_{ij}(t) = \left. \frac{\partial \delta(z^2)}{\partial z^2} \right |_{z = \|\mathbf z_i(t) - \mathbf z_j(t)\|_2}, \forall i, j \in \mathcal V.
    \end{equation}
    The proof can be concluded by noting that the gradient of Eqn.~\ref{eqn-energy-su-con} equals to the right-hand-side of Eqn.~\ref{eqn-diffuse2}:
    \begin{equation}
        \left. \frac{\partial \tilde E(\mathbf Z, t; \mathbf\Omega, \tilde \delta)}{\partial \mathbf z_i} \right |_{\mathbf z_i = \mathbf z_i(t)} = \sum_{j\in \mathcal V} s_{ij}(t) (\mathbf z_j(t) - \mathbf z_i(t))).
    \end{equation}

\subsection{Proof for Proposition~\ref{prop-feattrans}}
    The starting point of the proof follows that of Theorem~\ref{thm-main}. First, we consider one-step gradient descent on the surrogate energy of Eqn.~\ref{eqn-energy-w} without the penalty term:
    \begin{equation}
        \tilde E(\mathbf Z, k;\mathbf \Omega, \tilde \delta, h^{(k)}) = \|\mathbf Z - h^{(k)}(\mathbf Z^{(k)}) \|_{\mathcal F}^2 + \lambda \left[\sum_{i,j} \omega_{ij} \|\mathbf z_i - \mathbf z_j\|_2^2 - \tilde \delta(\omega_{ij})\right ].    
    \end{equation}
    Replacing the evaluation at $\mathbf Z = \mathbf Z^{(k)}$ in Eqn.~\ref{eqn-iter-unfolding-2} by $\mathbf Z = h^{(k)}(\mathbf Z^{(k)})$, we can obtain the update of node embeddings
    \begin{equation}
    \begin{split}
        \tilde{\mathbf Z}^{(k+1)} & = h^{(k)}(\mathbf Z^{(k)}) - \alpha \left.\frac{\partial \tilde E(\mathbf Z, k;\mathbf \Omega^{(k)}, \tilde\delta)}{ \partial \mathbf Z}\right|_{\mathbf Z = h^{(k)}(\mathbf Z^{(k)})} \\
        & = h^{(k)}(\mathbf Z^{(k)}) - \alpha'(\overline{\mathbf D}^{(k)} - \mathbf \Omega^{(k)}) h^{(k)}(\mathbf Z^{(k)}).
    \end{split}
    \end{equation}
    And, inserting the preconditioner term $(\overline{\mathbf D}^{(k)})^{-1}$ before the gradient part, we have
    \begin{equation}
        \tilde{\mathbf Z}^{(k+1)} = (1 - \alpha') h^{(k)}(\mathbf Z^{(k)}) + \alpha' (\overline{\mathbf D}^{(k)})^{-1}  \mathbf \Omega^{(k)} h^{(k)}(\mathbf Z^{(k)}).
    \end{equation}
    Then we take into account the penalty term $\psi$ in Eqn.~\ref{eqn-energy-w} and associate it with the non-linear activation $\sigma$ in Eqn.~\ref{eqn-diff-iter-w}. 
    The latter in fact can be treated as a proximal operator (which projects the output into the feasible region induced by the penalty) and the gradient descent step can be further modified to add a proximal operator:
    \begin{equation}\label{eqn-proof-proximal}
        \tilde{\mathbf Z}^{(k+1)} = \mbox{Prox}_{\psi} \left ((1 - \alpha') h^{(k)}(\mathbf Z^{(k)}) + \alpha' (\overline{\mathbf D}^{(k)})^{-1}  \mathbf \Omega^{(k)} h^{(k)}\mathbf Z^{(k)}) \right ),
    \end{equation}
    where $\mbox{Prox}_{\psi} (\mathbf z) = \arg\min_{\mathbf x} \|\mathbf x - \mathbf z\|_2^2 + \psi(\mathbf x)$. The above updating rule corresponds to a proximial gradient descent step which guarantees a strict minimization for the energy in Eqn.~\ref{eqn-energy-w}. In particular, if one considers ReLU activation, the proximal operator will be $\mbox{Prox}_{\psi} (\mathbf z) = \mbox{ReLU}(\mathbf z) = \max(\mathbf 0, \mathbf z)$ and the penalty function can be $\psi(\mathbf z) = \sum_{k} \mathbb I_{\infty}[z_k < 0]$, where $\mathbb I_{\infty}$ is an indicator function that assigns infinite penalty to negative value. The following proof for this proposition can re-use that of Theorem~\ref{thm-main} (the reasoning line after Eqn.~\ref{eqn-iter2-unfolding2}) by noting the connection between Eqn.~\ref{eqn-diff-iter-w} and Eqn.~\ref{eqn-proof-proximal}.

\subsection{Proof for Proposition~\ref{prop-trans-oversmooth}}
    Since the diffusivity $\mathbf S^{(k)}$ is row-normalized according to the definition of Eqn.~\ref{eqn-optimal-diffuse}, then using explicit scheme for solving the diffusion equation Eqn.~\ref{eqn-diffuse-source} would induce the feed-forward iteration (in the matrix form):
    \begin{equation}\label{eqn-proof-appnp-update2}
        \mathbf Z^{(k+1)} = ( 1 - \tau  ) \mathbf Z^{(k)} + \tau \mathbf S^{(k)} \mathbf Z^{(k)} + \tau \beta \mathbf H. 
    \end{equation}
    Then the proof can be concluded by following the reasoning line of Theorem~\ref{thm-main} and noting the equivalence between Eqn.~\ref{eqn-proof-appnp-update2} and a gradient descent step on the corresponding surrogate energy for Eqn.~\ref{eqn-energy2-appnp} (assuming $\alpha'=\alpha\lambda$ and $\eta' = \alpha \eta$):
\begin{equation}\label{eqn-iter-unfolding-appnp3}
    \tilde{\mathbf Z}^{(k+1)} = (\mathbf I - \alpha' ) \mathbf Z^{(k)} + \alpha' \mathbf S\mathbf Z^{(k)} + \eta' \mathbf H.
\end{equation}
The sufficient condition for the convergence of the above iteration is $\tau = \alpha \lambda \leq 1$.

\section{Different Energy Forms}\label{appx-inst}

We present more detailed illustration for the choices of $f$ and specific energy function forms in Eq.~\ref{eqn-energy}.

\begin{figure}[t!]
\centering
\subfigure[\model-s: $\delta(z^2) = 2z^2 - \frac{1}{4}z^4$]{
\begin{minipage}[t]{0.48\linewidth}
\centering
\label{fig-sparse}
\includegraphics[width=0.98\textwidth,angle=0]{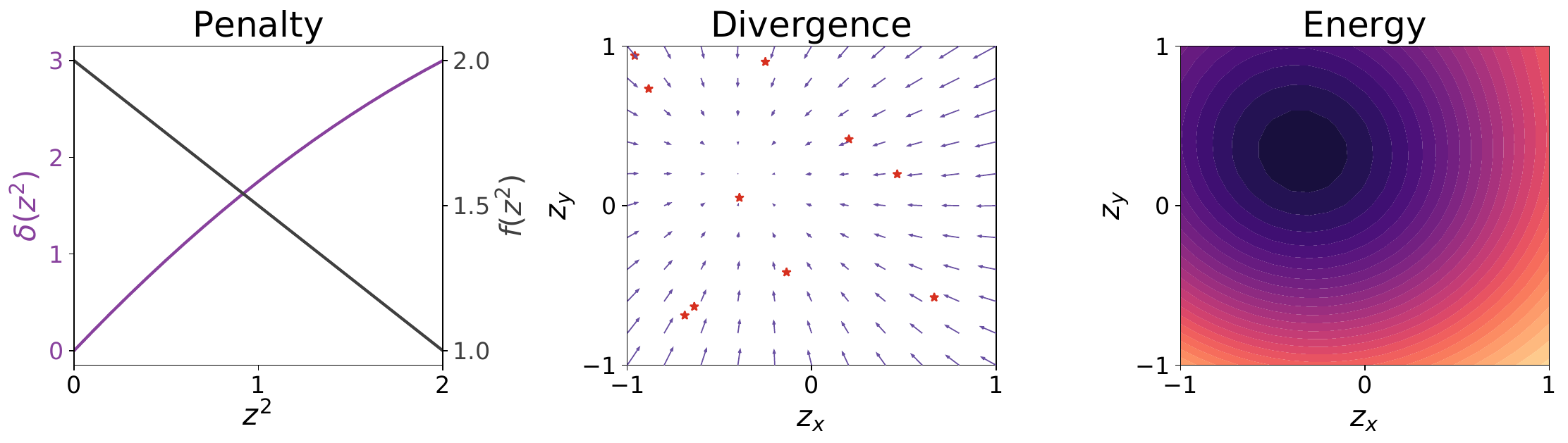}
\end{minipage}%
}%
\hspace{5pt}
\subfigure[\model-a: $\delta(z^2) = z^2 - 2 \log (e^{z^2/2-1} + 1)$]{
\begin{minipage}[t]{0.48\linewidth}
\centering
\label{fig-attn-a}
\includegraphics[width=0.98\textwidth,angle=0]{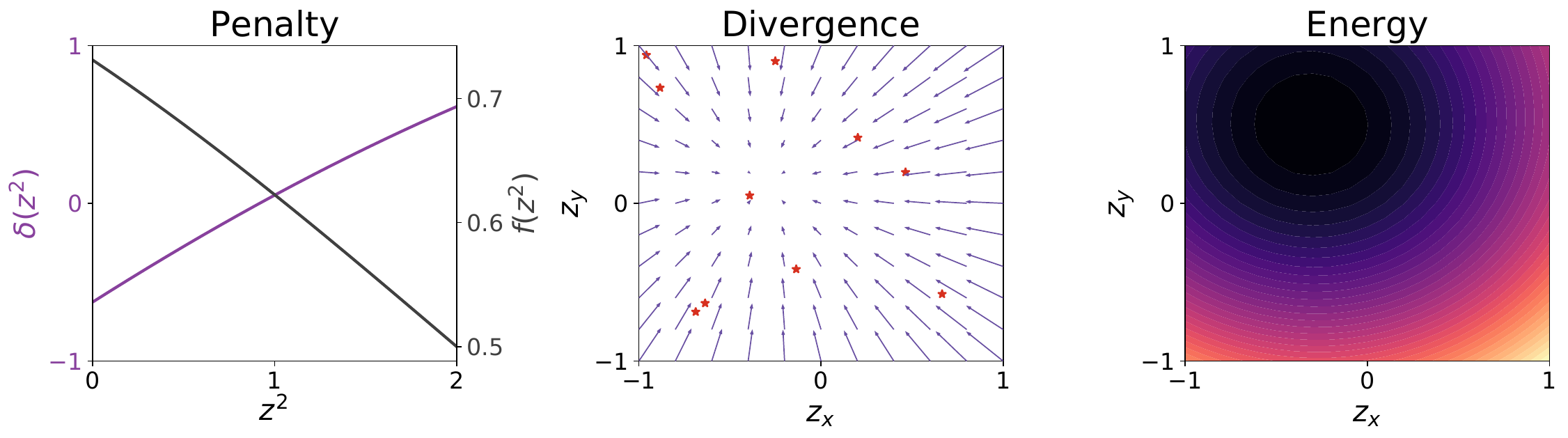}
\end{minipage}%
}
\caption{Plot of penalty curves $\delta(z^2)$ and $f(z^2)=\frac{\partial \delta(z^2)}{\partial z^2}$, divergence field (produced by 10 randomly generated instances marked as red stars) and cross-section energy field of an individual.}
\label{fig-endiff-vis}
\end{figure}

\textbf{Simple Diffusivity Model.} As discussed in Section~\ref{sec-trans-attn-difformer}, the simple model assumes $f(z^2) = 2 - \frac{1}{2}z^2$ that corresponds to $g(x) = 1+x$, where we define $z = \|\mathbf z_i - \mathbf z_j\|_2$ and $x = \mathbf z_i^\top \mathbf z_j$. The corresponding penalty function $\delta$ whose first-order derivative is $f$ would be $\delta(z^2) = 2z^2 - \frac{1}{4}z^4$. We plot the penalty function curves in Figure~\ref{fig-endiff-vis}(a). As we can see, the $f$ is a non-negative, decreasing function of $z^2$, which implies that the $\delta$ satisfies the non-decreasing and concavity properties to guarantee a valid regularized energy function. Also, in Figure~\ref{fig-endiff-vis}(a) we present the divergence field produced by 10 randomly generated instances (marked as red stars) and the cross-section energy field of one instance.  

\textbf{Advanced Diffusivity Model.} The diffusivity model defines $f(z^2) = \frac{1}{1+e^{z^2/2 - 1}}$ with $g(x) = \frac{1}{1+\exp(-x)}$, and the corresponding penalty function $\delta(z^2) = z^2 - 2 \log (e^{z^2/2-1} + 1)$. The penalty function curves, divergence field and energy field are shown in Figure~\ref{fig-endiff-vis}(b).

\section{Details of DIFFormer Model}\label{appx-alg}

In this section, we present the details for the feed-forward computation of DIFFormer. 

\subsection{DIFFormer's Feed-forward with A Matrix View}

\textbf{Input Layer.} For input data $\mathbf X = \{\mathbf x_i\}_{i=1}^N \in \mathbb R^{N\times D}$ where $\mathbf x_i$ denotes the $D$-dimensional input features of the $i$-th instance, we first use a shallow fully-connected layer to convert it into a $d$-dimensional embedding in the latent space:
\begin{equation}
    \mathbf Z = \sigma\left (\mbox{LayerNorm}(\mathbf W_I \mathbf X + \mathbf b_I) \right ),
\end{equation}
where $\mathbf W_I \in \mathbb R^{d\times D}$ and $\mathbf b_I \in \mathbb R^d$ are trainable parameters, and $\sigma$ is a non-linear activation (i.e., ReLU). Then the node embeddings $\mathbf Z$ will be used for feature propagation with our diffusion-induced Transformer model by letting $\mathbf Z^{(0)} = \mathbf Z$ as the initial states.

\textbf{Propagation Layer.} The initial embeddings $\mathbf Z^{(0)}$ will be transformed into $\mathbf Z^{(1)}, \cdots, \mathbf Z^{(L)}$ with $L$ layers of propagation. We next illustrate one-layer propagation from $\mathbf Z^{(k)}$ to $\mathbf Z^{(k+1)}$. We use the superscript $(k, h)$ to denote the $k$-th layer and the $h$-th head:
\begin{equation}
    \mathbf K^{(k, h)} =  \mathbf W_{K}^{(k, h)}\mathbf Z^{(k)}, \quad \mathbf Q^{(k, h)} =  \mathbf W_{Q}^{(k, h)}\mathbf Z^{(k)}, \quad
    \mathbf V^{(k, h)} =  \mathbf W_{V}^{(k, h)}\mathbf Z^{(k)},
\end{equation}
where $\mathbf W_{K}^{(k, h)}\in \mathbb R^{d\times d}, \mathbf W_{Q}^{(k, h)}\in \mathbb R^{d\times d}, \mathbf W_{V}^{(k, h)}\in \mathbb R^{d\times d}$ are trainable parameters of the $h$-th head at the $k$-th layer. We then adopt L2 normalization for the key and query vectors to constrain the vector norm:
\begin{equation}
    \tilde{\mathbf K}^{(k, h)} = \left [\frac{\mathbf K^{(k, h)}_{i}}{\|\mathbf K^{(k, h)}_{i}\|_2} \right ]_{i=1}^N, \quad \tilde{\mathbf Q}^{(k, h)} = \left [\frac{\mathbf Q^{(k, h)}_{i}}{\|\mathbf Q^{(k, h)}_{i}\|_2}\right ]_{i=1}^N,
\end{equation}
where $\mathbf K^{(k,h)}_{i}$ denotes the $i$-th row vector of $\mathbf K^{(k, h)}\in \mathbb R^{N\times d}$. Then the transformed embeddings will be fed into the all-pair propagation unit of DIFFormer-s or DIFFormer-a.
\begin{itemize}
    \item For DIFFormer-s: the all-pair propagation of the $h$-th head is achieved by
    \begin{equation}
        \mathbf R^{(k, h)} = \mbox{diag}^{-1}\left (N + \tilde{\mathbf Q}^{(k, h)} \left ((\tilde{\mathbf K}^{(k, h)} )^\top \mathbf 1 \right ) \right ), 
    \end{equation}
    \begin{equation}
    \mathbf P^{(k, h)} = \mathbf R^{(k, h)} \left [\mathbf 1 \left(\mathbf 1^\top \mathbf V^{(k, h)} \right ) + \tilde{\mathbf Q}^{(k, h)} \left ((\tilde{\mathbf K}^{(k, h)}  )^\top \mathbf V^{(k, h)} \right ) \right ],
    \end{equation}
    where $\mathbf 1_{N\times 1}$ is an all-one vector. The above computation only requires linear complexity w.r.t. $N$ since the bottleneck computation lies in $\tilde{\mathbf Q}^{(k, h)} \left ((\tilde{\mathbf K}^{(k, h)}  )^\top \mathbf V^{(k, h)} \right )$ where the two matrix products both require $O(Nd^2)$.
    \item For DIFFormer-a: we need to compute the all-pair similarity before aggregating the results
    \begin{equation}
        \tilde{\mathbf A}^{(k, h)} = \mbox{Sigmoid}\left (\tilde{\mathbf Q}^{(k, h)} (\tilde{\mathbf K}^{(k, h)})^\top \right ),
    \end{equation}
    \begin{equation}
        \mathbf R^{(k, h)} = \mbox{diag}^{-1} \left( \tilde{\mathbf A}^{(k, h)} \mathbf 1 \right ),
    \end{equation}
    \begin{equation}
        \mathbf P^{(k, h)} = \mathbf R^{(k, h)} \tilde{\mathbf A}^{(k, h)} \mathbf V^{(k, h)}. 
    \end{equation}
    The above computation requires $O(Nd^2+N^2d)$ due to the explicit computation of the $N\times N$ matrix $\tilde{\mathbf A}^{(k, h)}$.
\end{itemize}
If using input graphs, we add the updated embeddings of GCN-based propagation to the all-pair propagation's ones:
\begin{equation}
    \overline{\mathbf P}^{(k, h)} = \mathbf P^{(k, h)} + \mathbf D^{-\frac{1}{2}} \mathbf A \mathbf D^{-\frac{1}{2}} \mathbf V^{(k, h)},
\end{equation}
where $\mathbf A$ is the input graph and $\mathbf D$ denotes its corresponding diagonal degree matrix.

We then average the propagated results of multiple heads:
\begin{equation}
    \overline{\mathbf P}^{(k)} = \frac{1}{H} \sum_{h=1}^H \overline{\mathbf P}^{(k, h)}.
\end{equation}
The next-layer embeddings will be updated by
\begin{equation}
    \mathbf Z^{(k+1)} = \sigma' \left ( \mbox{LayerNorm}\left ( \tau \overline{\mathbf P}^{(k)} + (1 - \tau) \mathbf Z^{(k)}\right ) \right ),
\end{equation}
where $\sigma'$ can be identity mapping or non-linear activation (e.g., ReLU). 

\textbf{Output Layer.} After $K$ layers of propagation, we then use a shallow fully-connected layer to output the predicted logits:
\begin{equation}
    \hat {\mathbf Y} = \mathbf Z^{(K)} \mathbf W_O + \mathbf b_O,
\end{equation}
where $\mathbf W_O\in \mathbb R^{d\times C}$ and $\mathbf b_O \in \mathbb R^{C}$ are trainable parameters, and $C$ denotes the number of classes. And, the predicted logits $\hat {\mathbf Y}$ will be used for computing a loss of the form $l(\hat {\mathbf Y}, \mathbf Y)$ where $l$ can be cross-entropy for classification or mean square error for regression.

\section{Dataset Information}\label{appx-dataset}

In this section, we present the detailed information for all the experimental datasets, the pre-processing and evaluation protocol used in Section~\ref{sec-exp}. 

\begin{table*}[htbp]
    \centering
    \caption{Information for node classification datasets.}
    \label{tbl:dataset}
    \resizebox{0.75\textwidth}{!}{
    \begin{tabular}{l|ccccccc}
    \hline
    \textbf{Dataset} & \textbf{Type} & \textbf{\# Nodes} & \textbf{\# Edges} & \textbf{\# Node features} & \textbf{\# Class}  \\ 
    \hline
    Cora & Citation network & 2,708 & 5,429 & 1,433 & 7 \\
    Citeseer & Citation network & 3,327 & 4,732 & 3,703 & 6 \\
    Pubmed & Citation network & 19,717 & 44,338 & 500 & 3 \\
    Proteins & Protein interaction & 132,534 & 39,561,252 & 8  & 2 \\ 
    Pokec & Social network & 1,632,803 & 30,622,564 & 65 & 2 \\
    Chameleon & Information network & 2,277 & 31,421 & 2,325 & 5  \\
    Squirrel & Information network & 5,201 & 198,493 & 2,089 & 5 \\
    Actor & Social network & 7,600 & 30,019 & 932 & 5 \\
    \hline
    \end{tabular}
    }
\end{table*}

\subsection{Node Classification Datasets}
\texttt{Cora}, \texttt{Citeseer} and \texttt{Pubmed}~\citep{Sen08collectiveclassification} are commonly used citation networks for evaluating models on node classification tasks., These datasets are small-scale networks (with 2K$\sim$20K nodes) and the goal is to classify the topics of documents (instances) based on input features of each instance (bag-of-words representation of documents) and graph structure (citation links). 
Following the semi-supervised learning setting in \cite{GCN-vallina}, we randomly choosing 20 instances per class for training, 500/1000 instances for validation/testing for each dataset. 

\texttt{Chameleon} and \texttt{Squirrel} are both Wikipedia networks where nodes represent Wikipedia articles and edges record the mutual links between pages. Node features consist of the presence of particular nouns in the articles. The prediction target is the average
monthly traffic for the web page, and \cite{geomgcn-iclr20} converts the labels into discrete classes by grouping nodes into five categories. A recent work~\citep{heter-iclr23} identifies that the original data splits adopted by \cite{geomgcn-iclr20} introduce overlapped nodes between training and test sets, which causes the data leakage. Therefore, we follow the new data splits released by \cite{heter-iclr23} that filter out the overlapped nodes.

\texttt{OGBN-Proteins}~\citep{ogb-nips20} is a multi-task protein-protein interaction network whose goal is to predict molecule instances' property. We follow the original splitting of \cite{ogb-nips20} for evaluation. \texttt{Pokec} is a large-scale social network with features including profile information, such as geographical region, registration time, and age, for prediction on users' gender. For semi-supervised learning, we consider randomly splitting the instances into train/valid/test with 10\%/10\%/80\% ratios. 
Table~\ref{tbl:dataset} summarizes the statistics of these datasets.

\subsection{Image and Text Classification Datasets}

We evaluate our model on two image classification datasets: STL-10 and CIFAR-10. We use all 13000 images from STL-10, each of which belongs to one of ten classes. We choose 1500 images from each of 10 classes of CIFAR-10 and obtain a total of 15,000 images. For STL-10 and CIFAR-10, we randomly select 10/50/100 instances per class as training set, 1000 instances for validation and the remaining instances for testing. We first use the self-supervised approach SimCLR~\citep{chen2020simple} (that does not use labels for training) to train a ResNet-18 for extracting the feature maps as input features of instances.
We also evaluate our model on 20Newsgroup, which is a text classification dataset consisting of 9607 instances. We follow \cite{LDS-icml19} to take 10 classes from 20 Newsgroup and use words (TFIDF) with a frequency of more than 5\% as features.

\subsection{Spatial-Temporal Datasets}
The spatial-temporal datasets are from the open-source library PyTorch Geometric Temporal \citep{rozemberczki2021pytorch}, with properties and summary statistics described in Table~\ref{tab:desc_discrete}. Node features are evolving for all the datasets considered here, i.e., we have different node features for different snapshots. For each dataset, we split the snapshots into training, validation, and test sets according to a 2:2:6 ratio in order to make it more challenging and close to the real-world low-data learning setting.
In details:
\begin{itemize}
    \item \texttt{Chickenpox} describes weekly officially reported cases of chickenpox in Hungary from 2005 to 2015, whose nodes are counties and edges denote direct neighborhood relationships. Node features are lagged weekly counts of the chickenpox cases (we included 4 lags). The target is the weekly number of cases for the upcoming week (signed integers).
    \item \texttt{Covid} contains daily mobility graph between regions in England NUTS3 regions, with node features corresponding to the number of confirmed COVID-19 cases in the previous days from March to May 2020. The graph indicates how many people moved from one region to the other each day, based on Facebook Data For Good disease prevention maps. Node features correspond to the number of COVID-19 cases in the region in the past 8 days. The task is to predict the number of cases in each node after 1 day.
    \item \texttt{WikiMath} is a dataset whose nodes describe Wikipedia pages on popular mathematics topics and edges denote the links from one page to another. Node features are provided by the number of daily visits between 2019 March and 2021 March. The graph is directed and weighted. Weights represent the number of links found at the source Wikipedia page linking to the target Wikipedia page. The target is the daily user visits to the Wikipedia pages between March $16^\text{th}$ 2019 and March $15^\text{th}$ 2021 which results in 731 periods.
\end{itemize}
   
\begin{table}[htb]
\small

\centering
\caption{Properties and summary statistics of the spatial-temporal datasets used in the experiments with information about whether the graph structure is dynamic or static, meaning of node features (the same as the prediction target) and the corresponding dimension ($D$), the number of snapshots ($T$), the number of nodes ($|V|$), as well as the meaning of edges/edge weights.}\label{tab:desc_discrete}
{
\setlength{\tabcolsep}{1.5pt}
\centering
\resizebox{0.9\textwidth}{!}{
\begin{tabular}{l|cccccccc}
\hline
\textbf{Dataset}  & \textbf{Graph structure} &\textbf{Node features/targets} &$D$& \textbf{Frequency} & $T$ & $|V|$&\textbf{Edges/Edge weights} \\
\hline
Chickenpox&Static&Weekly Chickenpox Cases&4 & Weekly & 522 & 20 &Direct Neighborhoods\\
Covid  & Dynamic&Daily Covid Cases&8 & Daily & 61 & 129&Daily Mobility \\
WikiMath &Static&Daily User Visits&14&Daily & 731 & 1,068&Page Links \\
 \hline
\end{tabular}
}}
\end{table}

\subsection{Particle Property Prediction Datasets}

\texttt{ActsTrack} and \texttt{SynMol}, collected by~\cite{miao2022interpretable}, are composed of physical particles with certain physical properties as predictive targets. Each particle is a set of points with underlying structures that are unobserved yet induce latent interactions. Specifically, \texttt{ActsTrack} records the protons (nodes) colliding at the detectors and flying through a magnetic field. The prediction target is the property of $z\rightarrow\mu\mu$ decay, a measured property of a particle in the system. The positive samples contain
particle hits from both a $z\rightarrow\mu\mu$ decay and some pileup interactions, and negative samples have
only hits from pileup interactions. 
\texttt{SynMol} is a molecular dataset where each molecule is composed of a set of atoms (nodes) with physical interactions in the 3D space. The task is to predict the molecular property given by functional groups carbonyl and unbranched alkane. Since the predictive tasks for two datasets are both binary classification, we use ROC-AUC as the metric. Similar to the experiments on images and texts, we consider the evaluation with low label rates and adopt a random split with the ratio $10\%/10\%/80\%$ for training/validation/testing.

\section{Implementation Details and Hyper-parameters}\label{appx-implementation}

\subsection{Experiments in Section~\ref{sec-exp-observed}}
We use feature transformation for each layer on two large datasets and omit it for citation networks. The head number is set as 1. We set $\tau=0.5$ and incorporate the input graphs for DIFFormer. For other hyper-paramters, we adopt grid search for all the models with learning rate from $\{0.0001, 0.001, 0.01, 0.1\}$, weight decay for the Adam optimizer from $\{0, 0.0001, 0.001, 0.01, 0.1, 1.0\}$, dropout rate from $\{0, 0.2, 0.5\}$, hidden size from $\{16, 32, 64\} $, number of layers from $\{2, 4, 8, 16\}$. For evaluation, we compute the mean and standard deviation of the results with five repeating runs with different initializations. For each run, we run for a maximum of 1000 epochs and report the testing performance achieved by the epoch yielding the best performance on validation set.

\subsection{Experiments in Section~\ref{sec-exp-partial}}
We do not use feature transformation for these datasets due to their small sizes and also set $\tau=0.5$. The head number is set as 1. These spatial-temporal dynamics prediction datasets contain available graph structures, we consider both cases, using the input graphs and not, in our experiments and discuss their impact on the performance. For other hyper-parameters, we also consider grid search for all models here with learning rate from $\{0.01, 0.05, 0.005\}$, weight decay for the Adam optimizer from $\{0, 0.005\}$, dropout rate from $\{0, 0.2, 0.5\}$, and report the test mean squared error (MSE) based on the lowest validation MSE. We average the results for five repeating runs and report as well the standard deviation for each MSE result. For each run, we run for a maximum of 200 epochs in total and stop the training process with 20-epoch early stopping on the validation performance. The data split is done in time order, and hence is deterministic. We report the results using the same hidden size (4) and number of layers (2) for all methods. 

\subsection{Experiments in Section~\ref{sec-exp-unobserved}}
We use feature transformation for layer-wise updating. The head number is set as 1. We set $\tau=0.5$. These datasets do not have input graphs so we only consider learning new structures for our model. For hyper-parameter settings, we conduct grid search for all the models with learning rate from $\{0.0001, 0.0005, 0.005, 0.01, 0.05\}$, weight decay for the Adam optimizer from $\{0.0001, 0.001, 0.01, 0.1\}$, dropout rate from $\{0, 0.2, 0.5\}$, hidden size from $\{32, 64, 100, 200, 300, 400 \} $, number of layers from $\{1, 2, 4, 6, 8, 10, 12\}$. We average the results for five repeating runs and report as well the standard deviation. For each run, we run for a maximum of 600 epochs and report the testing accuracy achieved by the epoch yielding the highest accuracy on validation set.


\section{More Experimental Results}\label{appx-res}

\subsection{Impact of Mini-batch Sizes on Large Graphs}\label{appx-res-batch}

The randomness of mini-batch partition on large graphs has negligible effect on the performance since we use large batch sizes for training, which is facilitated by the linear complexity of \model-s. Even setting the batch size to be 100000, our model only costs 3GB GPU memory on Pokec. As a further investigation on this, we add more experiments using different batch sizes on \texttt{Pokec} and the results are shown in Table~\ref{tab:dis-minibatch}.

\begin{table}[htb]
\small

\centering
\caption{Discussions on using different mini-batch sizes for training on \texttt{Pokec}. We report testing accuracy and training memory for comparison.}\label{tab:dis-minibatch}
{
\setlength{\tabcolsep}{1.5pt}
\centering
\resizebox{0.8\textwidth}{!}{
\begin{tabular}{c|c|c|c|c|c|c}
\hline
Batch size & 5000 & 10000 &20000&50000&100000&200000 \\
\hline
Test Acc (\%) & 65.24 $\pm$ 0.34&67.48 $\pm$ 0.81&68.53 $\pm$ 0.75&68.96 $\pm$ 0.63&69.24 $\pm$ 0.76&69.15 $\pm$ 0.52 \\
GPU Memory (MB)&1244&1326&1539&2060&2928&4011 \\
 \hline
\end{tabular}
}}
\end{table}

One can see that using small batch sizes would indeed sacrifice the performance yet large batch sizes can produce decent and low-variance results with acceptable memory costs.

\subsection{Comparison of Running Time and Memory Costs}\label{appx-res-time}

To further study the efficiency and scalability of our model, we provide more comparison regarding the training time per epoch and memory costs of two \model's variants, GCN, GAT and DenseGAT in Table~\ref{tab-time}. One can see that compared to GAT, \model-s costs comparable time on small datasets such as \texttt{Cora} and \texttt{WikiMath}, and is much faster on large dataset \texttt{Pokec}. As for memory consumption, \model-s reduces the costs by several times over DenseGAT, which clearly shows the efficiency of our new diffusion function designs. Overall, \model-s has nice scalability, decent efficiency and yield significantly better accuracy. In contrast, \model-a costs much larger time and memory costs than \model-s, due to its quadratic complexity induced by the explicit computation for the all-pair diffusivity. Still, \model-a accommodates non-linearity for modeling the diffusion strengths which enables better capacity for learning complex layer-wise inter-interactions.

\begin{table}[htb]
\small

\centering
\caption{Comparison of training time and memory of different models on \texttt{Cora}, \texttt{Pokec}, \texttt{STL-10} and \texttt{WikiMath}. OOM refers to out-of-memory when training on a GPU with 16GB memory.}\label{tab-time}
{
\setlength{\tabcolsep}{1.5pt}
\centering
\resizebox{0.8\textwidth}{!}{
\begin{tabular}{c|c|c|c|c|c|c}
\hline
 \multicolumn{2}{c|}{Method}    & GCN & GAT & DenseGAT  & \model-s & \model-a \\
\hline
\multirow{2}{*}{Cora} & Train time (s)   &  0.0584 & 0.0807 & 0.5165  & 0.1438 & 0.3292 \\
 & Training memory (MB) & 1168 & 1380 & 8460 &  1350 & 3893 \\
\hline
\multirow{2}{*}{Pokec} & Train time (s)   &  1.069 & 14.87 & 88.07 &  2.206 & OOM \\
& Training memory (MB) & 1812 & 2014 & 13174 &  2923 & OOM \\
\hline
\multirow{2}{*}{STL} & Train time (s)   & 0.0069  & 0.0424 & OOM  & 0.0323 & 0.3298 \\
 & Training memory (MB) & 1224 & 1980 & OOM &  1342 & 7680 \\
 \hline
\multirow{2}{*}{WikiMath} & Train time (s)   &  0.0081 & 0.0261 & 0.0364  & 0.0281 & 0.0350 \\
 & Training memory (MB) & 1048 & 1054 & 1316 &  1046 & 1142 \\
 \hline
\end{tabular}
}}
\end{table}

\vskip 0.2in
\bibliography{sample}

\end{document}